\documentclass[default,iicol]{sn-jnl}

\usepackage{graphicx}%
\usepackage{multirow}%
\usepackage{amsmath,amssymb,amsfonts}%
\usepackage{amsthm}%
\usepackage{mathrsfs}%
\usepackage[title]{appendix}%
\usepackage[table]{xcolor}%
\usepackage{textcomp}%
\usepackage{manyfoot}%
\usepackage{booktabs}%
\usepackage{algorithm}%
\usepackage{algorithmicx}%
\usepackage{algpseudocode}%
\usepackage{listings}%
\usepackage{booktabs}
\usepackage{float}
\usepackage{subcaption}
\usepackage[title]{appendix}
\usepackage{ccaption}

\definecolor{green_var}{RGB}{228, 232, 223}
\definecolor{red_var}{RGB}{249, 233, 232}
\theoremstyle{thmstyleone}%
\theoremstyle{thmstyletwo}%

\theoremstyle{thmstylethree}%

\begin{document}

\title[Article Title]{A Multimodal Knowledge-enhanced Whole-slide Pathology Foundation Model}


\author[1]{\fnm{Yingxue} \sur{Xu}}\email{\{yxueb, ywangrm, fzhouaf, jmabq, cheng.jin, syangcw\}@connect.ust.hk}
\equalcont{These authors contributed equally to this work.}

\author[1]{\fnm{Yihui} \sur{Wang}}
\equalcont{These authors contributed equally to this work.}
\author[1]{\fnm{Fengtao} \sur{Zhou}}
\equalcont{These authors contributed equally to this work.}
\author[1]{\fnm{Jiabo} \sur{Ma}}
\author[1]{\fnm{Cheng} \sur{Jin}}
\author[1]{\fnm{Shu} \sur{Yang}}
\author[2,3,4]{\fnm{Jinbang} \sur{Li}} \email{lzcy2008@126.com}
\author[2,3,4]{\fnm{Zhengyu} \sur{Zhang}} \email{1085524719@qq.com}
\author[2,3,4,5]{\fnm{Chenglong} \sur{Zhao}} \email{zcl.125@163.com}
\author[1]{\fnm{Huajun} \sur{Zhou}} \email{csehjzhou@ust.hk}
\author[6]{\fnm{Zhenhui} \sur{Li}} \email{lizhenhui@kmmu.edu.cn}
\author[7]{\fnm{Huangjing} \sur{Lin}} \email{hjlin@cse.cuhk.edu.hk}
\author[8]{\fnm{Xin} \sur{Wang}} \email{xinwang@cuhk.edu.hk}
\author[9,10]{\fnm{Jiguang} \sur{Wang}} \email{jgwang@ust.hk}
\author[11]{\fnm{Anjia}\sur{Han}} \email{hananjia@mail.sysu.edu.cn}
\author[12]{\fnm{Ronald Cheong Kin} \sur{Chan}} \email{ronaldckchan@cuhk.edu.hk}
\author[2,3,4]{\fnm{Li} \sur{Liang}}\email{lli@smu.edu.cn}
\author[13]{\fnm{Xiuming} \sur{Zhang}} \email{xm\_zhang@zju.edu.cn}
\author*[1,9,10,14,15]{\fnm{Hao} \sur{Chen}}\email{jhc@cse.ust.hk}

\affil[1]{\orgdiv{Department of Computer Science and Engineering }, \orgname{The Hong Kong University of Science and Technology}, \orgaddress{\state{Hong Kong SAR}, \country{China}}}
\affil[2]{\orgdiv{Department of Pathology, Nanfang Hospital and School of Basic Medical Sciences}, 
\orgname{Southern Medical University}
\orgaddress{\state{Guangzhou}, \country{China}}}
\affil[3]{\orgname{Guangdong Provincial Key Laboratory of Molecular Tumor Pathology}, \orgaddress{\state{Guangzhou}, \country{China}}}
\affil[4]{\orgname{Jinfeng Laboratory}, \orgaddress{\state{Chongqing}, \country{China}}}
\affil[5]{Department of Pathology, The First Affiliated Hospital of Shandong First Medical University and Shandong Provincial Qianfoshan Hospital, Jinan, Shandong, People's Republic of China}
\affil[6]{\orgdiv{Department of Radiology}, \orgname{The Third Affiliated Hospital of Kunming Medical University, Yunnan Cancer Hospital}, \orgaddress{\state{Kunming}, \country{China}}}
\affil[7]{\orgdiv{Department of Computer Science and Engineering }, \orgname{The Chinese University of Hong Kong}, \orgaddress{\state{Hong Kong SAR}, \country{China}}}
\affil[8]{\orgdiv{Department of Surgery, Prince of Wales Hospital}, \orgname{The Chinese University of Hong Kong}, \orgaddress{Hong Kong}, \country{China}}
\affil[9]{\orgdiv{Department of Chemical and Biological Engineering}, \orgname{The Hong Kong University of Science and Technology}, \orgaddress{\state{Hong Kong SAR}, \country{China}}}
\affil[10]{\orgdiv{Division of Life Science}, \orgname{The Hong Kong University of Science and Technology}, \orgaddress{\state{Hong Kong SAR}, \country{China}}}
\affil[11]{\orgdiv{Department of Pathology, The First Affiliated Hospital}, \orgname{Sun Yat-sen University},
\orgaddress{\state{Guangzhou}, \country{China}}}
\affil[12]{\orgdiv{Department of Anatomical and Cellular Pathology, Prince of Wales Hospital},\orgname{The Chinese University of Hong Kong}, \orgaddress{\state{Hong Kong SAR}, \country{China}}}
\affil[13]{\orgdiv{Department of Pathology, The First Affiliated Hospital, School of Medicine}, \orgname{Zhejiang University}, \orgaddress{\state{Hangzhou}, \country{China}}}
\affil[14]{\orgdiv{HKUST Shenzhen-Hong Kong Collaborative Innovation Research Institute}, Futian, Shenzhen, China}
\affil[15]{State Key Laboratory of Molecular Neuroscience, The Hong Kong University of Science and Technology, Hong Kong, PR China}

\abstract{Remarkable strides in computational pathology (CPath) have been made in the task-agnostic foundation model (FM) that advances the performance of a wide array of downstream clinical tasks. Despite the promising performance, there are still several challenges. First, prior works have resorted to either vision-only or image-caption data, disregarding pathology reports with more clinically authentic information from pathologists and gene expression profiles which respectively offer distinct knowledge for versatile clinical applications. Second, the current progress in pathology FMs predominantly concentrates on the patch level, where the restricted context of patch-level pretraining fails to capture whole-slide patterns. Even recent slide-level FMs still struggle to provide whole-slide context for patch representation. In this study, for the first time, we develop a pathology foundation model incorporating three levels of modalities: microscopic-level pathology slides, macroscopic-level pathology reports created by experts, and molecular-level gene expression data, which resulted in 26,169 slide-level modality pairs from 10,275 patients across 32 cancer types, amounting to over 116 million pathological patch images. To leverage these data for CPath, we propose a novel whole-slide pretraining paradigm that injects the multimodal whole-slide context into the patch representation, called \textbf{M}ultimodal \textbf{S}elf-\textbf{TA}ught P\textbf{R}etraining (mSTAR). The proposed paradigm revolutionizes the pretraining workflow for CPath, enabling the pathology FM to acquire the whole-slide context. To the best of our knowledge, this is the first attempt to incorporate three modalities at the whole-slide context for enhancing pathology FMs, expanding the modelling context from single to multiple modalities and from patch-level to slide-level. To systematically evaluate the capabilities of mSTAR, we built the largest spectrum of oncological benchmark, spanning 7 categories of oncological applications in 15 types of 97 practical oncological tasks. The experimental results show that mSTAR outperformed previous state-of-the-art (SOTA) FMs, particularly excelling in molecular prediction, report-related pathology tasks, and multimodal fusion. This highlights that incorporating more pathology-related modalities at the slide level during pre-training significantly enhances the generalizable capabilities related to the respective modalities, which validates the scalability of modalities in developing pathology foundation models.}




\maketitle
\section{Introduction}\label{sec1:intro}
\begin{figure*}
    \centering
    \includegraphics[scale=0.52]{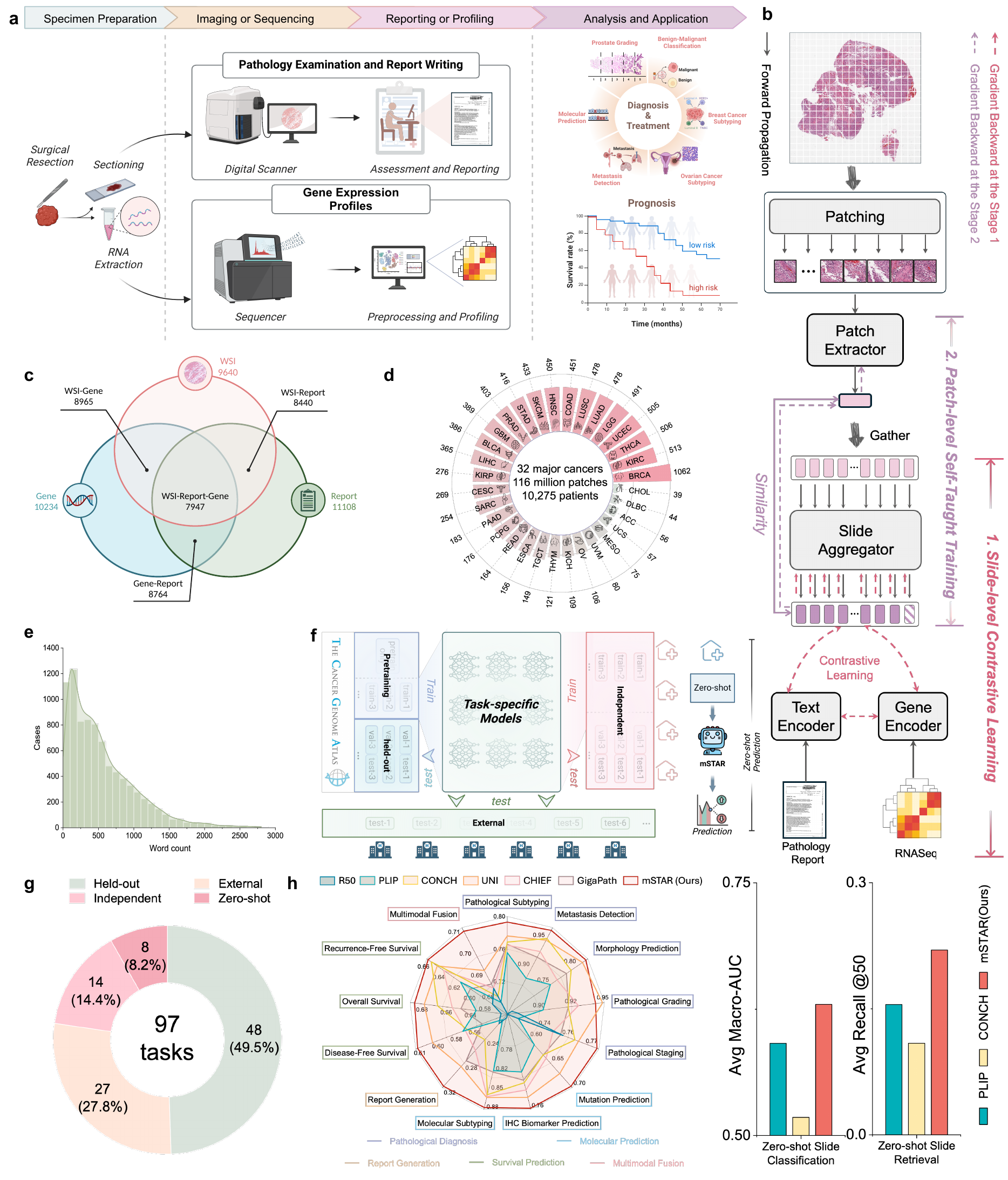}
\end{figure*}
\begin{figure*}
    \caption{\textbf{Overview of the study}. \textbf{a}, The workflow in clinical practice for diagnosis, treatment and prognosis of oncology, which primarily involves three common modalities data: WSIs, pathology reports and gene expression profiles. \textbf{b}, \textbf{The overview of mSTAR paradigm}. mSTAR consists of two stages: 1) Slide-level Contrastive Learning, and 2) Patch-level Self-Taught Training. \textbf{c-e}, statistics of data used in this study, including \textbf{c)} Venn Graph of cases across various modalities, \textbf{d)} the number of cases in pretraining data across different cancer types. \textbf{e)} the distribution of word count for pathology reports. \textbf{f}, evaluation scheme in this study: including held-out, independent, external and zero-shot. The illustration is presented in Sec.~\ref{sec:eval}. \textbf{g}, the distribution of datasets across different types of tasks for different evaluation scheme, and the detailed information about every dataset is presented in \textbf{Extended Data} Table~\ref{tab:all_ds}.
    \textbf{h}, The average performance spanning 15 types of 97 tasks across 7 categories of applications: Pathological Diagnosis, Molecular Prediction, Report Generation, Survival Prediction, Multimodal Fusion, Zero-shot Slide Classification, and Zero-shot Slide Retrieval. Zero-shot tasks, which require a well-aligned vision-language space, are evaluated for vision-language models only, i.e., PLIP, CONCH and mSTAR. (See \textbf{Extended Data} Table~\ref{tab:overall_performance})}
    \label{fig:overview}
\end{figure*}

The recent advancements in foundation models (FMs)~\cite{huang2023visual,chen2024towards,lu2024visual,xu2024whole,alfasly2023foundation} for computational pathology (CPath) have demonstrated considerable progress in an incredibly broad spectrum of clinical tasks, such as cancer diagnosis, treatment and prognosis. Despite encouraging performance in general-purpose pathology foundation models, there are still several unresolved challenges. 

First, massive multimodal data in line with clinical practices is under-utilized for pretraining, such as pathology reports and gene expression profiles. Existing pathology FMs either focus on vision-only~\cite{chen2024towards} or image-caption data~\cite{huang2023visual,lu2024visual}, in which the information provided by captions is insufficient to provide whole slide context for authentic slide-level oncological tasks although attempting to incorporate different modalities. The power of multimodal data has been repeatedly substantiated not only in the general machine learning community~\cite{team2023gemini,khattak2023maple} but also in the field of medical cancer research~\cite{chen2021multimodal,xu2023multimodal,zhou2023cross}. In the clinical workflow, as shown in \textbf{Fig.}~\ref{fig:overview}a, pathology reports often provide the most clinically relevant information of whole slides in real-world scenarios, while patients' gene expression profiles offer insights into quantitative molecular dynamics that can complement the qualitative morphological view provided by a slide. The integration of these slide-level multimodal data can establish a broad and holistic perspective, thereby undoubtedly enhancing the capabilities of pathology FMs to perform various clinical tasks.

Second, existing efforts in pathology FMs are predominantly aimed at the modelling of patch/ROI-level data~\cite{huang2023visual,chen2024towards, lu2024visual}, leading to limited contexts for slide-level oncological applications. Conventional models typically treat individual patch images as independent samples for pretraining a patch extractor, and subsequently employ multiple instance learning (MIL)~\cite{ilse2018attention,clam,shao2021transmil} to perform slide-level modelling based on embedded patch features. Recent concurrent works~\cite{xu2024whole,chief} have attempted to pretrain the slide-level FM. However, they are also achieved by pretraining a slide aggregator on top of pre-extracted patch features with a fixed trained patch extractor. This way poses an inherent limitation that the upper bound of pretraining performance is inevitably constrained by the quality of patch features. Furthermore, as the patch extractor and whole-slide aggregator undergo independent training processes, they still fail to provide a multimodal whole-slide context for patch representation, leading to suboptimal performance. 

\begin{figure*}[t]
    \centering
    \includegraphics[scale=0.47]{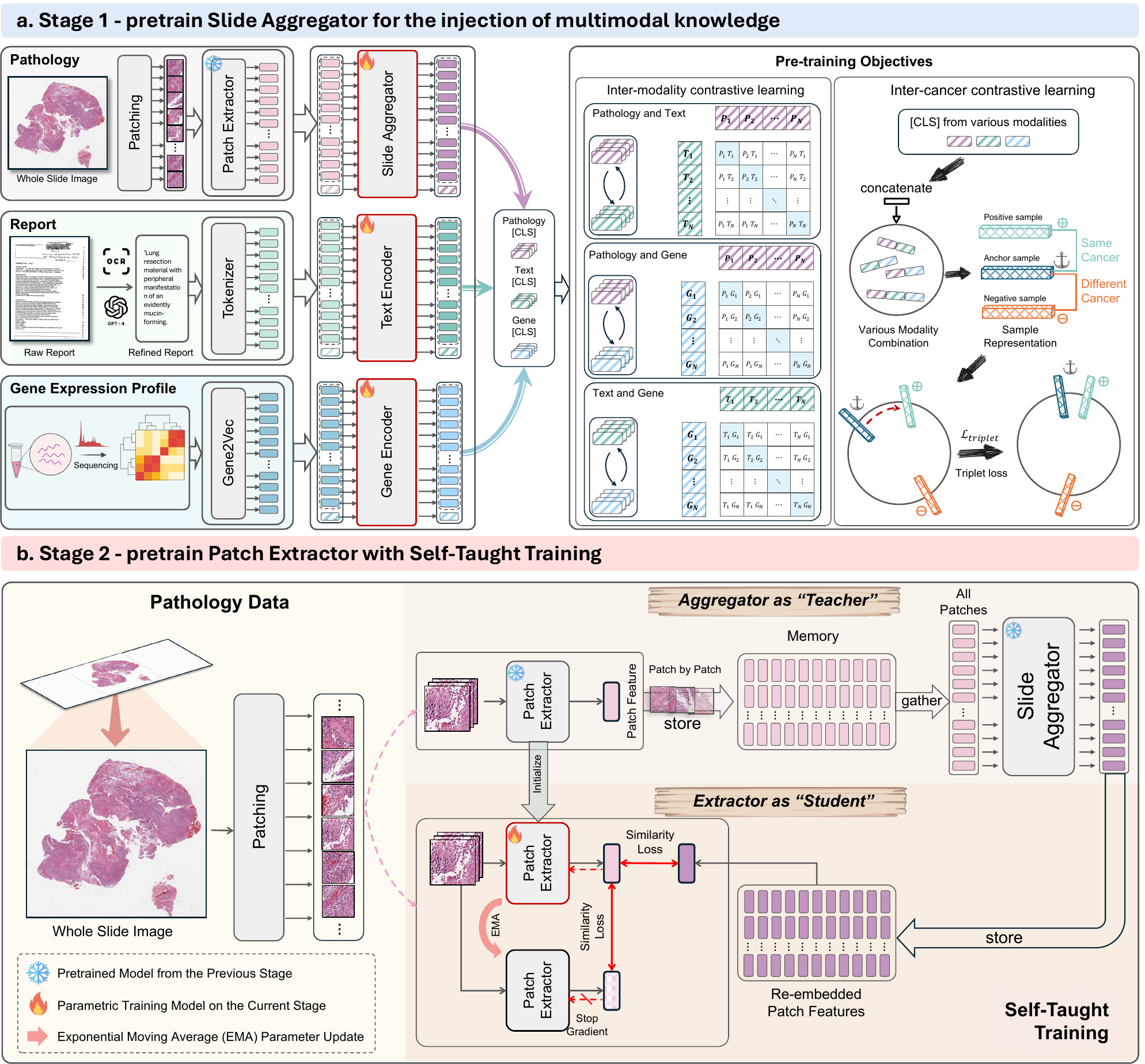}
    \caption{\textbf{The Overview of mSTAR Pipeline}. mSTAR is a whole-slide pretraining paradigm comprising two-stage pretraining. \textbf{a}, Stage 1 aims to inject multimodal knowledge into a slide aggregator by slide-level contrastive learning among WSIs, pathology reports and gene expression data. \textbf{b}, Stage 2 aims to seamlessly propagate multimodal knowledge learned at the slide level into the patch extractor by Self-Taught training, which leverages the slide aggregator pretrained in Stage 1 as ``Teacher'' and enforces patch extractor to be ``Student''.}
    \label{fig:framework}
\end{figure*}

In this study, for the first time, we simultaneously incorporate three-level complementary modalities for developing pathology foundation models: microscopic-level pathology slides, macroscopic-level pathology reports created by specialized pathologists, and molecular-level gene expression data. To build up this model, we collected 26,169 slide-level modality pairs from 10,275 patients across 32 cancer types (\textbf{Fig.}~\ref{fig:overview}c-e), encompassing over 116 million pathological patch images. To leverage these multimodal data for CPath, we developed a novel whole-slide pretraining paradigm, termed \textbf{M}ultimodal \textbf{S}elf-\textbf{TA}ught P\textbf{R}etraining (mSTAR) (\textbf{Fig.}~\ref{fig:overview}b and ~\ref{fig:framework}), which injects multimodal whole-slide context into the patch representation for the first time. Specifically, this paradigm first pretrains a slide aggregator that absorbs multimodal knowledge via slide-level contrastive learning at the first stage. This slide aggregator will act as a bridge that injects whole-slide contextual multimodal knowledge into the patch extractor at the second stage through self-taught training. 
The proposed paradigm revolutionizes the workflow of pretraining for computational pathology, which allows the pathology FM to possess powerful whole-slide abilities, broadening the scope of contextual modelling from unimodal to multimodal knowledge and from patch-level to slide-level. 

To systematically investigate the capabilities of mSTAR in real-world clinical scenarios, we evaluate 7 categories of oncological applications from 15 types of 97 practical clinical tasks, covering patients' oncological journey (\textbf{Fig.}~\ref{fig:overview}f and \textbf{Extended Data} Table~\ref{tab:all_ds}). This results in the largest spectrum of downstream benchmark to date, in which half of the datasets are external or independent cohorts from multi-center medical institutions, underscoring the evaluation of generalization ability for foundation models.

The experimental results have demonstrated that, mSTAR showcased superiority in 14 out of 15 types of tasks on average compared to previous state-of-the-art (SOTA) FMs (\textbf{Fig.}~\ref{fig:overview}g), especially in molecular prediction, report generation, multimodal fusion and zero-shot tasks with significant margins. These results underscore that the incorporation of more modalities at slide level in the pre-training substantially endows pathology foundation models with enhanced capabilities related to the respective modalities. Therefore, this validates the impressive scalability of modalities in establishing pathology foundation models, which potentially provide the guidance for enhancing pathology foundation models by injecting knowledge of more pathology-related modalities.

\section{Results}\label{sec2}
\subsection{The Overview of mSTAR}
The proposed mSTAR aims to provide a novel whole-slide pretraining paradigm that injects multimodal knowledge into the pathology foundation model. Compared with existing pathology foundation models, mSTAR has the following innovative designs to fully unleash its power in a wide spectrum of pathological downstream tasks. First, clinical multimodal data are fully harnessed in pretraining to endow the pathology FM with multimodal knowledge for comprehensive perspectives in clinical tasks. Second, the whole-slide pretraining paradigm provides an alternative way to obtain whole-slide contexts for pathology FMs through self-taught training. To the best of our knowledge, this is the first work to inject multimodal knowledge at the whole-slide context into a pathology FM, broadening the contextual understanding for CPath from patch-level to slide-level and from unimodal to multimodal knowledge. The overview of mSTAR is shown in \textbf{Fig.} ~\ref{fig:framework}, consisting of two stages of pretraining.

In the first stage, the objective is to inject multimodal knowledge into the slide aggregator by slide-level contrastive learning among three modalities, i.e., WSIs, pathology reports and RNA-Seq profiles. Note that the pretrained slide aggregator will act as a bridge that propagates multimodal knowledge into the patch extractor in the next stage. To this end, as shown in \textbf{Fig.}~\ref{fig:framework}a, we first utilized a pretrained patch extractor, a state-of-the-art pathology foundation model named UNI~\cite{chen2024towards}, to encode each patch image of a slide into patch features. Then the resulting patch features are fed into a slide aggregator and integrated into a slide-level representation which is subsequently aligned with other modalities through inter-modality contrastive learning. Furthermore, to mitigate the influence of heterogeneity across different types of cancers, the pretraining of the slide aggregator is also supervised by inter-cancer contrastive learning. This approach brings samples of the same cancer type closer together while concurrently pushing samples of different cancer types apart.

In the second stage, the pretrained slide aggregator acquiring multimodal knowledge, can serve as the teacher model to seamlessly propagate multimodal knowledge at the slide-level context into the patch extractor, called Self-Taught Training (\textbf{Fig.} ~\ref{fig:framework}b). Specifically, the patch extractor is pretrained through encouraging the extracted patch features to be as similar as possible to those re-embedded by the pretrained aggregator. At the same time, to avoid catastrophic forgetting, we also enforce a similarity constraint between the extracted features and those embedded by the exponential moving average (EMA) patch extractor.
 
With these two stages, multimodal knowledge at the whole-slide context can be seamlessly embedded into foundation models. As a result, the model acquires the ability to comprehend both patches and the entire WSI, which facilitates downstream tasks at different levels. In the end, the pathology foundation model can achieve advanced abilities with the extended context from patch-level to slide-level and from unimodal to multimodal knowledge. More details of mSTAR can be found in Section~\ref{sec:methods_pretrain}.

To validate the effectiveness of each component, we conducted ablation studies on different combinational modalities (\textbf{Extended Data} Table ~\ref{tab:abla_modal}) and different pretraining loss functions (\textbf{Extended Data} Table ~\ref{tab:abla_cancer}). Through ablating each component, we demonstrate the effectiveness of every part based on changes in performance. To this end, 9 held-out datasets for survival analysis are used to evaluate the ablated variants. We need to clarify that the purpose of this experiment is to validate whether adding each component is effective, hence it focused solely on evaluating the performance of the pretrained aggregator (stage 1) without injecting multimodal knowledge into the patch extractor (stage 2).
From the results in the \textbf{Extended Data} Table ~\ref{tab:abla_modal}, we observed that the contributions from combinations of Pathology+Reports and Pathology+RNASeq to performance are comparable, and the synergy of three modalities resulted in further improvements. For the ablation of pertaining loss functions, both objective functions make a difference in the pretraining, and their combination elevates performance to a higher level.
\begin{figure*}
    \centering
    \includegraphics[scale=0.45]{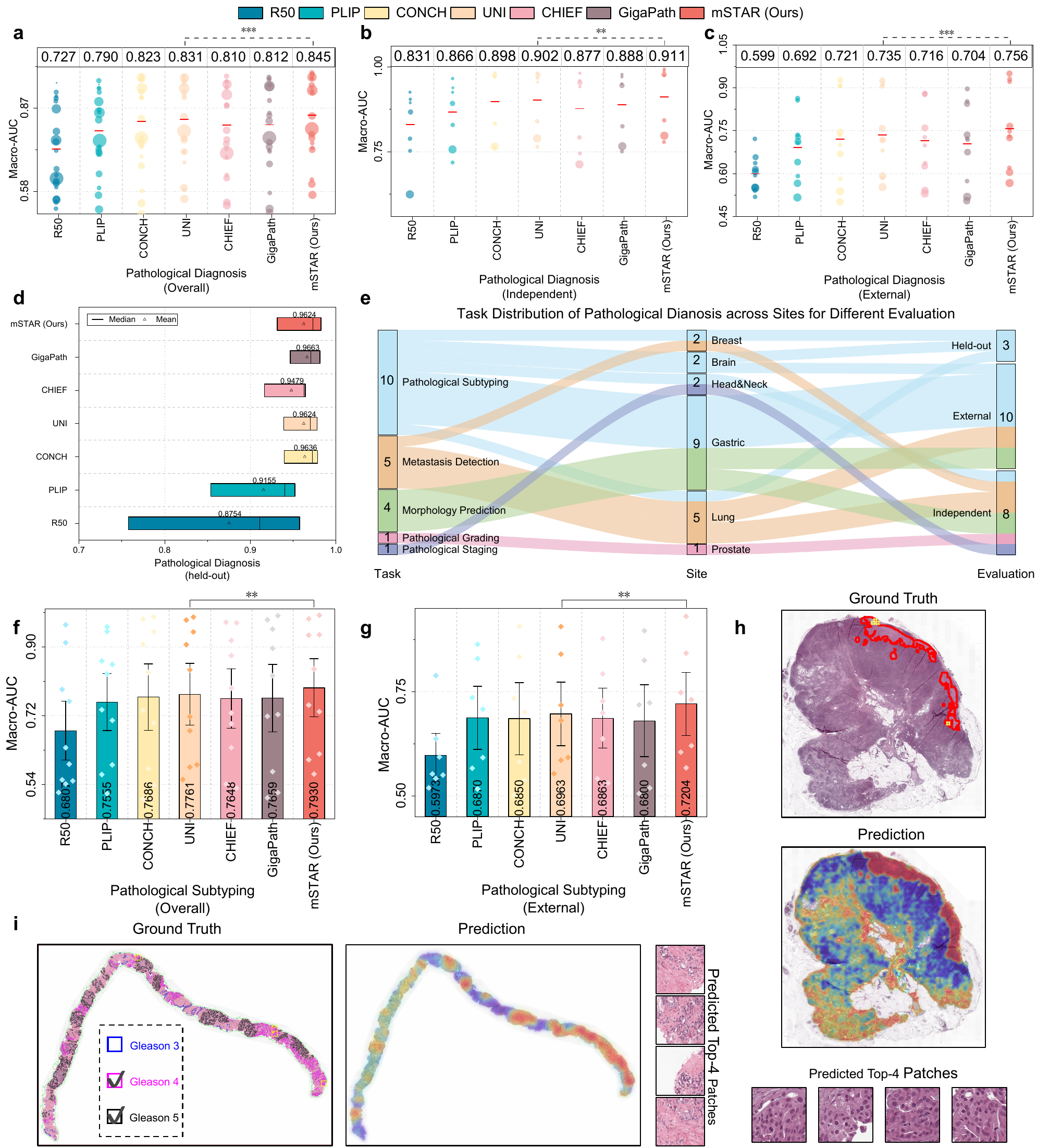}
    \caption{\textbf{Performance of Pathological Diagnosis} on 21 datasets. \textbf{a}, The overall performance on pathological diagnosis. \textbf{b}, The performance on 8 independent datasets. \textbf{c}, The performance on 10 external datasets. The {\color{red}{red}} lines and the values reported at the top of figures \textbf{a}, \textbf{b} and \textbf{c} refer to the averaged performance across datasets. Each point represents a dataset, with the size of the point indicating the standard deviation. \textbf{d}, The performance on 3 held-out datasets. \textbf{e}, Task distribution of pathological diagnosis across sites for different evaluation. \textbf{f}, The overall performance on Pathological Subtyping across 10 datasets. \textbf{g}, The performance on 6 external datasets of Pathological Subtyping. \textbf{h-i} The visualized validation of attention scores from mSTAR on h) CAMELYON and i) PANDA datasets. \textit{P-value} for every group of experiments is given through one-sided Wilcoxon signed-rank test between mSTAR and the second-best FM. * represents $P<0.05$, ** means $P<0.01$ and *** indicates $P<0.001$. Detailed Performances of every dataset are presented in \textbf{Extended Data} Fig.~\ref{fig:dianosis-raw} and Tab.~\ref{tab:diagnosis}.}
    \label{fig:dianosis}
\end{figure*}
\subsection{Pathological Diagnosis}
We start with evaluating the pathological diagnostic capabilities based on pathological morphology, including pathological subtyping, metastasis detection, morphology prediction, pathological grading and pathological staging. These tasks commonly appear in pathology reports, forming a fundamental component of such reports and thus holding significant clinical importance. To evaluate these tasks, we collected 21 datasets from both publicly available and institutional sources consisting for 3 types of evaluation strategies, i.e., 8 independent cohorts on the 7:1:2 split, 3 held-out cohorts that are TCGA data held out from pretraining data and 10 external cohorts for testing only.

Specifically, for pathological subtyping task, we include breast cancer on BRCA-PathSubtype~\cite{weinstein2013cancer} as a held-out cohort, brain tumor on GBMLGG\_PathSubtype~\cite{weinstein2013cancer} as a held-out cohort and EBrains\_PathSubtype~\cite{ebrains_data} as an external cohort, head and neck cancer on HANCOCK\_PathSubtype~\cite{hancock} as an independent cohort, gastric cancer NFGC\_PathSubtype, YN1\_PathSubtype and YN3\_PathSubtype as 3 external cohorts, lung cancer on TCGA-NSCLC~\cite{weinstein2013cancer} as a held-out cohort and Lauren classification of gastric cancer on NFGC\_Lauren and YN3\_Lauren as two external cohorts, resulting in 3 held-out, one independent and 6 external cohorts. For metastasis detection task, we perform breast metastasis detection on CAMELYON~\cite{c16_dataset,c17_dataset} as an independent cohort, lung metastasis detection on NF\_Metastatic as an independent cohort and QFS\_Metastatic as an external cohort, and meanwhile we further predict their primary locations on NF\_Metastatic\_Fine as an independent cohort and QFS\_Metastatic\_Fine as an external cohort. For morphology prediction, we assess whether perineural invasion is present on NFGC\_Perineural as an independent cohort and YN3\_Perineural as an external cohort, while we evaluate whether vascular invasion is present on NFGC\_Vascular as an independent cohort and YN3\_Vascular as an external cohort. Additionally, we evaluate pathological grading on PANDA~\cite{panda_dataset} as an independent cohort and pathological staging on HANCOCK-TStage~\cite{hancock} as an independent cohort. The task distribution for pathology diagnosis is demonstrated in \textbf{Fig.}~\ref{fig:dianosis}e, which covers common types of cancerous sites. The details of every dataset are described in Section~\ref{sec:downstream}.

As baselines, we evaluate the recent pathology foundation models (FMs) including PLIP~~\cite{huang2023visual}, CONCH~\cite{lu2024visual}, UNI~\cite{chen2024towards}, CHIEF~\cite{chief} and GigaPath~\cite{gigapath} as well as the classical R50~\cite{he2016identity}.
To perform these tasks, following the standard practice in computational pathology~\cite{chen2024towards}, we used foundation models to extract features from each patch and adopted attention-based multiple instance learning (ABMIL)~\cite{ilse2018attention} trained from scratch as the slide-level aggregator to perform slide-level prediction. ABMIL is a simple yet robust MIL approach, which is usually used for evaluation in previous foundation model research~\cite{chen2024towards,lu2024visual}. In particular, CHIEF and GigaPath claimed that their patch extractor should be used in conjunction with their pretrained aggregator. Therefore, we finetuned their pretrained aggregator paired with their extracted features on every downstream dataset to ensure the best performance.

All comparisons are based on the metric of Macro-AUC, a commonly used and objective classification metric, which does not rely on the selection of the decision threshold and is insensitive to the sample ratio of various classes. To examine statistical differences between mSTAR and the second-best FMs, the one-sided Wilcoxon signed-rank test was performed on various datasets.

From an overall perspective, we assessed the average performance for mSTAR and compared foundation models across 21 diverse datasets. The overall result demonstrates that mSTAR achieved the best performance with a +1.37\%  increase overall ($P < 0.001$) compared to the second-best model, UNI, as shown in \textbf{Fig.}~\ref{fig:dianosis}a. Compared with slide-level FMs, mSTAR obtained +3.31\% ($P < 0.001$) performance gain over GigaPath, the best slide-level baseline. From the perspective of consistency, mSTAR stood out on 18 out of 21 datasets, ranking at the first place. To evaluate the generalizability, we assessed 8 independent datasets and 10 external datasets. For independent cohorts, mSTAR demonstrates about +1\% improvement ($P < 0.01$) compared to the second-best FM and +2.32\% increase ($P < 0.001$) over GigaPath. It is worth noting that mSTAR exhibits superior generalizable capability on external cohorts with +2.14\% improvement ($P < 0.001$) over the second-best FM, and meanwhile exhibits +4.16\% ($P < 0.001$) increase over GigaPath. For held-out cohorts, mSTAR showcases performance comparable to that of other FMs.

Pathological subtyping is of utmost importance in clinical practice, as it forms the foundation for developing personalized treatment plans and enhancing treatment effectiveness. Therefore, we specifically evaluate mSTAR's performance on such crucial tasks. The overall performance across 10 datasets demonstrates +1.69\% increase ($P < 0.01$) compared to the second-performing FM. Furthermore, mSTAR shows significant improvement of +2.41\%  ($P < 0.01$) on external cohorts, suggesting the strong generalizable capability.

To validate whether the predictions of mSTAR align with clinical understanding, we visualize the predicted attention scores from mSTAR and compared the results with the given human-annotated ground-truth ROI on CAMELYON and PANDA datasets, as shown in \textbf{Fig.}~\ref{fig:dianosis}h and i. 

From the results of \textbf{Fig.}~\ref{fig:dianosis}h and i, we can see the areas of interest for mSTAR successfully matches with the ground truth. These indicate mSTAR possesses intelligence for pathological diagnosis.

\subsection{Molecular Prediction}
\begin{figure*}
    \includegraphics[scale=0.47]{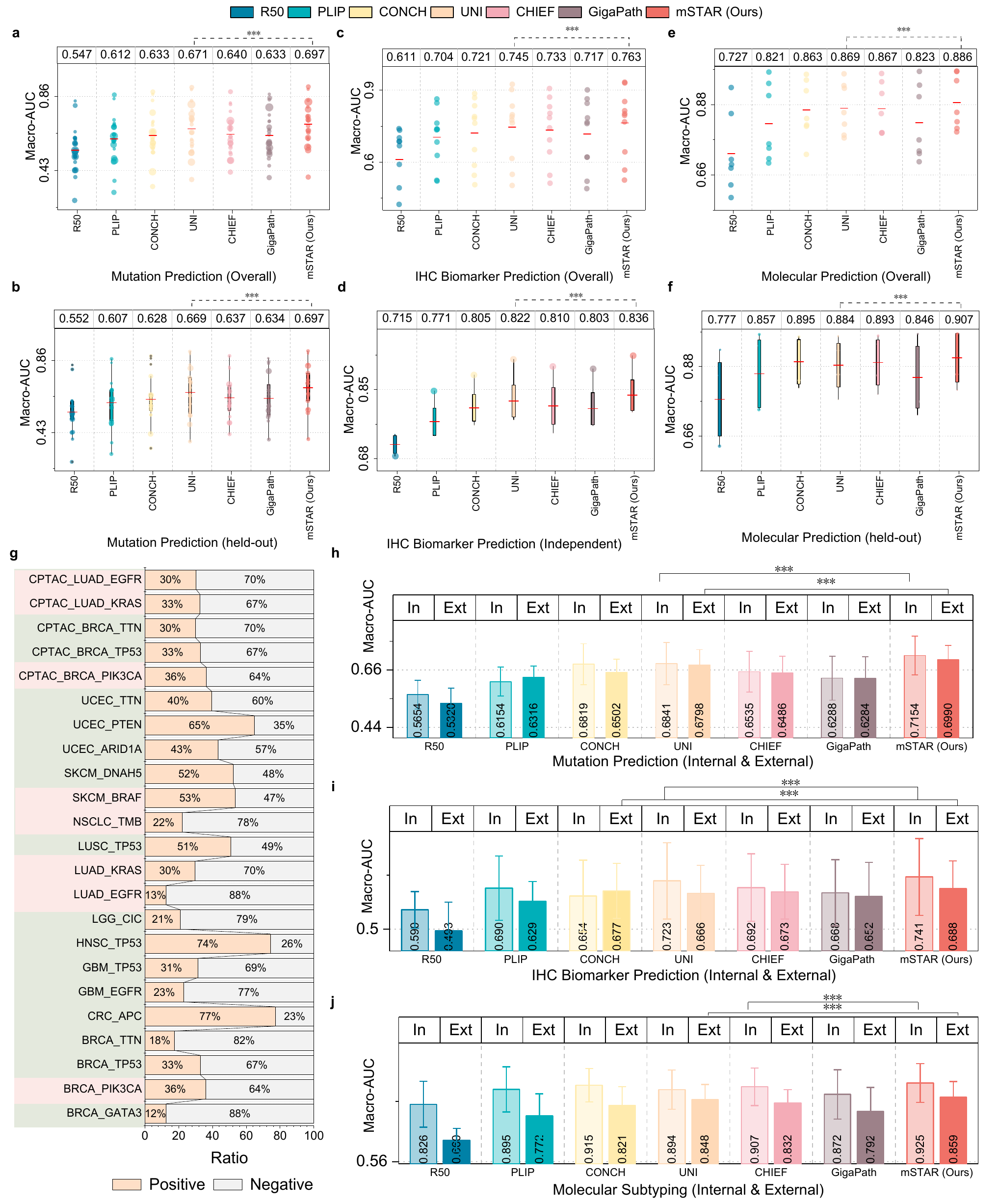}
\end{figure*}
\begin{figure*}
\centering
    \caption{\textbf{Performance of Molecular Prediction} on 40 datasets across 10 cancer types. \textbf{a}, Overall Performance of \textbf{Gene Mutation Prediction} on 23 datasets. \textbf{b}, Performance of Mutation Prediction on 18 held-out datasets. \textbf{c}, Overall Performance of \textbf{Immunohistochemistry (IHC) Biomarker Prediction} on 10 datasets. \textbf{d}, Performance of IHC Biomarker Prediction on 4 independent datasets. \textbf{e}, Overall Performance of \textbf{Molecular Subtyping} on 7 datasets. \textbf{f}, Performance of Molecular Subtyping on 4 held-out datasets. The {\color{red}{red}} lines and the values reported at the top of figures \textbf{a-f} refer to the averaged performance across datasets. Each point represents a dataset, with the size of the point indicating the standard deviation. \textbf{g}, Positive and Negative Ratios of gene mutation for every mutation dataset, including genes with \colorbox{green_var}{high-frequency mutations} highlighted in green and genes related to \colorbox{red_var}{FDA-approved therapies} highlighted in red. \textbf{h-j, Internal (In) v.s. External (Ext) Evaluation.} \textbf{(h)}, Performance of Mutation Prediction on 5 internal and 5 external datasets. \textbf{(i)}, Performance of IHC Biomarker Prediction on 3 internal and 3 external datasets. 
    \textbf{(j)}, Performance of Molecular Subtyping on 3 internal and 3 external datasets. \textit{P-value} for every group of experiments is given through one-sided Wilcoxon signed-rank test between mSTAR and the second-best FM. * represents $P<0.05$, ** means $P<0.01$ and *** indicates $P<0.001$. Detailed performances of every dataset spanning 10 cancer types are presented in \textbf{Extended Data} Fig.~\ref{fig:molecular-raw} and Tab.~\ref{tab:mutation_abmil}-\ref{tab:molecular}.}
    \label{fig:molecular-main}
\end{figure*}
Molecular prediction has significant clinical implications in targeted therapy and risk stratification, allowing for tailored treatment plans. For example, HER2-positive breast cancer can be treated with HER2-targeted drugs like trastuzumab~\cite{her2targeted}. However, unlike pathological examination, genome sequencing remains largely inaccessible, especially in underdeveloped areas due to its high cost. Fortunately, with the benefit of pretraining on the modality combination of WSIs and gene expression data, mSTAR is more likely to possess a promising capability of molecular prediction only based on easily accessible pathological images. Therefore, in this study, we investigate molecular prediction solely based on pathological images for 3 categories of crucial molecular tasks, including gene mutation prediction, immunohistochemistry (IHC) biomarker prediction and molecular subtyping. To this end, we collected 35 datasets sourced from both public databases and medical institutions for evaluation and followed the same setting as the pathological diagnosis for slide-level prediction. The results are as follows:
\\
\\
\noindent
\textbf{Gene Mutation Prediction.} Following the setting of CHIEF~\cite{chief}, we predicted gene mutation related to FDA (Food and Drug Administration)-approved targeted therapies presented in OncoKB~\cite{chakravarty2017oncokb} and high-frequency mutations~\cite{highfremuation} in 10 cancer types held out from pretraining data. The positive mutation ratios are presented in \textbf{Fig.}~\ref{fig:molecular-main}g. The overall performance (\textbf{Fig.}~\ref{fig:molecular-main}a) exhibits the superiority of mSTAR in mutation prediction with +2.6\% increases ($P < 0.001$) in Macro-AUC, compared to the second-best FM. On held-out cohorts, mSTAR surpassed the second-best FM by +2.81\%. In particular, among the 18 genes (\textbf{Extended Data} Fig.~\ref{fig:molecular-raw}a and Table~\ref{tab:mutation_abmil})-\ref{tab:mutation_ext}, mSTAR excels in the prediction of \textit{ARID1A} in endometrial carcinoma (UCEC) with +5.23\% increase ($P < 0.001$), \textit{KRAS} with +5.14\% ($P < 0.001$) in lung adenocarcinoma (LUAD), \textit{GATA3} with +3.2\% ($P < 0.001$) improvement and \textit{PIK3CA} +2.46\% ($P < 0.001$) increase in invasive breast carcinoma (BRCA), \textit{KRAS} in cutaneous melanoma (SKCM) with +2.73\% increase and \textit{EGFR} with +1.81\% ($P < 0.001$) in LUAD. All of these gene mutations have significant clinical relevance~\cite{ucec-ARID1A,KRAS-luad1,KRAS-luad2,gata3-brca,PIK3CA-brca,egfr-luad1,egfr-luad2}, indicating the potential of mSTAR in biomedical research.

For external validation, mSTAR still obtains about 0.7 Macro-AUC on average and outperformed the second-best FM by about +2\% improvements across 5 external cohorts (\textbf{Fig.}~\ref{fig:molecular-main}h). Specifically,  mSTAR achieves +2.04\% improvement ($P < 0.001$) in \textit{TP53} of BRCA, +2.06\% increase ($P < 0.001$) in \textit{EGFR} of LUAD and +1.11\% increase ($P < 0.001$) in \textit{KRAS} of LUAD. 

Additionally, mSTAR predicts the mutation status in 14 of the 18 genes with Macro-AUC greater than 0.6 on held-out cohorts, as shown in \textbf{Extended Data} Fig.~\ref{fig:molecular-raw}b and Table~\ref{tab:mutation_abmil}. Mutations with excellent performance greater than 0.8 include \textit{TP53} in BRCA (0.8366; 95\% CI 0.8105-0.8627) and glioblastoma multiforme (GBM) (0.8282; 95\% CI 0.7780-0.8784), \textit{CIC} in low-grade glioma (LGG) (0.9157; 95\% CI 0.8952-0.9362), \textit{PTEN} in UCEC (0.9008; 95\% CI 0.8737-0.9279). This showcases mSTAR can provide reliable prediction of these crucial biomarkers~\cite{tp53-breast, tp53-gbm, cic-lgg, pten-ucec1, pten-ucec2} for biomedical research.
\\
\\
\textbf{IHC Biomarker Prediction.} Immunohistochemistry (IHC) is widely used in clinical pathology, primarily for detecting specific proteins in tissue samples and distinguishing between tumors with similar pathological features, enabling more precise targeted therapy and improving patients' outcomes. However, IHC examination usually requires extra expensive costs. Therefore, if easily accessible H\&E slides can be used to predict IHC biomarkers, it would significantly advance the widespread adoption of precision cancer diagnostics, especially in underdeveloped areas. To assess mSTAR's performances on IHC biomarker prediction, we collected 10 datasets for evaluation, including 3 held-out datasets and 3 external as well as 4 independent datasets from collaborative medical institutions. Specifically, we involved common biomarkers comprising ER, HER2, PR and CK5 for breast cancer, along with CK7 for lung cancer.

From an overall perspective, mSTAR outperforms the second-performing FM by +1.8\% ($P < 0.001$, \textbf{Fig.}~\ref{fig:molecular-main}c). Across 4 independent datasets, mSTAR performs the best overall (\textbf{Fig.}~\ref{fig:molecular-main}d) by +1.44\% ($P < 0.001$) over UNI, the second-best FM, while consistently surpassing other FMs on all 4 independent datasets (\textbf{Extended Data} Fig.~\ref{fig:molecular-raw}b) with significant differences ($P < 0.001$). Specifically, mSTAR can achieve over 0.8 of Macro-AUC on 3 out of 4 tasks including ER and CK5 for breast cancer and CK7 for lung cancer, as well as almost 0.8 of Macro-AUC on HER2 (0.7951$\pm$0.0125). This indicates mSTAR is capable of offering reliable prediction for these vital and common biomarkers, probably resulting in a great reduction of the cost of IHC examination.

To test the generalization of mSTAR, 3 external datasets spanning ER, PR and HER2 in breast cancer are collected from the collaborative hospital for evaluation. The overall performance is presented in \textbf{Fig.}~\ref{fig:molecular-main}g. mSTAR consistently showcases superiority in both internal and external evaluations with +1.81\% ($P < 0.001$) and +1.02\% ($P < 0.001$) increases over the second-best FMs, UNI and CONCH, respectively. For the examination of ER (\textbf{Extended Data} Table~\ref{tab:IHC}), although we observed a decline in performance compared to the internal cohorts, mSTAR still maintains Macro-AUC above 0.85 (0.8526$\pm$0.0071), resulting in the promising generalization in the external cohort.
\\
\\
\textbf{Molecular Subtyping} aims to categorize cancers based on their molecular and genetic characteristics, thereby assisting in identifying patients with distinct responses to treatment, and prognostic outcomes. In this study, we investigate molecular subtyping on 4 cancers including Breast Invasive Carcinoma (BRCA), Colon Adenocarcinoma and Rectum Adenocarcinoma (CRC), Glioblastoma Multiforme and Brain Lower Grade Glioma (GBMLGG) and Head and Neck Squamous Cell Carcinoma (HNSC) on 4 held-out datasets (BRCA\_MolSubtype, CRC\_MolSubtype, GBMLGG\_MolSubtype and TCGA\_HNSC\_HPV) and 3 external datasets (ZJ1\_Breast MolSubtype, EBrains\_MolSubtype and HANCOCK\_HPV).

From the overall perspective, we observed a performance gain of +1.78\% ($P < 0.001$) over UNI, the second-best FM (Fig.~\ref{fig:molecular-main}e). When we delve into different evaluation strategies, +1.24\% ($P < 0.001$) improvement can be seen in held-out cohorts (Fig.~\ref{fig:molecular-main}f). When taking a close on internal and external cohorts of breast, brain and head\&neck cancers (Fig.~\ref{fig:molecular-main}j), mSTAR surpassed CHIEF (the second-best FM on internal cohorts) by +1.77\%, while exceeding UNI (the second-best FM on external cohorts) by +1.1\%. It is worth noting that in the FMs compared, mSTAR is the only one that maintains an AUC above 0.85 for both internal and external datasets, demonstrating strong generalization ability. Furthermore, mSTAR keeps the consistent superiority over 3 external cohorts (\textbf{Extended Data} Table~\ref{tab:molecular}).

To sum up, through the joint pretraining of pathological images and gene expression data, mSTAR demonstrates superior performance and strong generalization across mutation prediction, IHC biomarker prediction and molecular subtyping. This capability can provide reliable predictions in clinical applications and biomedical research, making it possible to utilize accurate molecular information in a cost-effective manner.
\subsection{Vision-Language Evaluation}
\begin{figure*}
    \centering
    \includegraphics[scale=0.43]{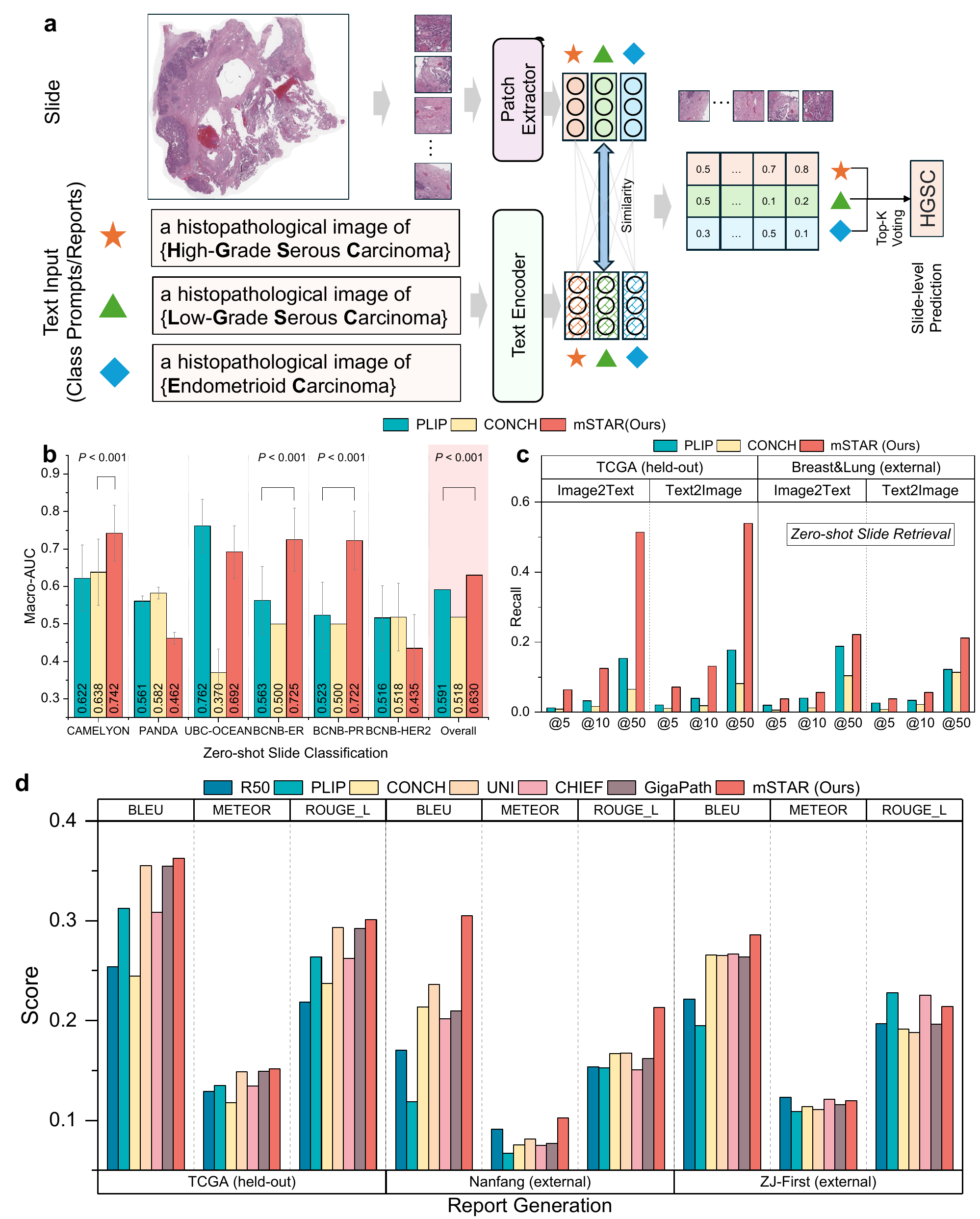}
    \caption{\textbf{Vision-language Evaluation.} \textbf{a, The scheme of zero-shot evaluation.} For zero-shot classification, we used class prompts as the text input. For zero-shot retrieval, the text input is a pathology report. \textbf{b, Performance of zero-shot slide classification} on 6 independent datasets. The `Overall' refers to the averaged performance across these 6 datasets. \textit{P-value} is given through one-sided Wilcoxon signed-rank test between mSTAR and the second-best FM. \textbf{c, Performance of zero-shot retrieval} on an external dataset for Image-to-Text and Text-to-Image tasks. The results on held-out TCGA dataset are presented for reference only to be compared with zero-shot's capability. \textbf{d, Performance of report generation} on one held-out TCGA dataset and two external datasets. \textit{P-value} for every group of experiments is given through one-sided Wilcoxon signed-rank test between mSTAR and the second-best FM. Detailed performances of every dataset are presented in \textbf{Extended Data} Tab.~\ref{tab:zeroshot_overall}-\ref{tab:report_gen}.}
    \label{fig:vision-language}
\end{figure*}
Strong language-related capabilities are one of the key features of foundational models, reflecting their potential in open-world scenarios where downstream tasks are conducted without further training, that is, zero-shot learning capability, especially in resource-constrained scenarios where access to sufficient data and computational resources may be limited. Furthermore, pathological report writing is a time-consuming process in pathologists' clinical workflow. As such, automatic report generation can significantly streamline workload for pathologists, which also heavily relies on the foundation model's language capabilities. With the benefit of the involvement of pathology reports during pretraining, mSTAR is expected to possess great language capabilities. Therefore, in this study, we assess mSTAR's language abilities from three aspects, that is, zero-shot slide classification, zero-shot retrieval and report generation.

Zero-shot's capability always relies on a well-aligned vision-language space. Therefore, we use vision-language foundation models as baselines, i.e., PLIP and CONCH. To produce slide-level predictions for patch extractors, following the setup of CONCH~\cite{lu2024visual}, MI-Zero~\cite{lu2023visual} was adopted through top-K patches voting based on patch similarities to class prototypes or reports embedded by the pretrained text encoders (\textbf{Fig.}~\ref{fig:vision-language}a).
\\
\textbf{Zero-shot Slide Classification}. In this study, we assess every FM on 6 slide classification tasks independent from TCGA data, CAMELYON, PANDA, UBC-OCEAN, BCNB-ER, BCNB-PR and BCNB-HER2.

Across 6 tasks, mSTAR outperforms other FMs on half of the tasks and performs best on the overall result (\textbf{Fig.}~\ref{fig:vision-language}b). Specifically, compared to the second-best FM, mSTAR achieves clear enhancement in these tasks by +3.9\% on average ($P<0.001$) with a significant difference. In particular, on CAMELYON, a remarkable rise of +10.4\% ($P<0.001$) is observed compared to CONCH, the second-best FM. Furthermore, we see performance enhancement over the second-best FM by +16.2\% ($P<0.001$) on BCNB-ER and +19.9\% ($P<0.001$) on BCNB-PR, respectively.
\\
\textbf{Zero-shot Slide Retrieval}. The capability of zero-shot whole-slide retrieval can assist pathologists in seeking similar cases for reference, effectively enhancing diagnostic precision and consistency as well as reducing the workload for pathologists.

In this study, we explore two settings: Image2Text refers to providing an image for the model to find the corresponding report, while Text2Image does the reverse. Although the source of held-out data is the same as pretraining data, the data itself is totally held out from pretraining data. We presented results on held-out data for reference only to be compared with zero-shot's results on external cohorts.
The results (\textbf{Fig.}~\ref{fig:vision-language}c and \textbf{Extended Data} Table~\ref{tab:retrieval}) demonstrate that mSTAR has a clear and significant advantage in this dataset, since mSTAR successfully aligned the vision and language spaces during the pre-training stage. To investigate whether mSTAR can exhibit the advantage on external data, we collected a dataset spanning breast and lung cancers from collaborative hospitals, comprising 500 cases of WSI-Report pairs. Despite performance decreases on the external cohort, results (\textbf{Fig.}~\ref{fig:vision-language}c and \textbf{Extended Data} Table~\ref{tab:retrieval}) demonstrate that mSTAR still performs the best among all vision-language FMs, with +9.4\% of Recall@50 on Text2Image and +3.6\% of Recall@50 on Image2Text.
\\
\\
\begin{figure*}
    \centering
    \includegraphics[scale=0.44]{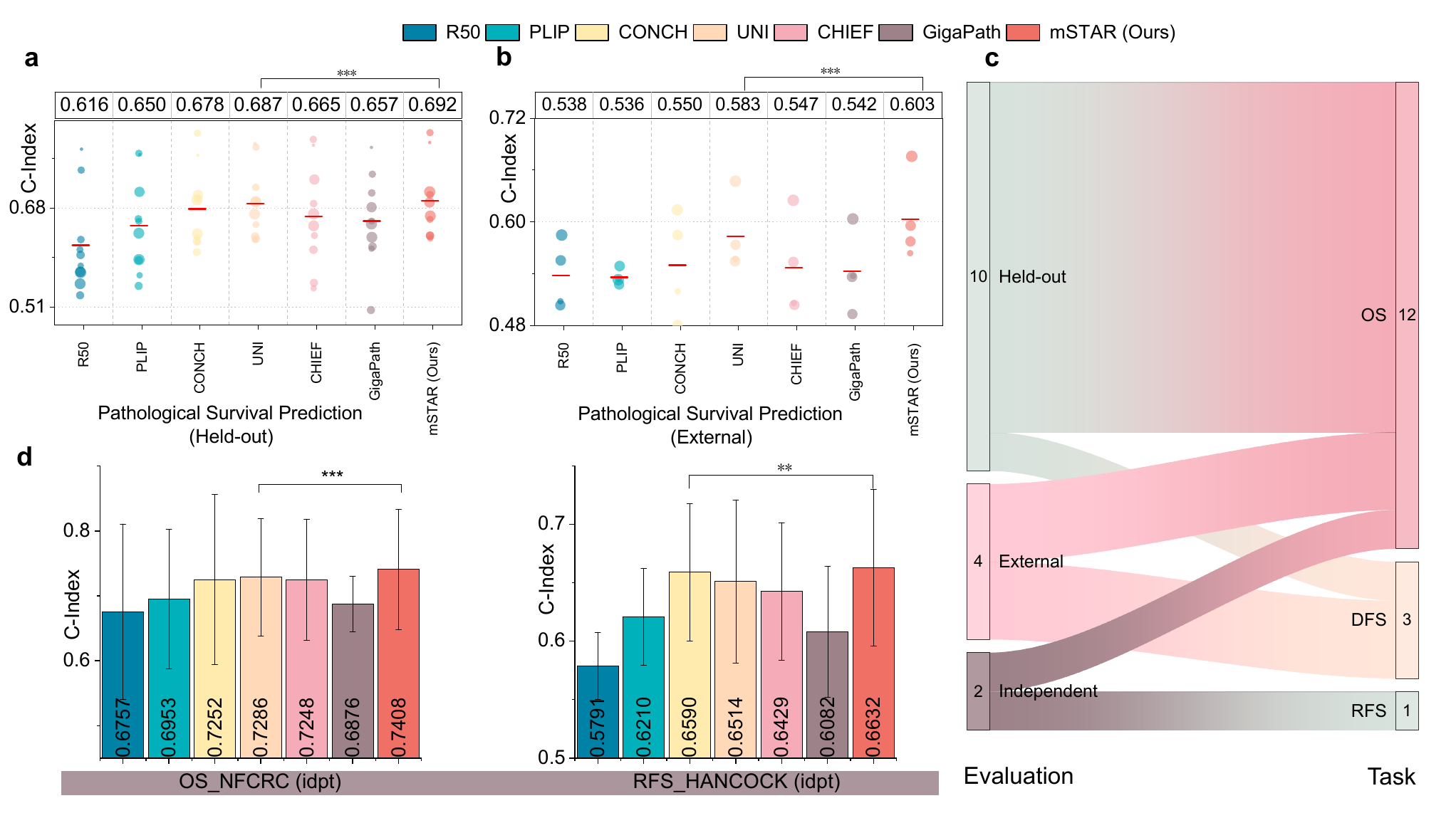}
    \caption{\textbf{Performance of Survival Prediction} on 16 datasets. \textbf{a}, Comparison of C-Index between mSTAR and compared methods on 9 held-out datasets. \textbf{b}, Comparison of C-Index between mSTAR and compared methods on 4 external datasets. The {\color{red}{red}} lines and the values reported at the top of figures \textbf{a} and \textbf{b} refer to the averaged performance across datasets. Each point represents a dataset, with the size of the point indicating the standard deviation. \textbf{c}, Task distribution of various survival endpoints for different evaluation. \textbf{d}, The performance (C-Index and 95\% CI) on independent cohorts. `out' refers to the partitions held out from pretraining data. `idpt' means independent datasets with a data source that differs from the pretraining data. `ext' represents external datasets where data originates from a source distinct from the training data used for fine-tuning and is used solely for testing, without any training involved. \textit{P-value} for every group of experiments is given through one-sided Wilcoxon signed-rank test between mSTAR and the second-best FM. * represents $P<0.05$, ** means $P<0.01$ and *** indicates $P<0.001$. Detailed performances of every dataset are presented in \textbf{Extended Data} Tab.~\ref{tab:survival}.}
    \label{fig:surv}
\end{figure*}
\textbf{Report Generation}. Automated generation of pathology reports has enormous potential in simplifying the report-writing process and reducing the workload burden on pathologists. To assess mSTAR's capability of report generation, we collected one pan-cancer TCGA dataset with 840 cases held out from pretraining data and two external cohorts including Nanfang of lung cancer from 250 patients and ZJ-First of breast cancer from 250 patients. Since pathology reports generally include numerous contents invisible in whole slide images, such as macro descriptions, we first leverage GPT-4o-mini to filter out these irrelevant descriptions. The prompts used for cleaning reports are presented in \textbf{Extended Data} Table~\ref{tab:prompt_repogen}. The detailed process regarding the quality control for reports is presented in Section~\ref{sec:downstream}. In this study, we finetuned HistGen~\cite{guo2024histgen}, a pathology report generation model, based on patch features extracted by different foundation models.

From the quantitative perspective, we evaluated multiple metrics including BLEU, METEOR and ROUGE-L to assess various aspects of the generated text, such as precision of n-grams (contiguous sequences of words), order, alignment, recall, etc. In the held-out TCGA cohort, across different metrics, mSTAR consistently outperformed the second-best approach (\textbf{Fig.}~\ref{fig:vision-language}d and \textbf{Extended Data} Table~\ref{tab:report_gen}). In one external cohort, Nanfang, we observe significant improvements in these three metrics compared to the second-best FM: +6.91\% of BLEU\_1, +1.17\% of METEOR and +4.61\% of ROUGE\_L. This indicates mSTAR has a better generalizable ability for report generation, instead of just memorizing the contents of reports. In another external cohort, ZJ-First, mSTAR demonstrates increases in BLEU metric, indicating mSTAR excels at generating precise long sentences. For METEOR, mSTAR achieves a comparable performance with the second-best FM, while a performance decline of 1.41\% is present in ROUGE\_L, which indicates the generated texts of mSTAR are less fluent.

We continued to qualitatively evaluate the quality of generated reports. The case studies for every cohort are presented in \textbf{Extended Data} Fig.~\ref{fig:cases_report_gen}. The texts highlighted in red are matched with the ground-truth report, while the ones highlighted in blue contradict the true report.

First, we investigate how mSTAR and the competitive FMs perform on held-out datasets. For the case (a) and (b) in \textbf{Extended Data} Fig.~\ref{fig:cases_report_gen}, the texts generated by R50 are almost unrelated to the ground truth and always the same. The reason probably is the pretraining materials are not specific to the pathology domain. The diversity of generated texts from PLIP, CONCH and CHIEF is better than that of R50. For example, PLIP, CONCH and CHIEF can identify the histologic type and simple margin information for case (a), despite missing more details about diagnosis. However, CONCH and CHIEF showcase poor relatedness and the generated texts from PLIP are too short for case (b). UNI and GigaPath are aware of more content types that need to be generated, although the specific predictions are always inaccurate. In other words, the generated texts from UNI and GigaPath contain a lot of hallucinations contradicting to the ground-truth report, such as case (b). mSTAR is able to identify the necessary content and make more accurate predictions cautiously, leading to fewer hallucinations. However, it still fails to count, such as the number of lymph nodes. Then, when examining the external cohorts closely, they demonstrate the same characteristics. This indicates that mSTAR has the generalizable capability of report generation, instead of just memorizing the template reports. 

Given that the current generation capabilities are not perfect, mSTAR excels compared to other models overall. This can be attributed to two main factors. First, these foundational models are encoder-based and do not incorporate decoders during pretraining, resulting in a significant distribution gap between encoded features and generated texts. Second, the effectiveness of existing report generation methods that are fine-tuned on foundational features remains restricted.
\subsection{Survival Prediction}
Prognostic analysis is an intricate clinical endeavor, which can inform clinical guidelines and practices, helping healthcare providers make evidence-based decisions regarding patient care. It is so complicated that it always necessitates a thorough analysis from a multitude of facets. In this regard, multimodal data has proven instrumental in enabling more comprehensive prognostic assessments~\cite{chen2021multimodal,xu2023multimodal,zhou2023cross,zhangprototypical}. Therefore, it is crucial to explore the role of multimodal knowledge within the broader whole-slide context in enhancing prognostic estimation. In this study, we assessed 3 prognostic tasks, Overall Survival (OS), Disease-Free Survival (DFS) and Recurrence-Free Survival (RFS) on top of pathological tissue slides. To this end, we collected 10 held-out cohorts covering 10 cancer types from TCGA, and 4 external and 2 independent cohorts for generalizable validation from public databases and collaborative medical institutions. The distribution relationship between tasks and cohorts can be seen in \textbf{Fig.}~\ref{fig:surv}c. The distribution of samples for every cohort is presented in \textbf{Extended Data} Table~\ref{tab:all_ds}.

First, we investigate the performance of held-out cohorts over 10 datasets, which demonstrates a slight improvement of +0.5\% ($P<0.001$) overall, compared to UNI, the second-best FM. For consistency of performance increases, mSTAR performed best compared to other foundation models, achieving the the top performance on 6 out of 10 datasets. However, UNI, ranking at the second place, performs the best on 2 out of 10 datasets.

Although the improvement in the held-out cohort is not promising, mSTAR demonstrates strong generalization, achieving an average increase of +2\% ($P<0.001$) on 4 external cohorts, especially on OS\_ZJ1 of breast cancer for overall survival with +2.88\% increase ($P<0.001$). Across 4 external cohorts and 2 independent cohorts, mSTAR also shows consistent superiority, which indicates a great generalizable ability of survival prediction.

To demonstrate patients' stratification performance, we examined the Kaplan-Meier curves characterized by mSTAR for every task (\textbf{Extended Data} Fig.~\ref{fig:km_curve}), where 12 out of 14 tasks showcased the statistical difference between low-risk and high-risk groups, via Logrank Test~\cite{mantel1966evaluation}.
\subsection{Multimodal Fusion}
\begin{figure*}
    \centering
    \includegraphics[scale=0.4]{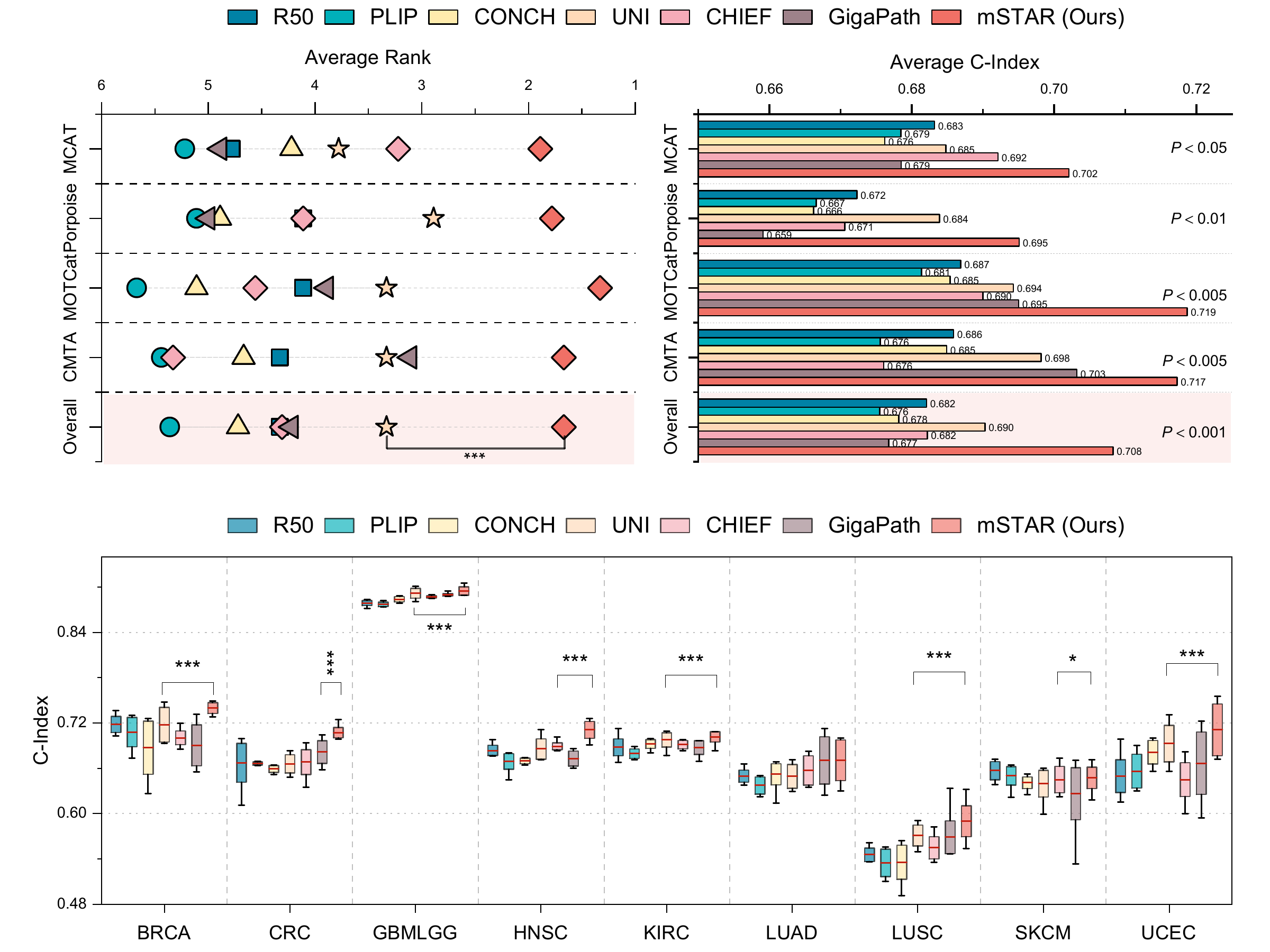}
    \caption{\textbf{Multimodal Fusion Performance of Overall Survival Prediction on Pathological Slides and Gene Expression Data}. The patch extractors of all foundation models are evaluated with different multimodal fusion models (MCAT, Porpoise, MOTCat and CMTA), trained from scratch across 9 TCGA held-out datasets. \textbf{a}, Performance of Ranking on 9 datasets of each FM on every multimodal fusion models and ``Overall'' that refers to the average results among these multimodal fusion methods. \textbf{b}, The average C-Index on 9 datasets. \textbf{c}, Performance (C-Index and 95\% CI) on each dataset. \textit{P-value} is given through one-sided Wilcoxon signed-rank test between mSTAR and the second-best FM. The colors of legends are shared across all sub-figures. *** indicates $P<0.001$. Detailed performances of every dataset are presented in \textbf{Extended Data} Tab.~\ref{tab:fusion_box}-\ref{tab:fusion_cmta}.}
    \label{fig:fusion}
\end{figure*}
Multimodal data typically provides a more comprehensive understanding of cancer, excelling in various clinical applications, such as treatment response prediction for neoadjuvant chemotherapy~\cite{yu20232mo} and prognostic analysis~\cite{chen2022pan}. However, multimodal data integration often suffers from the heterogeneity of different modalities, leading to limited performance. As a result, whether pathological features from foundation models can be well aligned to other modalities plays a crucial role in multimodal analysis. With the benefit of multimodal pretraining, they can align with each other by contrastive learning, thereby potentially alleviating inter-modal heterogeneity. Therefore, in this study, we examine whether mSTAR facilitates multimodal fusion by assessing multimodal overall survival prediction tasks.

To validate this, we replaced pathological features with ones extracted by various extractors in existing multimodal fusion models for 9 cancer survival prediction tasks held out from pretraining data, to observe the differences that would arise. Specifically, to reduce biases caused by multimodal integration approaches, 4 recent multimodal fusion models were employed in this study to make the multimodal slide-level prediction, including MCAT~\cite{chen2021multimodal}, Porpoise~\cite{chen2022pan}, MOTCat~\cite{xu2023multimodal} and CMTA~\cite{zhou2023cross}.

On the whole, mSTAR has clearly outperformed other SOTA methods by a wide margin. Considering average rank, mSTAR ranked between 1.22 and 1.67 among various fusion models and the overall rank is 1.47, which left the second-best approach UNI far behind (\textbf{Fig.}~\ref{fig:fusion}a) ranking at 2.68 on average. For average C-Index (\textbf{Fig.}~\ref{fig:fusion}b), mSTAR achieved consistent and notable enhancement in multimodal fusion with a significant difference, with average performance increases of +1.8\% ($P<0.001$). Among different multimodal fusion models, mSTAR outperformed the second-best FM by +1\% ($P=0.02$) for MCAT, +1.1\% ($P<0.01$) for Porpoise, +2.4\% ($P<0.005$) for MOTCat and +1.4\% ($P<0.005$) for CMTA.

Across various datasets, mSTAR surpassed the second-best FM on 8 out of 9 datasets (\textbf{Fig.}~\ref{fig:fusion}b and \textbf{Extended Data} Tab.~\ref{tab:fusion_box}). Among them, +2.22\% on BRCA ($P<0.001$), +2.55\% on CRC ($P<0.001$), +2.17\% on HNSC ($P<0.001$), +1.89\% on LUSC ($P<0.001$), +1.83\% on UCEC ($P<0.001$) are achieved with statistically significant differences. Specifically, based on MCAT, mSTAR surpassed other SOTA approaches on 5 out of 9 tasks (\textbf{Extended Data} Table~\ref{tab:fusion_mcat}), especially on BRCA (+1.9\%, $P<0.001$). mSTAR with Porpoise demonstrated superior performance in the majority of tasks, topping 6 out of 9 datasets (\textbf{Extended Data} Table~\ref{tab:fusion_porpoise}), which increased the second-best model by up to +3.8\% ($P<0.001$). In the case of MOTCat, mSTAR excelled in 6 out of 9 tasks (\textbf{Extended Data} Table~\ref{tab:fusion_motcat}) with performance increases of up to 3.2\% ($P<0.001$). For CMTA, mSTAR delivered the highest performance in 6 of 9 (\textbf{Fig.}~\ref{fig:fusion}f and \textbf{Extended Data} Table~\ref{tab:fusion_cmta}), advancing the second-best one by up to 2.9\% ($P<0.001$).

In a nutshell, the remarkable increases across various datasets and diverse multimodal fusion backbone models vividly demonstrate the tremendous contributions of multimodal knowledge embedded by slide-level multimodal contrastive learning in facilitating multimodal fusion.

\section{Discussion}\label{sec12}
In this study, we delve into how to harness the full potential of three-level multimodal data to advance the performance of the pathology foundation models effectively, by pretraining the model on over 116 million pathological images of 26k modality pairs from 10,275 patients across 32 major cancer types. Additionally, we explored a new whole-slide pretraining paradigm for CPath, which broadened the context of modelling for better performance on slide-level tasks. For systematical evaluation, we established the largest spectrum of oncological benchmark datasets, covering 7 categories of oncological applications comprising 15 types of 97 oncological tasks. With the benefit of the involvement of pathology reports and gene expression data in pretraining, diverse experimental results demonstrated that mSTAR excelled in not only molecular prediction but also pathological tasks frequently presented in pathology reports at the slide level, such as pathological subtyping, mutation prediction and report generation. Furthermore, multimodal pretraining facilitated multimodal fusion tasks due to a well-aligned multimodal space and endowed the model with more generalized zero-shot's capabilities.

In the realm of prior investigations into pathology foundation models, two prominent categories have emerged: vision-only models~\cite{chen2024towards, vorontsov2023virchow, xu2024whole} and vision-language models~\cite{lu2024visual,huang2023visual}. However, these approaches fail to tap into a vast wealth of information inherent in macroscopic-level pathology reports written by experts and molecular-level gene expression profiles. Pathology reports usually provide authentic expert knowledge in line with the clinical practice, while gene expression profiles serve as robust indicators of oncology status for clinical applications in diagnosis~\cite{hong2020rna} and prognosis~\cite{beer2002gene}. As shown in Extended Data Table~\ref{tab:abla_modal}, the involvement of pathology reports and gene expression data can bring extra performance gains. The superiority in molecular prediction and report-related oncological validates modality scalability in pathology foundation models, which potentially provides a guiding conclusion: pathology foundation models can benefit from a more diverse range of modalities.
Simultaneously, we notice that the majority of existing FMs primarily focus on patch/ROI-level models and short texts, in which restricted contexts hinder their practical performance in slide-level oncological applications. It is worth noting that, unlike CONCH supervised by generative loss, mSTAR without involving generative components in pretraining, still demonstrated encouraging performance in report generation with producing more comprehensive and coherent texts. 

Recently, beyond working on small patches/ROIs, we noticed that some studies~\cite{gigapath,chief} attempted to work on slide-level foundation models, which pretrained the model on patch features. However, the pretrained performance significantly depends on the quality of patch features, leading to under-performing results compared to mSTAR. In other words, their performance would be limited by the patch extractor. We believe that end-to-end pretraining is a promising solution in the future, while its current implementation is hindered by hardware limitations. Therefore, mSTAR bridges this gap through self-taught training to seamlessly transfer the knowledge captured by the slide aggregator into the patch extractor.

Distinct from previous researches, our study provides the uniqueness in three folds. First, our findings showcase the remarkable power of leveraging multimodal data, especially in enhancing multimodal capabilities. This validates the scalability of modalities, providing the guiding principle for building pathology foundation models. Second, we found a unified way to bridge the gap between slide-level and patch-level pretraining, bringing us closer to achieving end-to-end pretraining on raw slide data. We believe this innovative unified paradigm will revolutionize the workflow of pretraining for CPath. Moreover, this paradigm allows the injection of multimodal knowledge into pathology foundation models in an appropriate manner, which may hold the potential to harness more modalities to construct a stronger foundation model for CPath. Third, we established the widest range of oncological benchmark spanning 7 categories of 15 types of 97 oncological tasks.

Although preliminary results are encouraging, this study still has several limitations. First, the challenge of collecting paired multimodal data naturally limits the scale of pretraining data, compared to previous works of pathology foundation models. By expanding the scale of multimodal data for pretraining, we can expect to unlock further potential for enhancing various abilities, such as multimodal capabilities. Second, we still potentially have a long way to go before achieving the true end-to-end foundation model. Before that, mSTAR will serve as an alternative solution to seamlessly bridge slide-level and patch-level pretraining. However, there are still several challenges to be further explored, such as the appropriate way to propagate the pretrained knowledge embedded in the slide aggregator and the architectural design of slide aggregator. In mSTAR, due to a large number of patches of a WSI that would lead to extremely high computational costs, we selected TransMIL with linear time complexity as the slide aggregator. However, the increase in training speed comes at the expense of sacrificing a portion of the performance. Fortunately, a multitude of innovative architectures for handling long sequences are emerging, such as Mamba~\cite{gu2023mamba}, LongNet~\cite{ding2023longnet}, etc, which we explore in concurrent work~\cite{yang2024mambamil}. We believe that these new architectures will undoubtedly create new avenues in exploring more efficient and powerful pretraining paradigms for CPath. Third, due to the inherent challenge of gathering rare cancer data, there is still a need for further assessment of zero-shot's performance on real rare cancer cases, although zero-shot's performance can reflect the performance under the situation of limited data to some extent. In the future, we plan to incorporate more multimodal data into pretraining, such as multi-omics data, explore new efficient pretraining architecture, and keep moving forward in the collection of rare cancer data.
\bibliographystyle{unsrt}
\bibliography{sn-bibliography}

\section{Methods}\label{sec13}
\subsection{Pretraining Dataset Curation}
Data used for pretraining in this study were totally obtained from a publicly available source, the Cancer Genome Atlas Program (TCGA)~\cite{weinstein2013cancer}, in which we collected 9,640 cases (11,765 slides) of diagnostics formalin-fixed paraffin-embedded (FFPE) H\&E WSIs, 11,108 pathology reports and 10,234 cases of bulk RNA-Seq data across all 32 cancer types of TCGA. After quality control, we curated a dataset with 8,440 WSI-Report pairs, 8,965 WSI-RNA-Seq pairs and 8,764 Report-RNA-Seq pairs, resulting in 26,169 modality pairs, as shown in \textbf{Fig.} ~\ref{fig:overview}c. These data involve over 116 million pathological patch images. Given that numerous downstream tasks were evaluated on TCGA data, we held out some validation and test cases. For 9 cancer datasets comprising over 400 cases, we adopted a split ratio of 7:1:2 for train-validation-test folds. For those cases involving multiple slides, we combined their patches or features into a single case for pretraining at the patient level. This ensured slides belonging to one case were included within the same fold, thereby preventing potential data leakage. We also considered label stratification for survival labels at patient-level, since we primarily evaluated the performance of survival prediction on TCGA data. Note that all cases without survival labels were used for pretraining. Details of data splitting for these 9 cancer datasets are provided in \textbf{Extended Data} Table~\ref{tab:tcga_splits}. After data partitioning, we curated 22,127 modality pairs for contrastive learning, consisting of 7,083 WSI-Report pairs, 7,538 WSI-RNA-Seq pairs and 7,506 Report-RNA-Seq pairs. Among these, there were 7,947 cases with all three modalities for pretraining. For acquisition of high-quality data, we conducted the subsequent pre-processing procedures for each modality.
\\
\\
\textbf{WSI Pre-processing.} To conduct slide- (or patient-) level tasks on WSIs, our processing pipeline involved tissue segmentation, patching, and feature extraction (for pretraining aggregators and evaluation). For tissue segmentation, we employed the CLAM library~\cite{clam}, which performed binary thresholding on the saturation channel of a downsampled RGB slide, converted to the hue-saturation-value (HSV) color space. The resulting segmentation mask was obtained by filtering the contours based on their area. The hyperparameters of segmentation are released on our codebase. Furthermore, slides that were corrupted and those containing a small proportion of tissue region were excluded from this study. As a result, we acquired 9,608 cases of 11,727 slides for pretraining and evaluation.

To adhere to established practices of previous works~\cite{clam,chen2024towards,lu2024visual}, we partitioned the segmented tissue regions into 256 $\times$ 256 pixels patches at 20$\times$-equivalent magnification without overlaps and then resized all patches to 224 $\times$ 224 pixels for feature extraction. Using pretrained patch extractors that were kept frozen, we pre-extracted embeddings for each patch and stored them for subsequent evaluation purposes.
\\
\\
\textbf{Report Pre-processing.} For pathology reports, we curated open-source texts from TCGA and converted them from their original PDF format to editable text format via Amazon Web Services (AWS) Optical Character Recognition (OCR) tools, resulting in 9,523 Reports. For quality control, we curated these reports via the powerful language tool, GPT-4, with appropriate prompts provided in \textbf{Extended Data} Table~\ref{tab:prompt_gpt4}, and re-checked them manually to ensure the unchanged original intent. The statistical distribution of word counts for reports is demonstrated in \textbf{Fig.}~\ref{fig:overview}e, in which the majority of cases have word counts below 500.
\\
\\
\textbf{RNA-Seq Pre-processing.} We accessed RNA-Seq data of TCGA from cBioportal database, which were preprocessed and normalized using RSEM~\cite{li2011rsem}. An inherent difficulty in gene expression modelling arises from the variations in absolute magnitudes observed across different sequencing protocols~\cite{sarkar2021separating}. Therefore, we further applied a common preprocessing technique log1p transformation~\cite{haque2017practical} for gene expression values. Following previous works~\cite{yang2022scbert}, Gene2Vec~\cite{du2019gene2vec} contributed to better representing the gene names by enforcing that words with similar meanings are assigned similar representations. Therefore, we retained genes present in the Gene2Vec vocabulary. In the end, we obtained 9,890 cases RNA-Seq data, each consisting of genes with a length of 17,425.
\subsection{Pretraining Framework}
\label{sec:methods_pretrain}
To utilize multimodal knowledge at the whole-slide context for enhancing the pathology foundation model, we propose a whole-slide pretraining paradigm consisting of two-stage pretraining, as shown in \textbf{Fig.}~\ref{fig:framework}. In the first stage, we aim to inject multimodal knowledge into the slide aggregator by contrastive learning, including inter-modality contrastive learning (following CLIP~\cite{radford2021learning}) and inter-cancer contrastive learning. In the second stage, to seamlessly propagate multimodal knowledge at the slide-level context into the patch extractor, we leverage the slide aggregator pretrained in the first stage, serving as a ``Teacher'' model, to supervise the pretraining of the patch extractor, termed Self-Taught training. In this way, multimodal knowledge of the whole-slide context can be injected into the pathology FM.
\\
\textbf{Stage 1 - Pretrain Slide Aggregator.} In this stage, we aim to pretrain a slide aggregator that learns multimodal knowledge by contrastive learning with other modalities. Note that the pretrained slide aggregator plays a role of ``Teacher'' that propagates the learned knowledge into the patch extractor at the next stage. These modules to be trained are highlighted in red boxes in the \textbf{Fig.}~\ref{fig:framework}a, in which we pretrain a 2-layer TransMIL~\cite{shao2021transmil} as the slide aggregator for WSIs, a Bert-like text encoder (following BioBert-Base-v1.2~\cite{lee2020biobert}) for pathology reports, and a Performer (following scBERT~\cite{yang2022scbert}) for RNA-Seq data.

Given these transformer-like encoders, we need to tokenize raw data of every modality into token embeddings before feeding them into their respective encoders. For pathology, we obtained non-overlapping $224\times 224$ patches as early mentioned, and then for every patch, we used a pretrained patch extractor, UNI~\cite{chen2024towards}, to extract patch features, where a patch feature was regarded as a token embedding for the slide aggregator. After gathering 4,096 patch features for the $i$-th patient's WSIs, $\mathcal{P}_i=\{\mathbf{p}_i^m \}_{m=1}^{M}$, we fed them into the slide aggregator to integrate all patch features and got a 512-dimensional pathological [CLS] token embedding $\mathbf{P}_i$ as the slide-level representation, where $M$ is the number of patches and it was fixed into 4,096. For cases where the number of patches exceeds 4,096, a random selection of 4,096 patches is made, while for cases with fewer than 4,096 patches, padding is applied using the mean value. For those cases where one patient has more than one WSI, we simply concatenated them together. Note that all patch features were transformed into 512-dimensional features by a linear projection before being forwarded into the aggregator. 

For pathology reports, we adopted the text encoder for randomly truncated 512 tokens and outputted the report [CLS] token embedding $\mathbf{T}_i$. For cases where the length of the text is less than 512, the special token '[pad]' was padded. The RNA-Seq data was organized as a set of 2-tuple $(g_i, e_i)$ comprising of the gene name $g_i$ and its expression variable $e_i$. Following previous works~\cite{yang2022scbert,cui2024scgpt}, to assure that genes with potential co-expression get close together, we employed Gene2Vec~\cite{du2019gene2vec} to generate 200-dimensional gene embeddings for each gene name $g_i$. Gene expression can be viewed as the manifestation or presence of each gene, which has been well-documented within a biological system. Therefore, we applied the term-frequency-analysis method used in previous works~\cite{yang2022scbert,cui2024scgpt} to discretize the continuous expression variable $e_i$ through binning technique. Subsequently, the discrete variable was transformed into a 200-dimensional embedding, which was then integrated into the final gene token embedding $\mathbf{g}_i$ by addition. Through forwarding the gene encoder, we can get the gene [CLS] token embedding $\mathbf{G}_i$. It is worth noting that encoder outputs from report and gene modalities were transformed into 512-dimensional features by a linear projection for contrastive learning.

To optimize the model through pretraining, we incorporate two objectives including inter-modality contrastive learning and  inter-cancer contrastive learning. In the case of inter-modality contrastive learning, given the [CLS] representation of each modality, every two modalities can be paired together, which finally yielded three combinations: WSI-report $(\mathbf{P}_i, \mathbf{T}_i)$, WSI-gene $(\mathbf{P}_i, \mathbf{G}_i)$ and report-gene $(\mathbf{T}_i, \mathbf{G}_i)$. During pretraining between every modality pairs, a mini-batch consisted of $N$ samples, e.g., $\{(\mathbf{P}_i, \mathbf{T}_i)\}_{i=1}^{N}$ for WSI-report. Contrastive learning imposes a higher similarity in modality pairs from the same sample. Take WSI-report pairs as an example, and the loss function can be formulated as
\begin{equation}
\begin{aligned}
    \mathcal{L}_{P-T} = &-\frac{1}{2N}\sum_{i=1}^N\log \frac{\exp{( \mathbf{P}_i^\top\mathbf{T}_i / \tau)}}{\sum_{j=1}^{N}\exp{( \mathbf{P}_i^\top\mathbf{T}_j / \tau)}} \\
    &-\frac{1}{2N}\sum_{j=1}^N\log \frac{\exp{( \mathbf{T}_j^\top\mathbf{P}_j / \tau)}}{\sum_{i=1}^{N}\exp{( \mathbf{T}_j^\top\mathbf{P}_i / \tau)}}
\end{aligned}
\end{equation}
where $\tau$ is a scale factor of the contrastive loss and it was set by default following CLIP~\cite{radford2021learning}. Similarly, we can get $\mathcal{L}_{P-G}$ and $\mathcal{L}_{T-G}$ and finally combine them by addition.

To alleviate the heterogeneity of various cancer types, we utilized inherent cancer labels available in TCGA for the inter-cancer pretraining objective. Specifically, [CLS] tokens of available modalities (regardless of whether they involved two or three modalities) would be concatenated into a single anchor representation $\boldsymbol{a}_i$. Furthermore, positive and negative samples were obtained within the mini-batch, and they were from the same cancer and different cancers, respectively. Similarly, they were constructed in the same way by concatenating the [CLS] tokens from available modalities, leading to $\boldsymbol{a}^+$ and $\boldsymbol{a}^-$ for positive and negative samples, respectively. Subsequently, we enforced a triplet loss $\mathcal{L}_{triplet}$ for them to bring the samples of the same cancer closer than that of the negative sample:
\begin{equation}
    \mathcal{L}_{triplet} = \frac{1}{N} \sum_{i=1}^N \max{(d(\boldsymbol{a}_i, \boldsymbol{a}^+)-d(\boldsymbol{a}_i, \boldsymbol{a}^-)+\epsilon, 0)}
\end{equation}
where $\boldsymbol{a}^+$ and $\boldsymbol{a}^-$ represent the farthest positive samples and nearest negative samples within a mini-batch, respectively, following the hard sample mining technique~\cite{hermans2017defense}. Here we used $l_2$ distance for function $d(\cdot)$ and $\epsilon$ is the margin which was set 0.3 based on smoother stability of loss degradation in the training set. Through these two pretraining objectives, as a result, we can get a well-trained slide aggregator that absorbed multimodal knowledge, which would be the 'Teacher' for the patch extractor at the next stage.
\\
\textbf{Stage 2 - Pretrain Patch Extractor.} Upon finishing the first stage of pretraining, we can obtain a slide aggregator incorporating multimodal knowledge by being pretrained with multimodal data. In this stage, we leverage the pretrained slide aggregator as ``Teacher'' to seamlessly propagate multimodal knowledge into pathological patch extractor (ViT-L~\cite{dosovitskiy2020image}), as shown in \textbf{Fig.}~\ref{fig:framework}b, which is termed Self-Taught training. Specifically, for each WSI, we gathered their patch features $\mathcal{P}_i=\{\mathbf{p}_i^m \}_{m=1}^{M}$ of the $i$-th WSI and fed them into the aggregator pretrained in the previous stage, where $M$ refers to the number of patches of this WSI. Following the setting in the previous stage, $M$ was fixed as 4,096. In this way, every patch can be re-embedded into new features $\hat{\mathcal{P}}_i=\{\hat{\mathbf{p}}_i^m \}_{m=1}^{M}$ incorporating multimodal knowledge. With these re-embedded features as the objective guidance, we can pretrain a patch extractor by enforcing the extracted patch feature to get as close as possible to the ones re-embedded by the well-trained aggregator. To achieve this, for each patch, we can query its corresponding re-embedded feature $\hat{\mathbf{p}}_i^m$ encoded by the aggregator and further tuned the extractor with a loss function that minimizes the discrepancy between patch features encoded by the patch extractor and the corresponding re-embedded features incorporating multimodal knowledge:
\begin{equation}
    \min \sum_m^{M} ||f(\mathbf{p}_i^m) -  \hat{\mathbf{p}}_i^m||_1
\end{equation}
where $f(\cdot)$ is a linear projection for adjusting the dimension of features and it transformed them into 512-dimensional features. Additionally, to avoid the catastrophic forgetting problem, a siamese structure is employed for the patch extractor consisting of two identical branches, where the parameters of one branch are updated using gradient descent, while the parameters of the other branch are updated using an Exponential Moving Average (EMA) of the parameters from the previous branch, without any gradient updates. Afterward, we enforced a similarity constraint between the patch features $\mathbf{p}_i^m$ extracted by the branch with gradient updates and those $\overline{\mathbf{p}}_i^m$ embedded by the branch with EMA updates. In the end, we combined two objectives into a loss function for pretraining the patch extractor:
\begin{equation}
    \min \sum_m^{M_i} \lambda \cdot ||f(\mathbf{p}_i^m) -  \hat{\mathbf{p}}_i^m||_1 + (1-\lambda) \cdot ||\mathbf{p}_i^m -  \overline{\mathbf{p}}_i^m||_1
\end{equation}
where $\lambda$ is a balancing coefficient and it was set 0.6 based on  smoother stability of loss degradation. By doing this, the patch extractor was enhanced by multimodal knowledge at the whole-slide context.

\subsection{Downstream Tasks}
\label{sec:downstream}
\textbf{Comparisons and Baselines.} To investigate the benefit of enhancing the patch extractor by incorporating multimodal knowledge at the slide level, we compared mSTAR against one general baseline and three SOTA pretrained extractors commonly used in the CPath community: (1) ResNet50~\cite{he2016identity} pretrained on ImageNet-1K~\cite{deng2009imagenet}, a commonly used baseline in many slide-level tasks~\cite{shao2021transmil,xu2023multimodal}. (2) PLIP~\cite{huang2023visual}, a vision-language (V-L) architecture (CLIP~\cite{radford2021learning}) pretrained on OpenPath consisting of over 200k pathological patch-caption pairs. (3) CONCH~\cite{lu2024visual}, a V-L CoCa~\cite{yu2022coca} framework with an additional generative loss pretrained on over 1.17 million pathological patch–caption pairs. (4) UNI~\cite{chen2024towards}, a pure vision patch extractor pretrained on more than 100 million patches from over 100k WSIs, and (5) CHIEF~\cite{chief} as well as (6) GigaPath~\cite{gigapath}, 2 slide-level vision-only pathology foundation models pretrained on 60,530 and 171,189 slides. Through pre-extracted patch features via these encoders, we can get 1024-dimensional (1024-d) embeddings for ResNet50, UNI, and mSTAR, 512-d embeddings for PLIP and CONCH, 768-d embeddings for CHIEF and 1536-d embeddings for GigaPath.
\\
\\
\textbf{Models for Downstream Tasks}

\textbf{WSI Classification and Survival Prediction}. For slide-level prediction including classification and survival prediction, we follow the conventional two-stage MIL paradigm comprising pre-extraction of patch features as instances and the training of a MIL aggregator that integrates patch features (or instances) into a single slide-level (or bag) feature. The aggregator took all patch features of a WSI as an input and mapped them into a hidden embedding as a single slide-level representation. Subsequently, the slide-level representation was passed through a fully connected classifier head, resulting in logits. Lastly, based on logits, we performed two types of slide-level tasks including classification supervised by cross-entropy loss with slide labels, and survival prediction (an ordinal regression task) supervised by NLL loss~\cite{zadeh2020bias} with survival labels (event time in month), ranging from various diagnosis and prognosis tasks. Unless otherwise specified, we obtained slide-level predictions by training the widely used attention-based multiple-instance learning (ABMIL)~\cite{ilse2018attention}, a MIL aggregator that integrates all patch features of a WSI into the slide-level representation according to attention scores. For CHIEF and GigaPath, we fully follow the design in their original text that their pretrained patch extractors paired with the corresponding pretrained aggregator were employed. In particular, for patient-level tasks, such as survival prediction, we concatenate features of all slides belonging to a single patient as one case for the patient-level prediction.

We used the same hyper-parameters set for mSTAR and the competing FMs, in which the hidden dimensions are 512 and dropout keeps $p=0.25$ after each intermediate layer in the network for regularization. We trained each model for 30 epochs on the training split by an Adam optimizer of the learning rate of $2\times 10^{-4}$ along with a cosine learning rate scheduler. The full set of hyperparameters is summarized in \textbf{Extended Data} Table ~\ref{tab:mil_cls}.

\textbf{Multimodal Fusion}. In the experiments of multimodal fusion, we employed 4 existing SOTA multimodal integration models, MCAT~\cite{chen2021multimodal}, Porpoise~\cite{chen2022pan}, CMTA~\cite{zhou2023cross} and MOTCat~\cite{xu2023multimodal}. It is worth highlighting that the training and evaluation of multimodal datasets held out from TCGA followed the same splits as that of vision-only models, and we simply discarded those cases without paired RNA-Seq data. For the aforementioned four existing multimodal integration models, we followed their default hyperparameters for these models, and detailed hyperparameters for each model are presented in \textbf{Extended Data} Table ~\ref{tab:mm_porpoise}-\ref{tab:mm_cmta}. For Porpoise, the input length of RNA-Seq varies across different cancer datasets in TCGA and the hidden dimension for RNA-Seq is fixed as 25, while the hidden dimension of pathological features was first transformed into 512 and then 256. Both modality branches adopted the dropout technique with $p=0.1$. Lastly, features from two modalities were fused into a 256-dimensional slide-level feature. For MCAT and MOTCat, the hidden dimension of features was 256 for both modalities and dropout was 0.25 for regularization. Subsequently, features from two modalities were concatenated and integrated into a 256-dimensional slide-level representation. Similarly, CMTA followed the same hyperparameters except the hidden dimension of RNA-Seq which first became 1024 and then 256. For RNA-Seq data of MCAT, CMTA and MOTCat, embeddings were defined based on 6 functional categories according to ~\cite{liberzon2015molecular} provided in MCAT by default, including 1) Tumor Supression, 2) Oncogenesis, 3) Protein Kinases, 4) Cellular Differentiation, 5) Transcription, and 6) Cytokines and Growth. More training hyperparameters are provided in \textbf{Extended Data} Table~\ref{tab:mm_train}.

\textbf{Zero-shot Slide Classification and Retrieval}. We considered the pretrained model as a good zero-shot learner, and employed non-parametric MI-Zero~\cite{lu2023visual} that does not rely on parametric training for these tasks, a well-established zero-shot approach for pathology slides. Given that the zero-shot's capability heavily relies on the well-aligned modality spaces, we only compared against those approaches that are equipped with the text encoder by utilizing the pretrained text encoder as a good classification head, including PLIP and CONCH. The ensembling prompt of templates was used as the textual classification, which was utilized to compute the cosine similarity score with every patch feature. In the end, MI-Zero made the slide-level decision for every slide in the test set based on the majority voting of top-K scores. 

\textbf{Pathological Report Generation.} To do this, we finetune the specific model of report generation, our prior work HistGen~\cite{guo2024histgen}. Given patients' pathology features from WSIs of each FM, HistGen is able to produce a sequence of words. Specifically, given extracted pathological features from the foundation model, the encoder-decoder architecture of HistGen would encode them into the latent features for report decoding. Subsequently, these features are utilized by the text decoder to generate the report. The quality of the generated report is directly influenced by the quality of the pathological features encoded by each FM. For all optimization hyperparameters, refer to \textbf{Extended Data} Table ~\ref{tab:param_histgen}.
\\
\\
\textbf{Evaluation}
\label{sec:eval}

To systematically evaluate mSTAR's capabilities, as shown in Fig.~\ref{fig:overview}f, following the previous work~\cite{chief}, we adopted four evaluation strategies as follows: 

\textbf{`Held-out' (out)} represents the downstream dataset held out from pretraining data to avoid data contamination for evaluation. The training data included in pretraining data was used for training task-specific models, which were then used for inference on validation and test sets (i.e., held-out cohorts) that were held out from the pretraining data. 

\textbf{`Independent' (idpt)} underscores that the source of dataset is independent from that of the pretraining data. For these datasets, we always either label-stratified these datasets into 7:1:2 train-validation-test folds or employed 5-fold cross-validation independently. Note that the difference between \textit{Held-out} and \textit{Independent} lies in whether the data comes from the same source as the pretraining data. 

\textbf{`External' (ext)} is used for testing only and its data source is different from training data (from either held-out or independent cohorts) that was utilized to train task-specific models.

\textbf{`Zero-shot'} means that foundation models (e.g., mSTAR) are directly applied to make slide-level predictions without further training, rather than relying on additional task-specific models.

The details of all evaluation datasets are demonstrated in \textbf{Extended Data} Table~\ref{tab:all_ds}.
\\
\\
\textbf{Datasets}
\\
We present a description of each dataset used for evaluation, including 7 categories of oncological applications, covering 15 types of 97 practical clinical tasks. More details are summarized in \textbf{Extended Data} Table~\ref{tab:all_ds}.
\\
\textbf{BRCA\_PathSubtype~\cite{weinstein2013cancer} for Pathological Subtyping (2 classes).}
\\
The BRCA\_PathSubtype (Breast Invasive Carcinoma) dataset are sourced from TCGA including H\&E diagnostic histopathology WSIs. This dataset encompassed cases of primary IDC (Invasive Ductal Carcinoma) and ILC (Invasive Lobular Carcinoma). After excluding slides with inadequate proportional tumor, a total of 985 slides were gathered, comprising 787 IDC and 198 ILC slides. Following the splits for pretraining, which approximately yielded 7:1:2 train-validation-test folds (656:95:234 slides), we ensure validation and test sets held out from pretraining sources.
\\
\textbf{GBMLGG\_PathSubtype~\cite{weinstein2013cancer} and EBrains\_PathSubtype~\cite{ebrains_data} for Pathological Subtyping (3 classes).}
\\
The GBMLGG\_PathSubtype (Glioblastoma and Brain Lower Grade Glioma) dataset comprises 1276 H\&E diagnostic histopathology WSIs in total, consisting of three classes subtypes: Glioblastoma (GB) with 895 slides, Anaplastic Astrocytoma (AASTR) with 164 slides and Oligodendroglioma (ODG) with 217 slides. Following the splits for pretraining, which approximately yields 7:1:2 train-validation-test folds (839:200:237 slides), we ensure validation and test sets held out from pretraining materials. To evaluation models' generalizable ability, we collected samples of the same subtypes as GBMLGG\_PathSubtype from EBrains~\cite{ebrains_data} database, leading to 732 slides as an external cohort, EBrains\_PathSubtype for pathological subtyping. It consists of 559 slides of Glioblastoma (GB), 89 slides of Anaplastic Astrocytoma (AASTR) and 84 slides of Oligodendroglioma (ODG).
\\
\textbf{HANCOCK\_PathSubtype~\cite{hancock} for Pathological Subtyping (3 classes).}
\\
HANCOCK\_PathSubtype provides a dataset of head\&neck tumors for pathological subtyping of 3 categories: SCC\_Conventional-Keratinizing with 427 slides, SCC\_Basaloid with 144 slides and SCC\_Conventional-NonKeratinizing with 101 slides, resulting in 672 slides totally. We label-stratified the dataset into 7:1:2 train-validation-test splits, yielding 470:68:134 slides.
\\
\textbf{TCGA-NSCLC~\cite{weinstein2013cancer} for Pathological Subtyping (2 classes).}
\\
The TCGA-NSCLC (Non-Small Cell Lung Cancer) dataset comprised NSCLC H\&E diagnostic slides from TCGA, including cases of primary lung adenocarcinoma (LUAD) and lung squamous cell carcinoma (LUSC). After tissue segmentation, a total of 1,053 slides were obtained, consisting of 541 LUAD and 512 LUSC slides. Similarly, we used the same pretraining splits train-validation-test of an approximate ratio 7:1:2 (664:100:289 slides) to avoid data contamination.
\\
\textbf{NFGC\_Lauren and YN3\_Lauren for Lauren Subtyping of Gastric Cancer (3 classes).}
\\
Lauren subtyping is a common classification system for gastric cancer based on morphology, which typically divides tumors into Diffuse-type, Intestinal-type and Mixed-type that indicate different prognostic outcomes and treatment responses. We utilized the TCGA-STAD dataset as an internal cohort to train a model for Lauren classification. Since the data of TCGA-STAD has been used for pretraining, we collected one external gastric cancer cohort (NFGC\_Lauren) of 388 slides from NanFang Hospital (NFH) and another external cohort of 319 slides from the Third Affiliated Hospital of Kunming Medical University in Yunnan (YN3\_Lauren) for testing only. NFGC consists of 159 slides of Diffuse-type, 102 slides of Intestinal-type and 127 slides of Mixed-type. For YN3, there are 143 slides of Diffuse-type, 90 slides of Intestinal-type and 86 slides of Mixed-type.
\\
\textbf{NFGC\_PathSubtype, YN1\_PathSubtype and YN3\_PathSubtype for Pathological Subtyping (3 classes).}
\\
With TCGA-STAD as an internal, we evaluate the ability of pathological subtyping for 3 crucial categories: Tubular Stomach Adenocarcinoma, Signet Ring Cell Carcinoma of the Stomach and Stomach Adenocarcinoma. For external validations, we collected 3 cohorts, NFGC\_PathSubtype, YN1\_PathSubtype and YN3\_PathSubtype, from 3 medical centers including NFH, the First Affiliated Hospital of Kunming Medical University in Yunnan (YN1) and YN3, leading to 385, 254 and 315 slides for testing only. Specifically, NFGC\_PathSubtype of NFH includes 166 slides of Tubular Stomach Adenocarcinoma, 163 slides of Signet Ring Cell Carcinoma of the Stomach and 66 slides of Stomach Adenocarcinoma. YN1\_PathSubtype consists of 59 slides of Signet Ring Cell Carcinoma of the Stomach and 195 slides of Stomach Adenocarcinoma, while YN3\_PathSubtype comprises 82 slides of Signet Ring Cell Carcinoma of the Stomach and 233 slides of Stomach Adenocarcinoma. Note that all data of external cohorts are used for testing only.
\\
\textbf{CAMELYON~\cite{c16_dataset,c17_dataset} for Breast Metastasis Detection (2 classes).}
\\
This dataset comprises 399 slides from the Cancer Metastases in Lymph Nodes Challenge 2016 (CAMELYON16)~\cite{c16_dataset} and 500 slides from the CAMELYON17 challenge~\cite{c17_dataset}, resulting in 899 slides for the breast metastasis detection of two classes (``normal'' v.s. ``metastasis''). After removing a corrupted slide, we obtained a total of 898 WSIs (557 normal, 341 metastasis). For training and evaluation, we employed the label-stratified 7:1:2 train-validation-test splits (629:90:179 slides).
\\
\textbf{NF\_Metastatic, NF\_Metastatic\_Fine, QFS\_Metastatic and QFS\_Metastatic\_Fine for Lung Metastasis Detection (2 classes and 6 classes).}
\\
Lung Metastasis Detection includes two tasks: metastasis detection and its primary site prediction, denoted by `Metastatic' and `Metastatic\_Fine'. We curated NF\_Metastatic dataset (1,198 slides, 705 cases) from NFH, in which `Metastatic' aims to identify if the tumor is metastatic (314 cases) or primary (391 cases). Another dataset NF\_Metastatic\_Fine (705 cases) is also established from NFH, in which `Metastatic\_Fine' is performed to predict the primary site of metastatic cancer. The primary sites include six distinct classes: LUAD (391 cases), breast (55 cases), colon (186 cases), kidney (25 cases), liver (34 cases), and carcinoma of unknown primary (CUP, 14 cases). Both two datasets are label-stratified into 7:1:2 train-validation-test splits (493:70:142 cases). For external cohorts, we incorporated 530 WSIs (430 cases) from Shandong Provincial Qianfoshan Hospital (QFS), leading to QFS\_Metastatic and QFS\_Metastatic\_Fine cohorts for testing only. QFS\_Metastatic dataset included 237 primary cases and 193 metastatic cases, while QFS\_Metastatic\_Fine comprised 237 LUAD cases, 50 breast cases, 96 colon cases, 30 kidney cases, 10 liver cases, and 7 CUP cases.
\\
\textbf{NFGC\_Perineural and YN3\_Perineural for Perineural Invasion Detection in Gastric Cancer.}
\\
The morphological presence of Perineural Invasion (PNI) often indicates a more aggressive tumor and poorer survival rates. As such, it is crucial for prognostic evaluation, treatment decisions, and assessing recurrence risk to detect PNI. To this end, NFGC\_Perineural dataset (396 cases) was collected NFH, consisting of 255 positive and 141 negative cases. As an internal cohort for training and evaluation, the data was divided into training, validation, and test sets in a ratio of 7:1:2 (277:39:80 cases). Furthermore, an additional cohort of 319 cases (112 positive and 207 negative), YN3\_Perineural, was obtained from YN3 for external validation.
\\
\textbf{NFGC\_Vascular and YN3\_Vascular for Vascular Invasion Detection in Gastric Cancer.}
\\
Vascular Invasion in gastric cancer indicates the presence of tumor cells in blood vessels and is linked to poorer prognosis, higher metastasis risk, and increased recurrence rates. To identify it, we used a dataset consisting of 395 cases from NFH, known as the NFGC\_Vascular dataset. This dataset comprises 197 positive cases and 198 negative cases. For model training and evaluation, the data was divided into training, validation, and test sets in a 7:1:2 ratio. Furthermore, we included an external validation set of 319 cases from YN3, which contains 122 positive and 197 negative cases.
\\
\textbf{PANDA~\cite{panda_dataset} for Prostate ISUP grading (6 classes).}
\\
Derived from the PANDA challenge~\cite{panda_dataset}, the ISUP (International Society of Urological Pathology) grading task includes a collection of 10,616 prostate cancer core needle biopsies for prostate cancer evaluation of 6 grades (also known as ``classes''). After tissue segmentation, slides with a low tumor proportion were excluded, which resulted in 10,202 slides. For training and evaluation, we label-stratified PANDA into 7:1:2 train–validation–test folds (7,143:1,019:2,040 slides).
\\
\textbf{HANCOCK-TStage~\cite{hancock} for Pathological T-Staging (4 classes).}
\\
In clinical practice, pathologists will divide patients into different stages according to the severity, which can guide treatment decisions and assess the likelihood of metastasis. To assess this task, we utilize HANCOCK-TStage dataset consisting of 705 patients and divided it into 7:1:2 train–validation–test folds (496:67:128 cases) for validation. To be specific, the dataset includes 259 T1, 256 T2, 123 T3 and 67 T4 cases.
\\
\textbf{18 TCGA Datasets for Mutation Prediction.}
\\
We used the public TCGA data from the studies held out from pretraining materials to evaluate the performance of gene-level mutation prediction. For every study, we involve high-frequent mutated genes and FDA-approved drug-related genes, leading to 17 datasets across 9 held-out studies. The positive rates are presented in \textbf{Extended Data} Table~\ref{tab:pos_rate}. Additionally, the prediction of tumor mutation burden (TMB), a predictive biomarker in solid tumors that is especially important for immunotherapy, was also evaluated in the TCGA-NSCLC study.
\\
\textbf{5 CPTAC~\cite{cptac_data} Datasets for Mutation Prediction.}
\\
With internal cohorts from TCGA, we utilized the data from CPTAC database for external validation on the ability of mutation prediction. The datasets with over 100 cases and mutation rate of at least 30\%, and overlap with internal datasets are included, resulting in BRCA\_PIK3CA (116 cases), BRCA\_TP53 (116 cases), BRCA\_TTN (120 cases), LUAD\_KRAS (175 cases) and LUAD\_EGFR (175 cases). The positive rates are presented in \textbf{Extended Data} Table~\ref{tab:pos_rate}.
\\
\textbf{10 IHC Biomarker Datasets.}
\\
Immunohistochemistry (IHC) typically serves as the biomarker to assess tumor types and differentiation, guide the choice of targeted and immunotherapies, and monitor recurrence in clinical practice. We collected three IHC biomarkers tasks from TCGA: estrogen receptor (ER) with 949 cases, progesterone receptor (PR) with 948 cases, and human epidermal growth factor receptor (HER2) with 646 cases. For training and evaluation, these datasets are divided into train-val-test splits following the pretraining splits to avoid data contamination. With these datasets as internal cohorts, we curated the corresponding tasks from The First Affiliated Hospital of Zhejiang University School of Medicine (ZJ1), leading to IHC\_ZJ1\_ER (1,548 cases),  IHC\_ZJ1\_HER2 (1,344 cases) and IHC\_ZJ1\_PR (1,556 cases) for testing only. As doctors in clinical settings typically annotate the fine-grained labels for ER and HER2, we further assessed their expression levels, resulting in two datasets: IHC\_ZJ1\_ER\_Level and IHC\_ZJ1\_HER2\_Level with 7:1:2 splits (1,083:154:311 cases for ER and 940:134:270 cases for HER2) for training and evaluation. Furthermore, we also evaluate other biomarkers commonly seen in clinical practice: Cytokeratin 5 (CK5, 961 cases) of breast cancer and Cytokeratin 7 (CK7, 419 cases) of lung cancer. For training and evaluation, we divided them with 7:1:2 into train-val-test splits (672:96:193 cases for CK5 and 293:42:84 cases for CK7). The label distribution of these biomarkers can be found in \textbf{Extended Data} Table~\ref{tab:pos_rate_ihc} and~\ref{tab:ihc_level}.
\\
\textbf{BRCA\_MolSubtype~\cite{weinstein2013cancer} and ZJ1\_Breast\_MolSubtype for Molecular Subtyping (4 classes).}
\\
BRCA\_MolSubtype~ is derived from TCGA, consisting of Triple-Negative Breast Cancer (TNBC) (94 cases), HER2 (56 cases), LumA (228 cases) and LumB (127 cases) classes. For training and evaluation, we label-stratified the dataset into train–validation-test cohorts (323:53:129 cases). For external validation, an external cohort was established with 2,045 cases (585 TNBC, 292 HER2, 307 LumA and 861 LumB).
\\
\textbf{GBMLGG\_MolSubtype~\cite{weinstein2013cancer} and EBrains\_MolSubtype~\cite{ebrains_data} for Molecular Subtyping (2 classes).}
\\
GBMLGG\_MolSubtype~ is derived from TCGA for identifying IDH status, consisting of Positive (362 cases) and Negative (190 cases) classes. For training and evaluation, we label-stratified the dataset into train–validation-test cohorts (401:64:87 cases). For external validation, an external cohort was established with 428 cases (361 Positive and 67 Negative).
\\
\textbf{TCGA\_HNSC\_HPV~\cite{weinstein2013cancer} and HANCOCK\_HPV~\cite{hancock} for Molecular Subtyping (2 classes).}
\\
HPV-p16 status is a significant prognostic biomarker regrading different outcomes. To predict HPV-p16 status, we leveraged TCGA\_HNSC\_HPV (405 cases) derived from TCGA as an internal cohort for identifying HPV status, consisting of Positive (41 cases) and Negative (364 cases) classes. For training and evaluation, we label-stratified the dataset into train–validation-test cohorts (284:39:118 cases). For external validation, an external cohort was established with 332 cases (191 Positive and 141 Negative).
\\
\textbf{CRC\_MolSubtype~\cite{weinstein2013cancer}  for Molecular Prediction (4 classes).}
\\
The dataset (492 cases) used in this study is derived from the TCGA CRC (Colon Adenocarcinoma and Rectum Adenocarcinoma) dataset, which includes the Colon Adenocarcinoma (COAD)  and Rectum Adenocarcinoma (READ) datasets. It comprises four consensus molecular subtypes (CMSs): 74 CMS1, 211 CMS2, 68 CMS3 and 139 CMS4. To facilitate training and evaluation, we stratified the dataset based on labels into train-validation-test cohorts with proportions of 325:49:118 cases, respectively.
\\
\textbf{10 TCGA cohorts, 4 external cohorts and 2 independent cohorts for Survival Prediction.}
\\
``In pretraining splits, we employed case- and label-stratified 7:1:2 training-validation-test splits for 9 TCGA cancer datasets of over 400 cases. We evaluated the capability of survival analysis on the same validation and test sets totally excluded from pretraining data. More information about the 9 TCGA cancer datasets were provided in \textbf{Extended Data} Table~\ref{tab:tcga_splits}. We first evaluated 9 Overall Survival (OS) tasks of TCGA across 9 cancer types. With OS\_HNSC as the internal cohort, we further collected OS\_HANCOCK (747 cases) as an external cohort. Given OS\_BRCA as the internal cohort, we curated OS\_ZJ1 (454 cases) from ZJ1 of breast cancer for external validation. Additionally, we further evaluate the DFS task of breast cancer on TCGA, resulting in DFS\_BRCA (878 cases), serving as the internal cohort with the splits (619:84:175 cases). For external validation, DFS\_ZJ1 (454 cases) was curated as an external cohort. Although TCGA-STAD was not held out from pretraining data, we utilized it as a training set to train a model and evaluate it on the curated DFS\_YN1 (260 cases) sourced from YN1 for external validation. For independent cohorts, we curated OS\_NFCRC (294 cases) of colon cancer from NFH for OS prediction using 5-fold cross-validation, and meanwhile we utilized RFS\_HANCOCK (747 cases) for Recurrence-Free Survival (RFS) prediction based on 5-fold cross-validation as well.
\\
\textbf{9 Survival Prediction Datasets for Multimodal Fusion.}
\\
We collected RNA-Seq data from cBioPortal~\footnote{\url{https://www.cbioportal.org/}} for 9 TCGA held-out studies to evaluate the performance of multimodal fusion. We followed the same splits as those used in unimodal survival prediction tasks, and excluded those without the paired RNA-Seq data. The final splits of the train-val-test are presented in \textbf{Extended Data} Table~\ref{tab:all_ds}.
\\
\textbf{UBC-OCEAN~\cite{ubc_ocean_paper1,ubc_ocean_paper2} for Ovarian Cancer Subtyping (5 classes).}
\\
The UBC-OCEAN (University of British Columbia - Ovarian Cancer subtypE clAssification and outlier detectioN) dataset consists of 538 slides, which aims to classify ovarian cancer subtypes into 5 categories. After performing tissue segmentation, a total of 527 slides were acquired (98 CC, 122 EC, 221 HGSC, 43 LGSC and 43 MC). The class information is presented in \textbf{Extend Data} Table~\ref{tab:cls_ubc}.
\\
\textbf{BCNB datasets~\cite{xu2021predicting} for ER (2 classes), PR (2 classes) and HER2 prediciton (2 classes) of Biopsy Slides.}
\\
The Early Breast Cancer Core-Needle Biopsy (BCNB) WSI dataset, encompasses core-needle biopsy WSIs obtained from patients diagnosed with early breast cancer. We collected 1038 WSIs paired with ER, PR, HER2 status after tissue extraction.
\\
\textbf{Pancancer TCGA~\cite{weinstein2013cancer} and Breast\&Lung Datasets for Zero-shot Slide Retrieval.}
\\
We utilized pan-cancer TCGA datasets (934 cases) held out from pretraining data to evaluate the performance of Zero-shot Slide Retrieval for Slide-to-Report (Image2Text) and Report-to-Slide (Text2Image) retrieval. Furthermore, to evaluate the generalizability of zero-shot slide retrieval, we curated another cohort consisting of breast and lung cancers from ZJ1 and NFH, resulting in Breast\&Lung (500 cases) to ensure a sufficiently large search space. Given that original reports of ZJ1 and NFH are in Chinese, we first translated them into English via GPT-4o-mini before performing retrieval.
\\
\textbf{TCGA Dataset, Nanfang and ZJ-First for Pathological Report Generation.}
\\
During pretraining, we employed training-validation-test splits for some cancer datasets of over 400 cases and other data were put into pretraining materials. Following this setting, we considered all pretraining data containing pathology reports as the training set, and the held-out validation-test sets were re-used, resulting in 7073:452:934 cases for train-validation-test splits. Given TCGA dataset as the internal cohort, we additionally collected two external cohorts: Nanfang (250 cases) and ZJ-First (250 cases) from NFH and ZJ1 of lung cancer and breast cancer, respectively. Similarly, considering original reports of ZJ1 and NFH are in Chinese, they are translated into English via GPT-4o-mini before performing report generation.
\\
\\
\textbf{Evaluation Metrics}
\\
For classification tasks, Macro-AUC and its 95\% confidence interval (CI) are reported considering alleviating the impact of unbalanced data, which doesn't depend on the selection of the decision threshold and is not affected by the sample ratio of classes. For survival prediction tasks, we report the commonly used Concordance Index (C-Index) and its 95\% CI, which is defined as the probability that two randomly selected individuals will have risk predictions correctly ordered. For zero-shot slide retrieval, we reported Recall @5, @10 and @50. In pathological report generation, in line with our prior studies HistGen~\cite{guo2024histgen}, we report various metrics, BLEU@K~\cite{papineni2002bleu}, METEOR~\cite{denkowski2011meteor} and ROUGE-L~\cite{lin2004rouge}, to assess the accuracy of predicted captions against the ground-truth captions from different perspectives. BLEU@K measures the similarity between machine-generated text and ground truth by comparing the presence and frequency of n-grams. METEOR is a metric that evaluates precision and recall by matching unigrams while also factoring in synonyms and word variations between the original text and the reference. On the other hand, ROUGE-L measures the similarity in n-gram overlap between the generated texts and the ground truth.
\\
\\
\textbf{Statistical Analysis.} Unless otherwise specified, we employ non-parametric bootstrapping with 1,000 bootstrap~\cite{efron1994introduction} replicates to estimate 95\% confidence intervals (CI) for all experiments. For each evaluation experiment, the model performing best in the validation split was chosen to be evaluated on test sets or external sets. To assess the observed differences in performance between the two models, we utilize a one-sided Wilcoxon signed-rank test~\cite{wilcoxon1992individual} for statistical significance, following the previous work~\cite{xu2024whole}.
\\
\\
\textbf{Computing Software and Hardware.} We conducted all experiments and analyses in this study using Python (v3.11.5) and PyTorch (v2.2.1, CUDA 11.7) (\url{https:// pytorch.org}) unless stated otherwise, and these can be reproduced with open-source libraries as described below. To pretrain aggregator, the implementation of the text encoder pretrained on PubMed was maintained by the codebase (\url{https://github.com/dmis-lab/biobert}) and its pretrained weights can be assessed in the open-source timm library from Hugging Face (\url{https://huggingface.co}). For extractor pretraining, we initialize the backbone with the pretrained weights of UNI codebase (\url{https://github.com/mahmoodlab/uni}). OpenSlide (v3.4.1) and openslide-python (v1.3.1) were utilized to support the processing of WSIs in conjunction with CLAM (\url{https://github.com/mahmoodlab/CLAM}). Implementations of other visual pretrained encoders compared in the study can be accessed through the following links: ResNet-50 pretrained on ImageNet-1K (\url{https://github.com/mahmoodlab/CLAM}), PLIP (\url{https://github.com/PathologyFoundation/plip}), CONCH (\url{https://github.com/mahmoodlab/CONCH}), CHIEF (\url{https://github.com/hms-dbmi/CHIEF}) and GigaPath (\url{https://github.com/prov-gigapath/prov-gigapath}). Implementations of zero-shot learning for WSIs were provided in MI-Zero (\url{https://github.com/mahmoodlab/MI-Zero}). For training MIL models for downstream tasks, we adapted the code of ABMIL from the CLAM codebase (\url{https://github.com/mahmoodlab/CLAM}). For multimodal survival prediction, we used the off-the-shelf multimodal fusion models: MCAT (\url{https://github.com/mahmoodlab/MCAT}), Porpoise (\url{https://github.com/mahmoodlab/PORPOISE}), MOTCat (\url{https://github.com/Innse/MOTCat}) and CMTA (\url{https://github.com/FT-ZHOU-ZZZ/CMTA}). For pathological report generation, HistGen (\url{https://github.com/dddavid4real/HistGen}) is applied. We used 4 $\times$ 80 GB NVIDIA H800 GPUs (graphics processing unit) for pretraining aggregator and a single 80 GB NVIDIA H800 GPU for pretraining extractor. These GPUs were set up for multi-GPU, multi-node training, employing distributed data-parallel (DDP) techniques. All other experiments for downstream tasks were conducted on single 24 GB NVIDIA 3090 GPUs or single 80 GB H800 GPU.

\backmatter

\bmhead{Data availability} This study incorporates a total of 97 oncological tasks for downstream evaluation, in which 69 tasks are evaluated on public datasets and 28 tasks are assessed on private cohorts. Pretraining data are curated from TCGA and cBioportal. Regarding the data from Nanfang Hospital of Southern Medical University (NFH), Shandong Provincial Qianfoshan Hospital (QFS), The First Affiliated Hospital of Kunming Medical University in Yunnan (YN1), The Third Affiliated Hospital of Kunming Medical University in Yunnan (YN3), and The First Affiliated Hospital of Zhejiang University School of Medicine (ZJ1), these datasets are not publicly available due to patient privacy obligations, institutional review board requirements, and data use agreements. However, researchers interested in accessing de-identified data may submit a reasonable request directly to the corresponding author, subject to obtaining the necessary ethical approvals and complying with institutional policies. The details of these datasets are demonstrated in \textbf{Extended Data} Table~\ref{tab:public_source}.

\bmhead{Code availability} The code and weights of mSTAR have been made available on GitHub (\url{https://github.com/Innse/mSTAR}).

\bmhead{Ethics declarations} This study has been reviewed and approved by the Human and Artefacts Research Ethics Committee (HAREC). The protocol number is HREP-2024-0212.

\bmhead{Author contributions}
Y.X., Y.W. and H.C. conceived the study. Y.X. designed the pretraining approach and experiments, organized the data for downstream evaluation, and prepared the manuscript. Y.W. performed foundation model pretraining and conducted evaluation on downstream tasks including molecular prediction, survival prediction, zero-shot prediction and report generation. F.Z. conducted evaluation of pathological diagnosis, molecular prediction and multimodal fusion. J.M., C.J. and H.Z. collected the data and assisted in results analysis and coding. S.Y. participated in coding for pretraining. J.L. provided insightful interpretation and specialized clinical validation. Z.Z., C.Z., Z.L., L.L and X.Z. provided preprocessed data for downstream tasks. H.L. assisted in the experimental design. X.W., A.H. and R.C.K.C. assisted in the design of evaluation on clinical tasks and the interpretation of experimental results. J.W. assisted in curating RNA-Seq data and offered suggestions for experimental designs. H.C. supervised the research.

\bmhead{Acknowledgements}
This work was supported by the National Natural Science Foundation of China (No. 62202403), Innovation and Technology Commission (Project No. PRP/034/22FX and ITCPD/17-9), Research Grants Council of the Hong Kong Special Administrative Region, China (No. R6003-22 and C4024-22GF) and HKUST Frontier Technology Research for Joint Institutes with Industry (No. OKT24EG01).

\bmhead{Declarations} The authors have no conflicts of interest to declare.

\noindent

\bigskip






\begin{appendices}
\section{Extended Data}
\begin{table*}[ht]
  \centering
  \caption{Data Statistics of all datasets used for evaluation in this work.}
  \scalebox{0.59}{
  \centering
\begin{tabular}{l|l|l|l|cccc|cc|c}
    \toprule
    \pmb{No.} & \pmb{Application} & \pmb{Task} & \pmb{Dataset} & \pmb{Train} & \pmb{Val} & \pmb{Test} & Total & \pmb{Evaluation} & \pmb{Internal} & \pmb{Level} \\
    \midrule
    \rowcolor[rgb]{ .753,  .902,  .961} \textcolor[rgb]{ .063,  .282,  .38}{1} & \textcolor[rgb]{ .063,  .282,  .38}{Pathlogical Diagnosis} & \textcolor[rgb]{ .063,  .282,  .38}{Pathological Subtyping} & \textcolor[rgb]{ .063,  .282,  .38}{BRCA-PathSubtype } & \textcolor[rgb]{ .063,  .282,  .38}{656} & \textcolor[rgb]{ .063,  .282,  .38}{95} & \textcolor[rgb]{ .063,  .282,  .38}{234} & \textcolor[rgb]{ .063,  .282,  .38}{985} & \textcolor[rgb]{ .063,  .282,  .38}{Held-out} & \textcolor[rgb]{ .063,  .282,  .38}{-} & \textcolor[rgb]{ .063,  .282,  .38}{Slide} \\
    \textcolor[rgb]{ .063,  .282,  .38}{2} & \textcolor[rgb]{ .063,  .282,  .38}{Pathlogical Diagnosis} & \textcolor[rgb]{ .063,  .282,  .38}{Pathological Subtyping} & \textcolor[rgb]{ .063,  .282,  .38}{GBMLGG\_PathSubtype } & \textcolor[rgb]{ .063,  .282,  .38}{839} & \textcolor[rgb]{ .063,  .282,  .38}{200} & \textcolor[rgb]{ .063,  .282,  .38}{237} & \textcolor[rgb]{ .063,  .282,  .38}{1276} & \textcolor[rgb]{ .063,  .282,  .38}{Held-out} & \textcolor[rgb]{ .063,  .282,  .38}{-} & \textcolor[rgb]{ .063,  .282,  .38}{Slide} \\
    \rowcolor[rgb]{ .753,  .902,  .961} \textcolor[rgb]{ .063,  .282,  .38}{3} & \textcolor[rgb]{ .063,  .282,  .38}{Pathlogical Diagnosis} & \textcolor[rgb]{ .063,  .282,  .38}{Pathological Subtyping} & \textcolor[rgb]{ .063,  .282,  .38}{HANCOCK\_PathSubtype } & \textcolor[rgb]{ .063,  .282,  .38}{470} & \textcolor[rgb]{ .063,  .282,  .38}{68} & \textcolor[rgb]{ .063,  .282,  .38}{134} & \textcolor[rgb]{ .063,  .282,  .38}{672} & \textcolor[rgb]{ .063,  .282,  .38}{Independent} & \textcolor[rgb]{ .063,  .282,  .38}{-} & \textcolor[rgb]{ .063,  .282,  .38}{Slide} \\
\bottomrule
    \end{tabular}%
  }
\end{table*}%

\begin{table*}\ContinuedFloat
\centering
\caption{(Continuous Table) Data Statistics of all datasets used for evaluation in this work.}
\scalebox{0.7}{
\centering
    \begin{tabular}{l|l|l|l|cccc|cc|c}
    \toprule
    \pmb{No.} & \pmb{Application} & \pmb{Task} & \pmb{Dataset} & \pmb{Train} & \pmb{Val} & \pmb{Test} & \pmb{Total} & \pmb{Evaluation} & \pmb{Internal} & \pmb{Level} \\
    \midrule
    \textcolor[rgb]{ .063,  .282,  .38}{78} & \textcolor[rgb]{ .063,  .282,  .38}{Multimodal Fusion} & \textcolor[rgb]{ .063,  .282,  .38}{Overall Survival} & \textcolor[rgb]{ .063,  .282,  .38}{OS\_BRCA } & \textcolor[rgb]{ .063,  .282,  .38}{705} & \textcolor[rgb]{ .063,  .282,  .38}{101} & \textcolor[rgb]{ .063,  .282,  .38}{201} & \textcolor[rgb]{ .063,  .282,  .38}{1007} & \textcolor[rgb]{ .063,  .282,  .38}{Held-out} & \textcolor[rgb]{ .063,  .282,  .38}{-} & \textcolor[rgb]{ .063,  .282,  .38}{Patient} \\
    \rowcolor[rgb]{ .753,  .902,  .961} \textcolor[rgb]{ .063,  .282,  .38}{79} & \textcolor[rgb]{ .063,  .282,  .38}{Multimodal Fusion} & \textcolor[rgb]{ .063,  .282,  .38}{Overall Survival} & \textcolor[rgb]{ .063,  .282,  .38}{OS\_CRC } & \textcolor[rgb]{ .063,  .282,  .38}{377} & \textcolor[rgb]{ .063,  .282,  .38}{56} & \textcolor[rgb]{ .063,  .282,  .38}{108} & \textcolor[rgb]{ .063,  .282,  .38}{541} & \textcolor[rgb]{ .063,  .282,  .38}{Held-out} & \textcolor[rgb]{ .063,  .282,  .38}{-} & \textcolor[rgb]{ .063,  .282,  .38}{Patient} \\
    \textcolor[rgb]{ .063,  .282,  .38}{80} & \textcolor[rgb]{ .063,  .282,  .38}{Multimodal Fusion} & \textcolor[rgb]{ .063,  .282,  .38}{Overall Survival} & \textcolor[rgb]{ .063,  .282,  .38}{OS\_GBMLGG } & \textcolor[rgb]{ .063,  .282,  .38}{383} & \textcolor[rgb]{ .063,  .282,  .38}{55} & \textcolor[rgb]{ .063,  .282,  .38}{112} & \textcolor[rgb]{ .063,  .282,  .38}{550} & \textcolor[rgb]{ .063,  .282,  .38}{Held-out} & \textcolor[rgb]{ .063,  .282,  .38}{-} & \textcolor[rgb]{ .063,  .282,  .38}{Patient} \\
\bottomrule
    \end{tabular}
    }
    \label{tab:all_ds}%
\end{table*}

\begin{table*}[htbp]
  \centering
  \caption{Average Performance of 15 types of oncological tasks. ** means $P < 0.01$ and *** indicates $P < 0.001$.}
  \scalebox{0.65}{
    \begin{tabular}{l|l|ccccccc}
    \toprule
    Task  & Metric & R50   & PLIP  & CONCH & UNI   & CHIEF & GigaPath & mSTAR (Ours) \\
    \midrule
    Pathological Subtyping*** & Macro-AUC & 0.6807±0.154 & 0.7555±0.149 & 0.7686±0.174 & 0.7761±0.161 & 0.7648±0.156 & 0.7659±0.177 & \pmb{0.7930±0.152} \\
    \midrule
    Metastasis Detection*** & Macro-AUC & 0.8205±0.109 & 0.8824±0.085 & 0.9511±0.029 & 0.9482±0.025 & 0.9285±0.041 & 0.9187±0.058 & \pmb{0.9652±0.027} \\
    \midrule
    Morphology Prediction*** & Macro-AUC & 0.6785±0.109 & 0.7502±0.092 & 0.7888±0.114 & 0.8101±0.099 & 0.7629±0.120 & 0.7802±0.101 & \pmb{0.8164±0.097} \\
    \midrule
    Pathological Grading & Macro-AUC & 0.8872±0.008 & 0.9046±0.008 & 0.9220±0.008 & \pmb{0.9457±0.006} & 0.9291±0.007 & 0.9221±0.007 & 0.9409±0.006 \\
    \midrule
    Pathological Staging*** & Macro-AUC & 0.7527±0.062 & 0.7184±0.055 & 0.7617±0.055 & 0.7645±0.052 & 0.7450±0.056 & 0.7508±0.061 & \pmb{0.7786±0.054} \\
    \midrule
    Mutation Prediction*** & Macro-AUC & 0.5473±0.132 & 0.6122±0.140 & 0.6331±0.139 & 0.6711±0.143 & 0.6399±0.120 & 0.6331±0.131 & \pmb{0.6973±0.129} \\
    \midrule
    IHC Biomarker Prediction*** & Macro-AUC & 0.6108±0.124 & 0.7039±0.112 & 0.7215±0.131 & 0.7455±0.136 & 0.7334±0.127 & 0.7174±0.138 & \pmb{0.7631±0.134} \\
\bottomrule
    \end{tabular}%
    }
  \label{tab:overall_performance}%
\end{table*}%

\begin{figure*}
    \centering
    \includegraphics[scale=0.45]{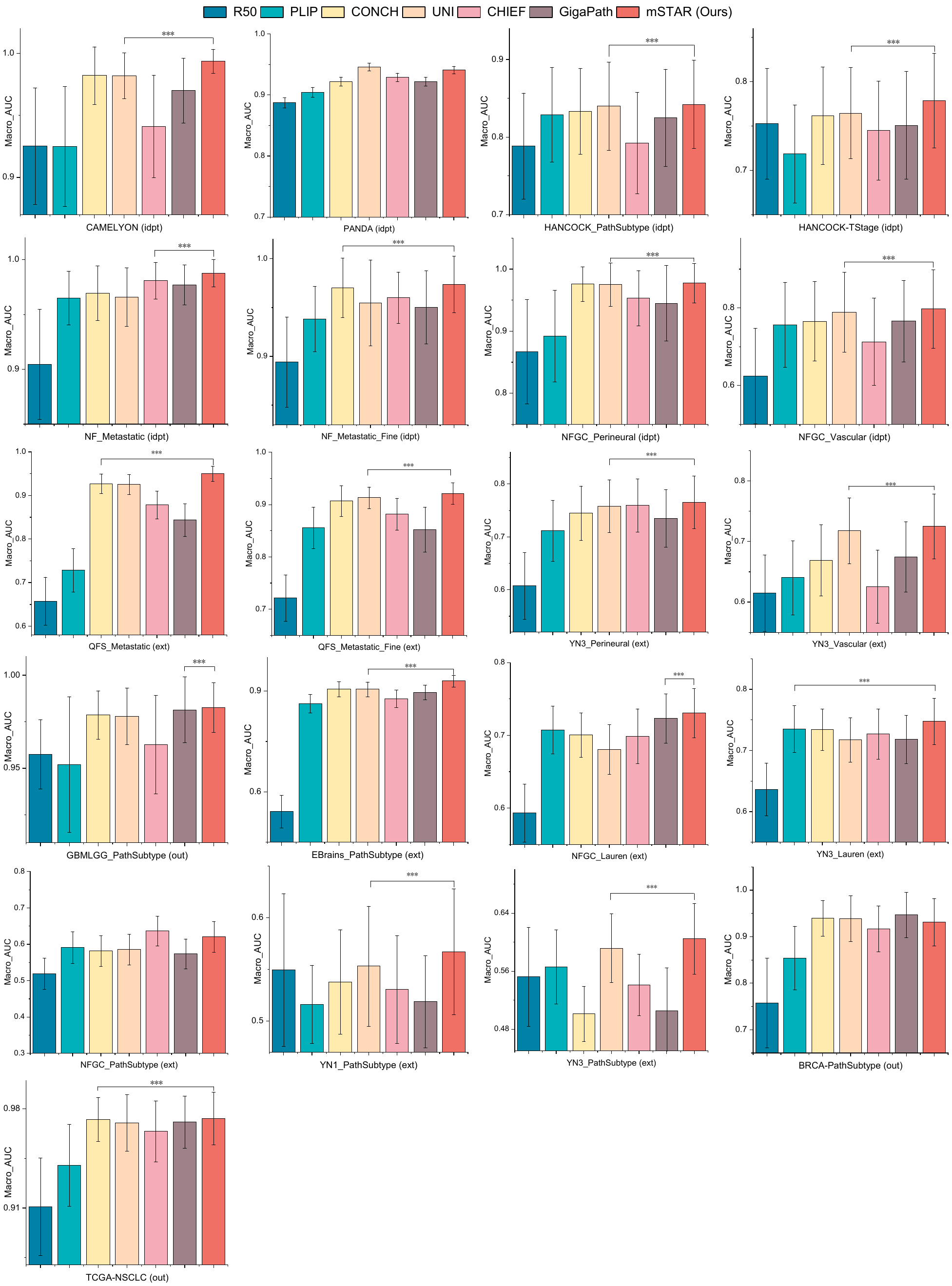}
    \caption{\textbf{Performance of Pathological Diagnosis.} * represents $P<0.05$, ** means $P<0.01$ and *** indicates $P<0.001$. `out' refers to held-out datasets. `idpt' means independent datasets. `ext' represents external datasets.}
    \label{fig:dianosis-raw}
\end{figure*}

\begin{figure*}
    \centering
    \includegraphics[scale=0.42]{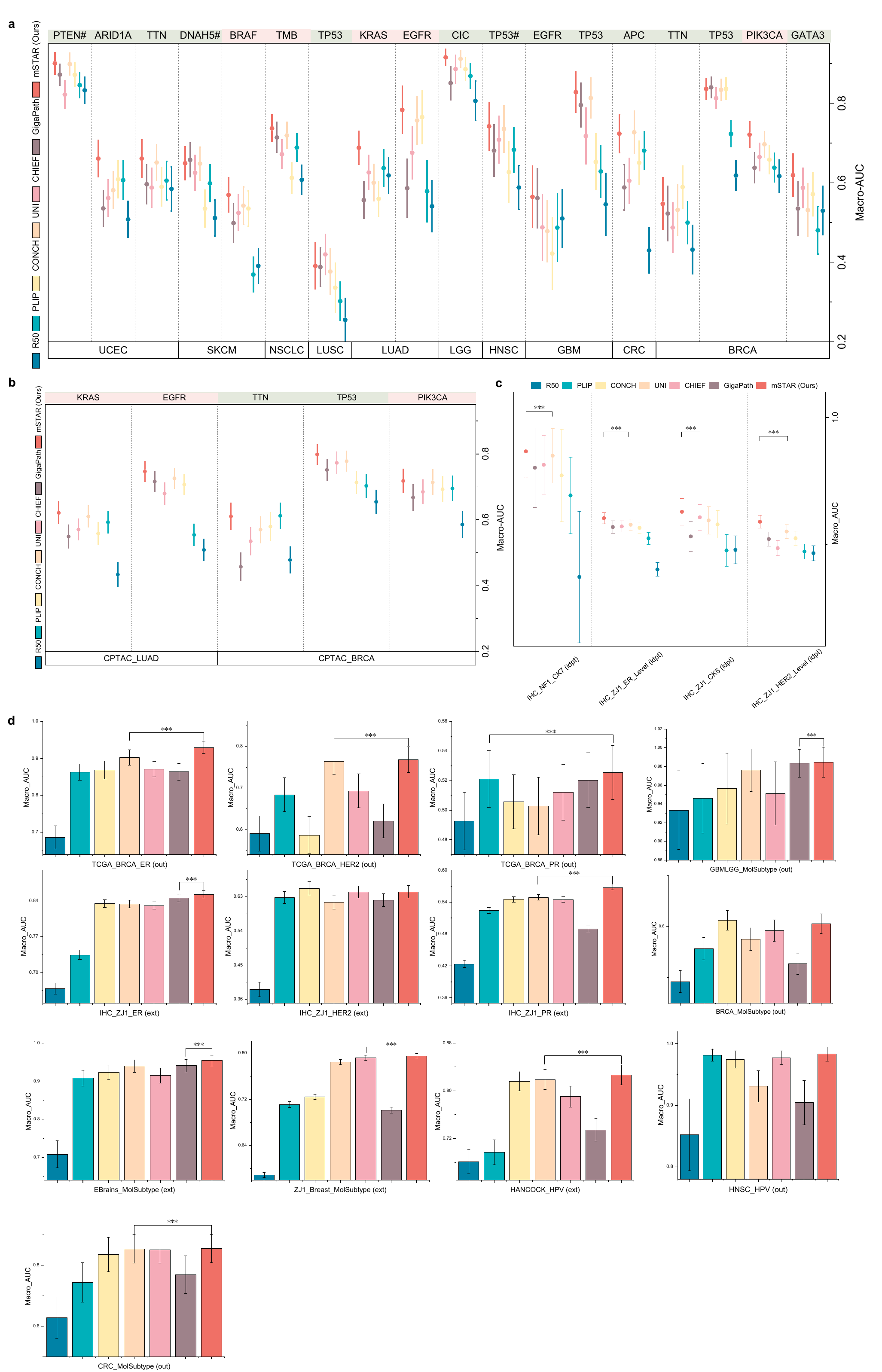}
    \caption{\textbf{Performance of Molecular Prediction.} \pmb{a}, Mutation Prediction. Among the genes where mSTAR performed the best, results showed a statistically significant difference compared to the second-best FM, except for the genes marked with `\#'. \pmb{b}, Performance of Mutation Prediction on external cohorts. \pmb{c}, Independent IHC Biomarker Prediction. \pmb{d}, Held-out and External IHC Biomarker Prediction, and Held-out Molecular Subtyping. * represents $P<0.05$, ** means $P<0.01$ and *** indicates $P<0.001$. `out' refers to held-out datasets. `idpt' means independent datasets. `ext' represents external datasets. \colorbox{green_var}{High-frequency mutations} are highlighted in green and genes related to \colorbox{red_var}{FDA-approved therapies} are highlighted in red.}
    \label{fig:molecular-raw}
\end{figure*}

\begin{figure*}
    \centering
    \includegraphics[scale=0.45]{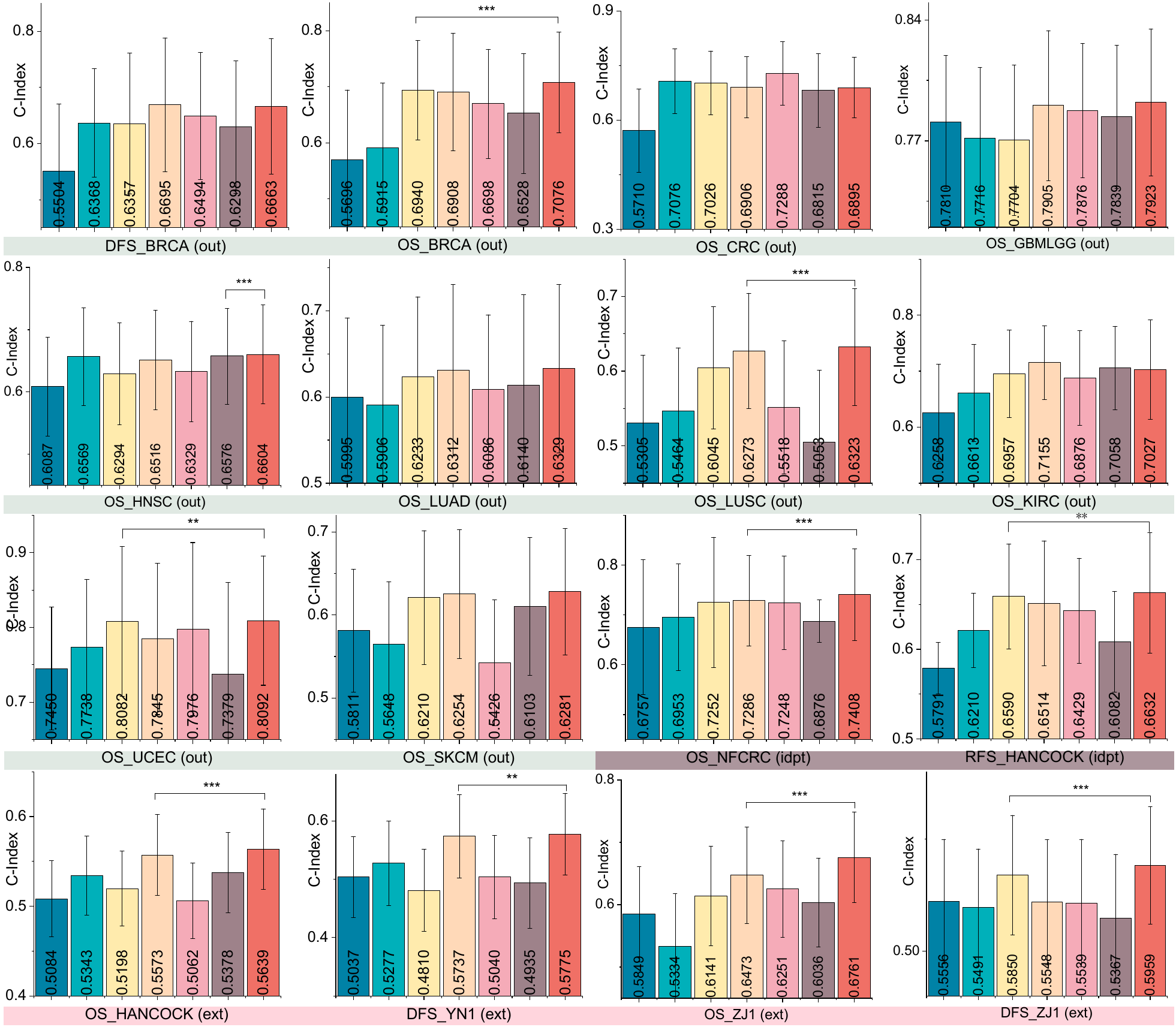}
    \caption{\textbf{Performance of Survival Prediction.} * represents $P<0.05$, ** means $P<0.01$ and *** indicates $P<0.001$. `out' refers to held-out datasets. `idpt' means independent datasets. `ext' represents external datasets.}
\end{figure*}
\begin{figure*}
    \centering
    \includegraphics[scale=0.32]{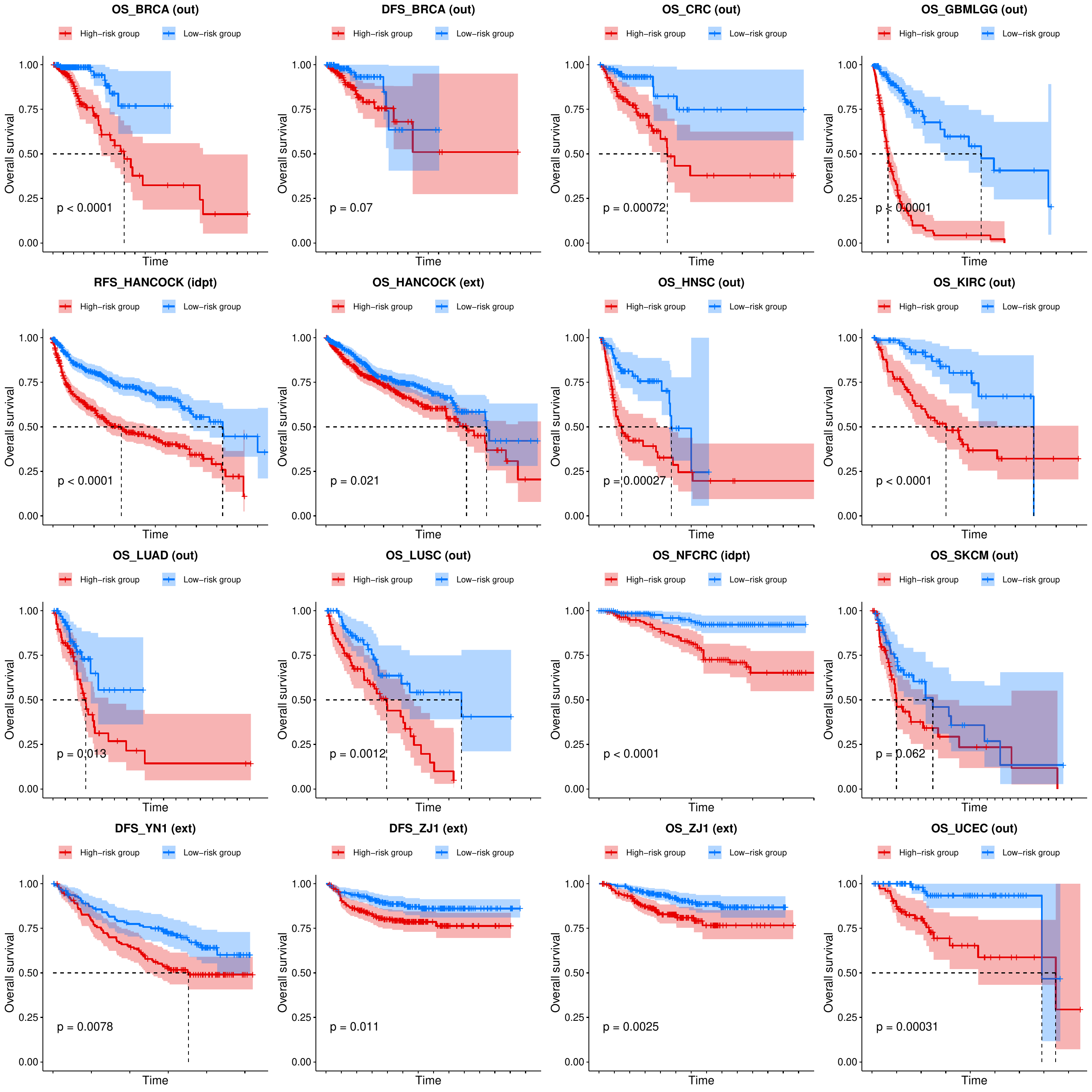}
    \caption{Kaplan-Meier curves of mSTAR on every survival prediction task. $P$-value is given by Logrank Test~\cite{mantel1966evaluation}.}
    \label{fig:km_curve}
\end{figure*}
\begin{figure*}[h]
    \centering
    \begin{subfigure}{\textwidth}
        \includegraphics[scale=0.45]{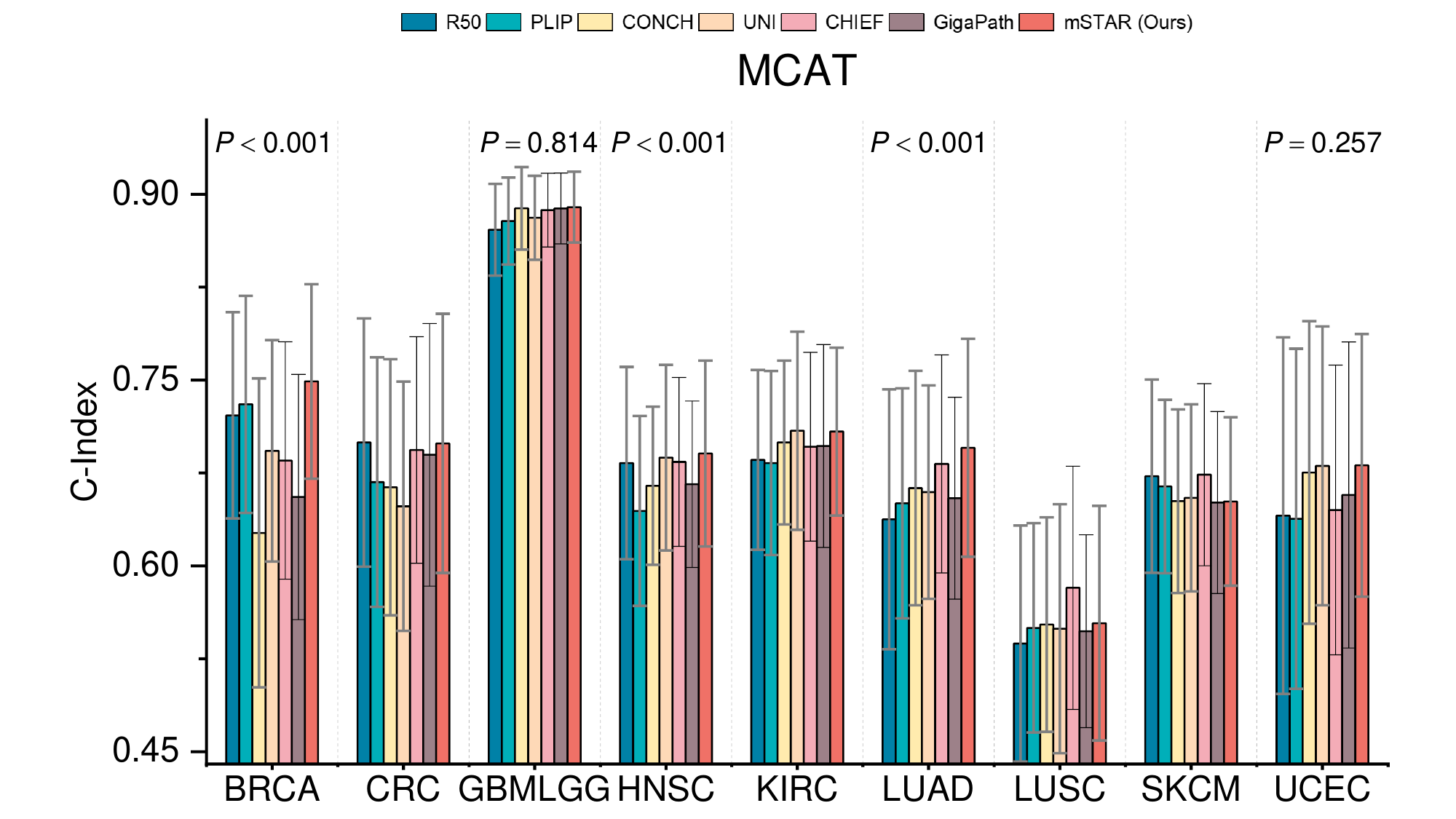}
        \centering
        \subsection{(a)}
    \end{subfigure}
    \begin{subfigure}{\textwidth}
        \includegraphics[scale=0.45]{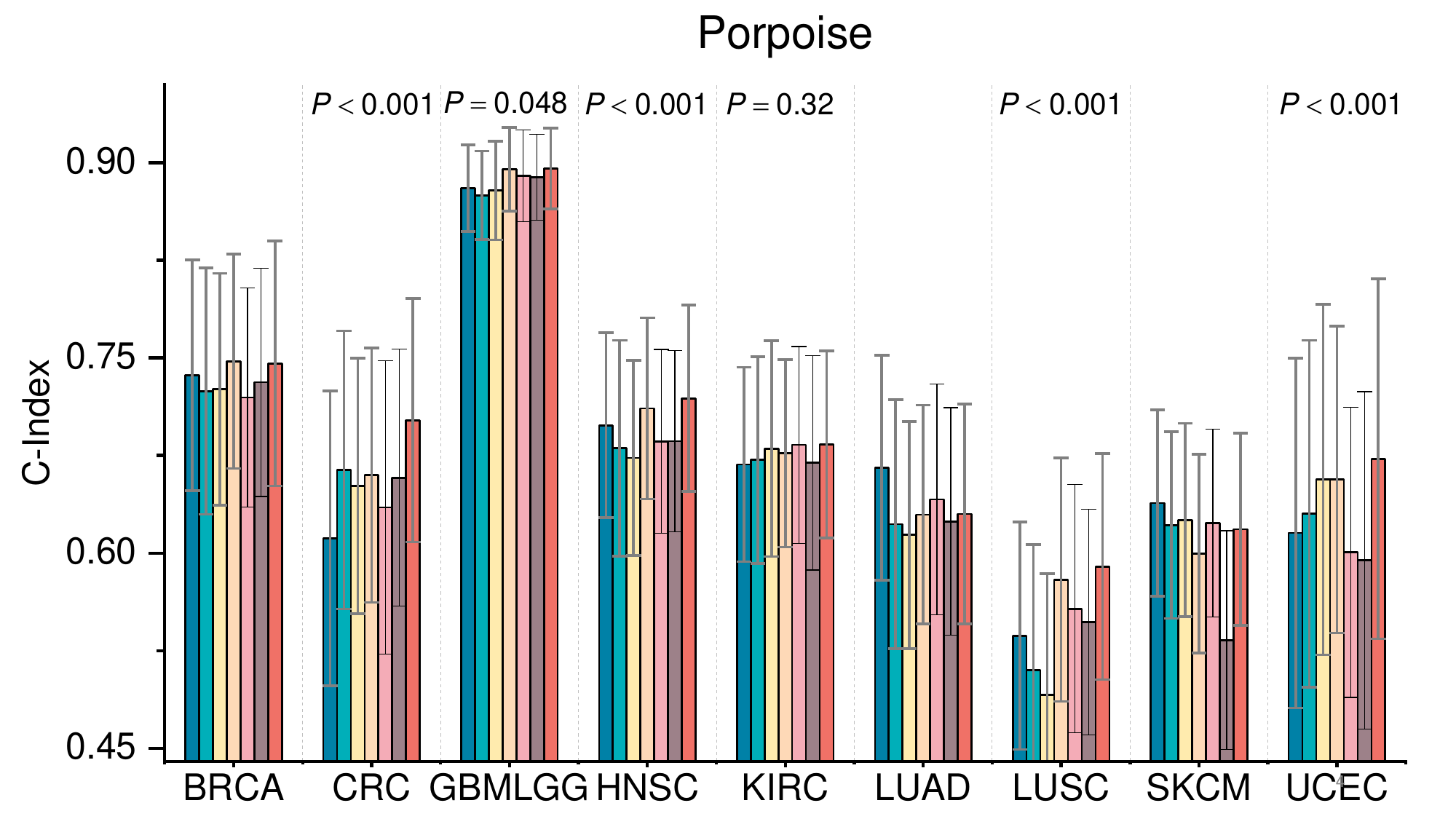}
        \centering
        \subsection{(b)}
    \end{subfigure}
\end{figure*}
\begin{figure*}[h]\ContinuedFloat
\centering
    \begin{subfigure}{\textwidth}
        \includegraphics[scale=0.45]{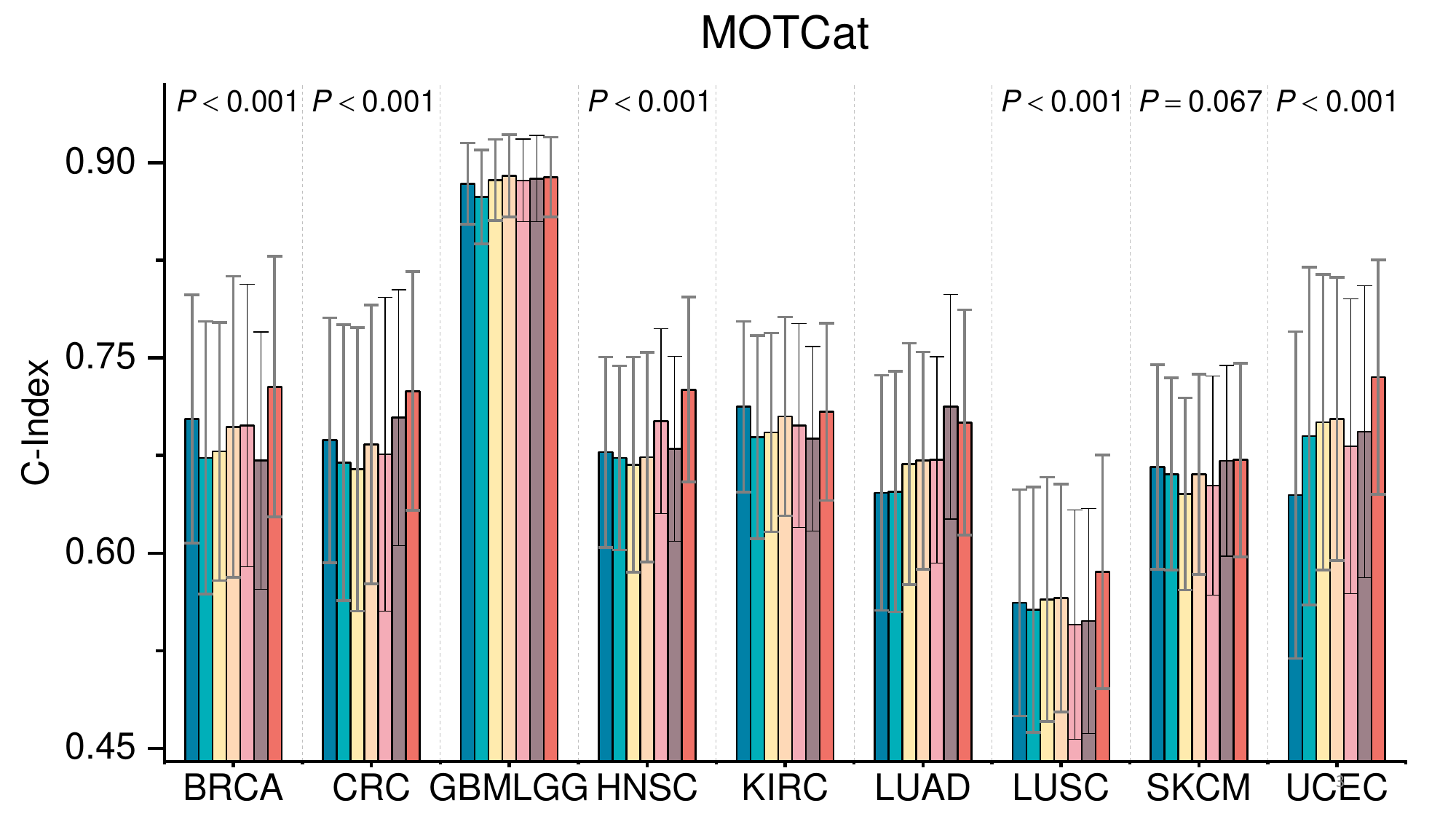}
        \centering
        \subsection{(c)}
    \end{subfigure}
    \begin{subfigure}{\textwidth}
        \includegraphics[scale=0.45]{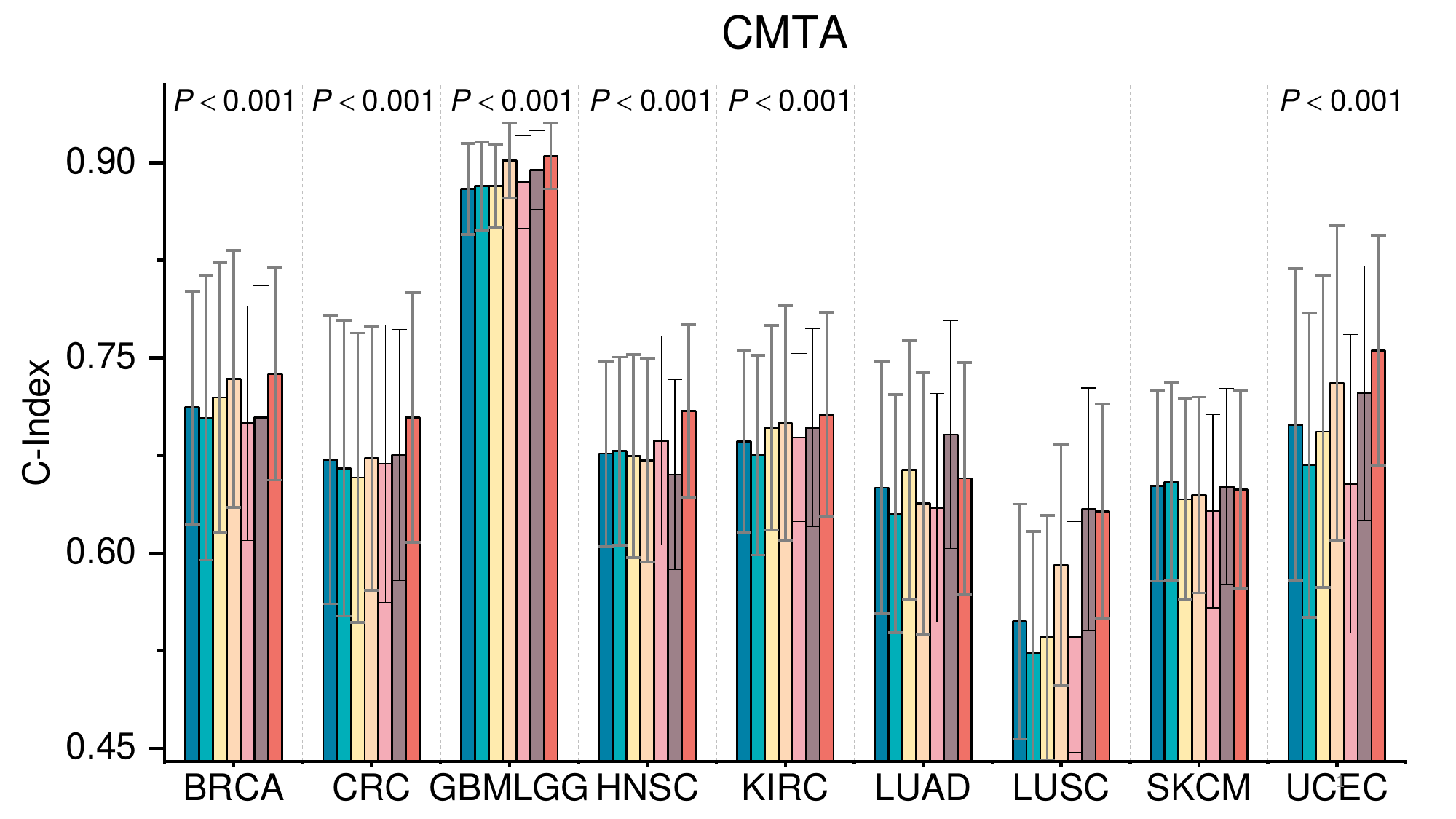}
        \centering
        \subsection{(d)}
    \end{subfigure}
    \caption{\textbf{Performance (C-Index and 95\% CI) of Multimodal Fusion} from 4 multimodal fusion models across 9 multimodal held-out datasets on Overall Survival consisting of pathological slides and RNASeq data: (a) MCAT, (b) Porpoise, (c) MOTCat and (d) CMTA. If the performance of mSTAR is the best one compared against the compared FMs, P-value would be given through one-sided Wilcoxon signed-rank test between mSTAR and the second-best FM. The colors of legends are shared across all sub-figures.}
\end{figure*}

\begin{figure*}
    \centering
    \begin{subfigure}{\textwidth}
    \includegraphics[scale=0.45]{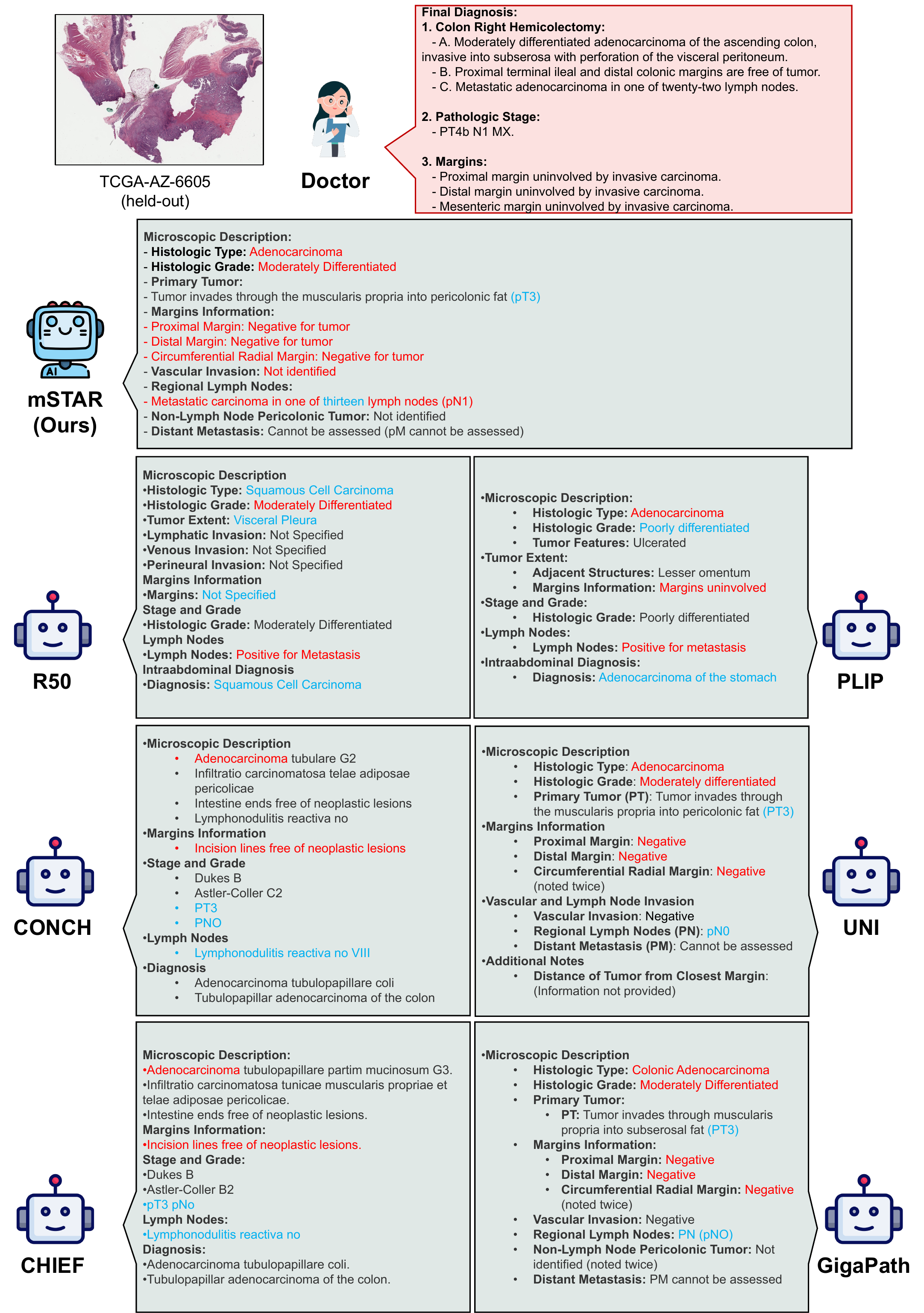}
    \subcaption{TCGA-AZ-6605 (held-out)}
    \end{subfigure}
    
\end{figure*}
\begin{figure*}[h]\ContinuedFloat
\centering
    \begin{subfigure}{\textwidth}
    \centering
    \includegraphics[scale=0.45]{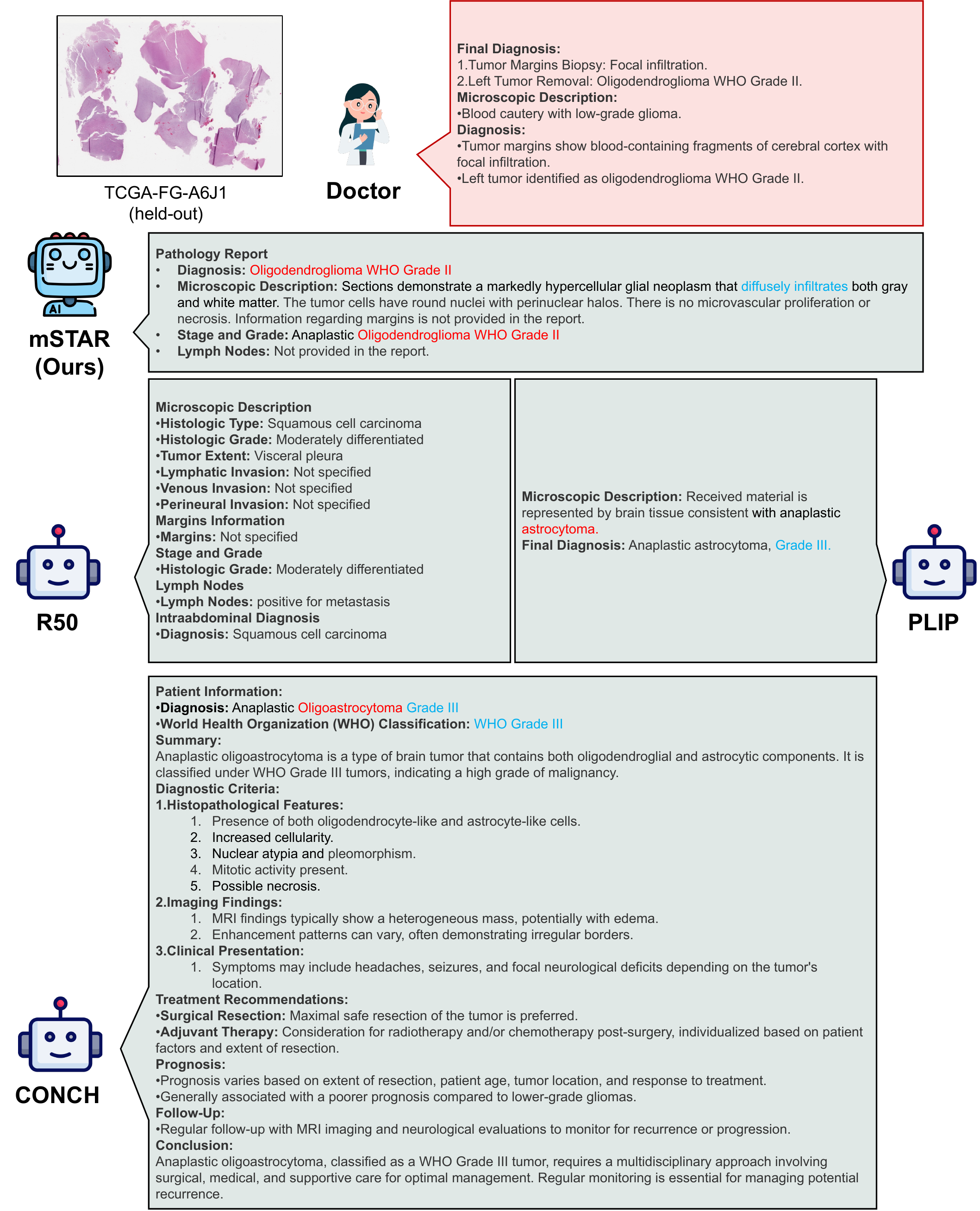}
    \subcaption{TCGA-FG-A6J1 (part1, held-out)}
    \end{subfigure}
\end{figure*}
\begin{figure*}[h]\ContinuedFloat
\centering
    \begin{subfigure}{\textwidth}
    \centering
    \includegraphics[scale=0.48]{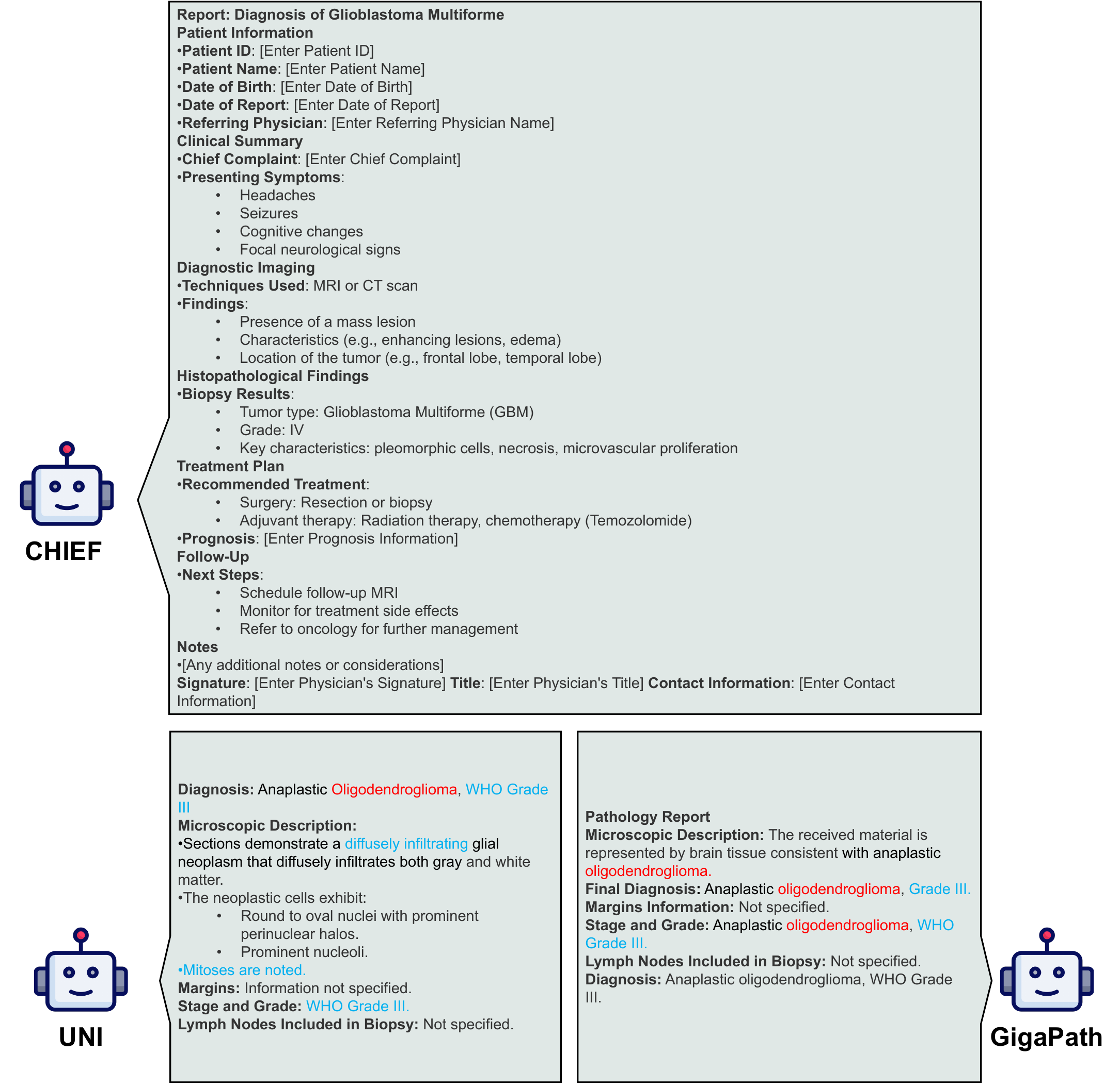}
    \subcaption{TCGA-FG-A6J1 (part2, held-out)}
    \end{subfigure}
\end{figure*}
\begin{figure*}[h]\ContinuedFloat
\centering
    \begin{subfigure}{\textwidth}
    \centering
    \includegraphics[scale=0.42]{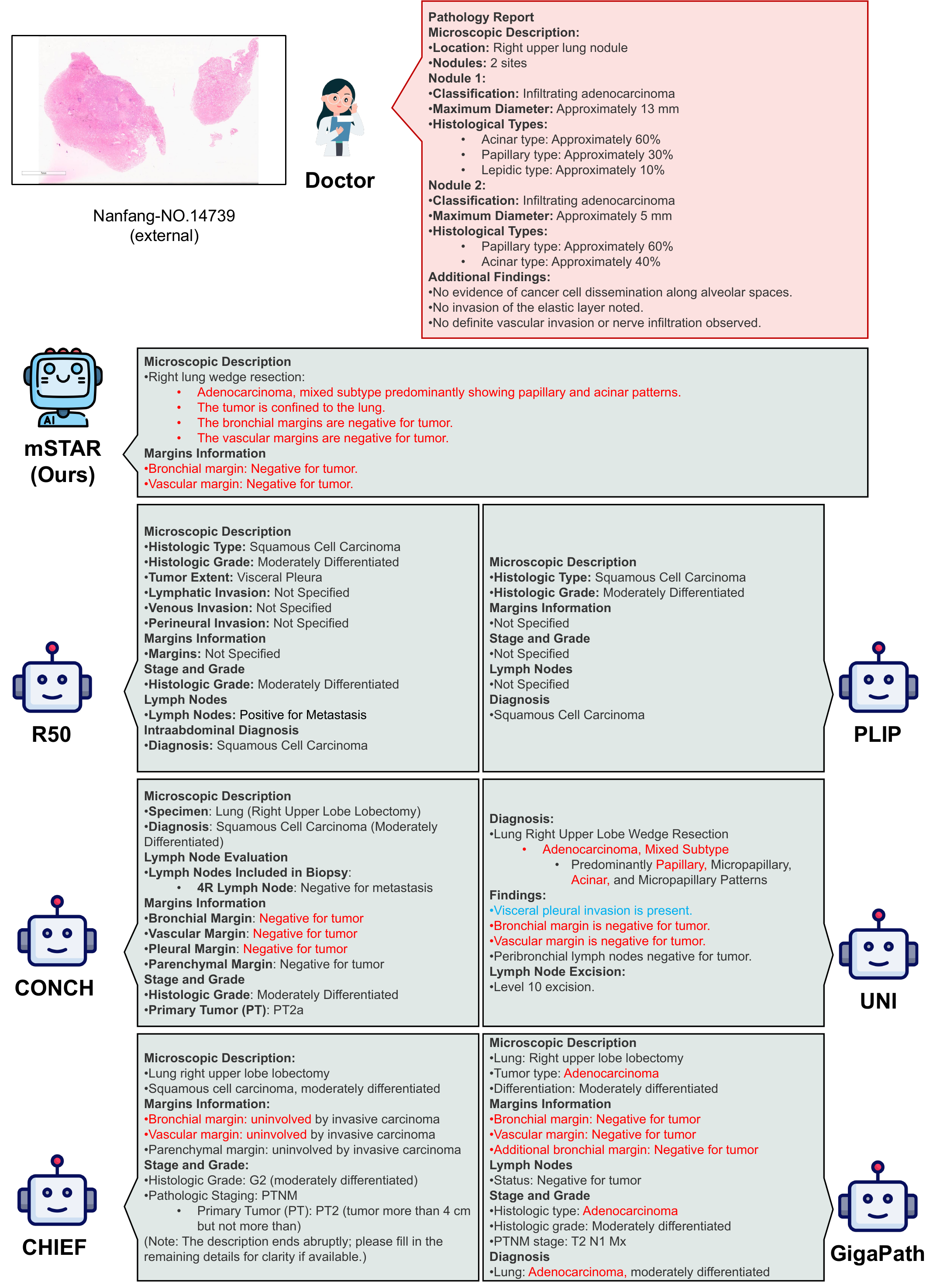}
    \subcaption{Nanfang-NO.14739 (external)}
    \end{subfigure}
\end{figure*}
\begin{figure*}[h]\ContinuedFloat
\centering
    \begin{subfigure}{\textwidth}
    \centering
    \includegraphics[scale=0.41]{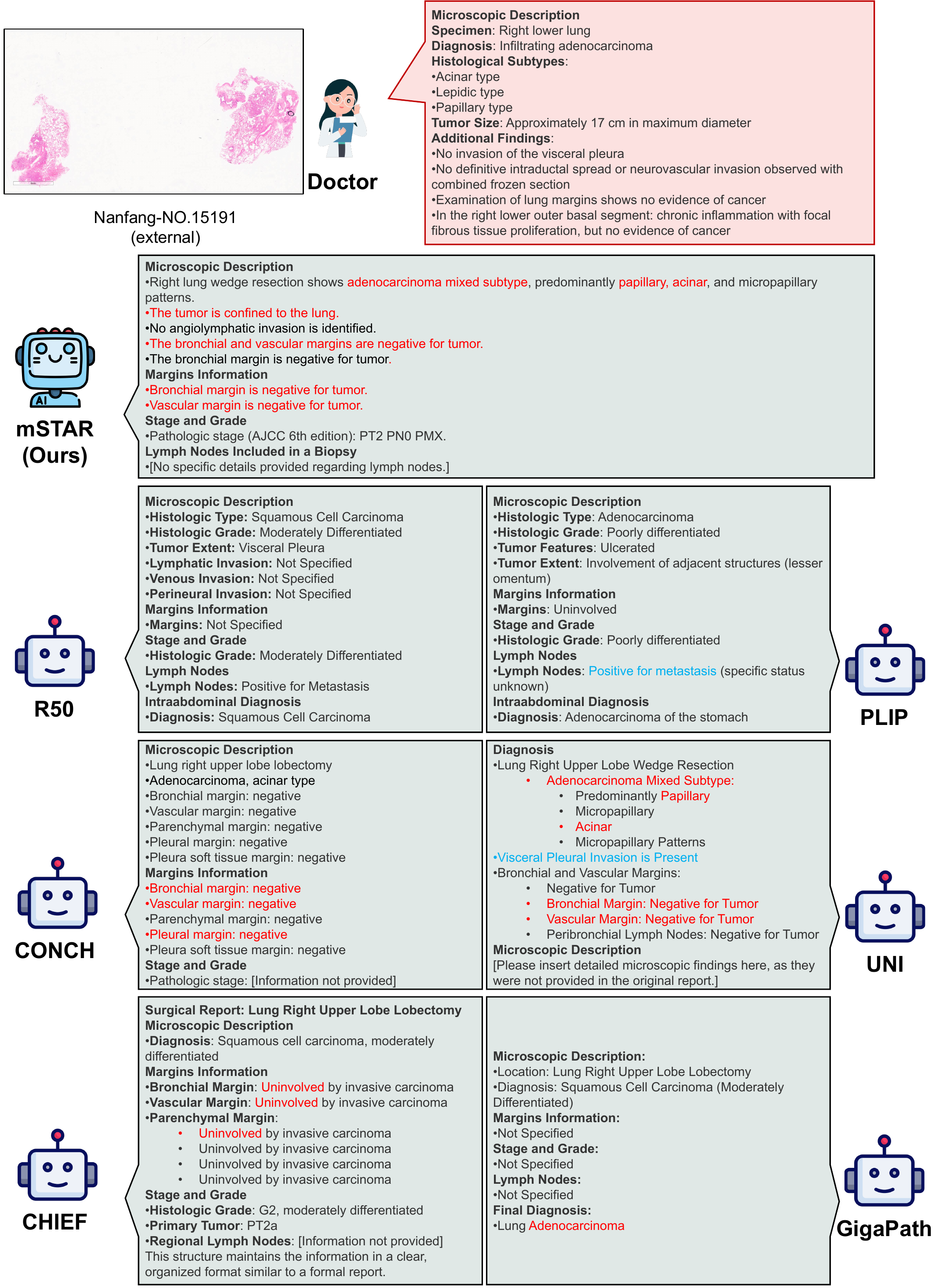}
    \subcaption{Nanfang-NO.15191 (external)}
    \end{subfigure}
\end{figure*}
\begin{figure*}[h]\ContinuedFloat
\centering
    \begin{subfigure}{\textwidth}
    \centering
    \includegraphics[scale=0.41]{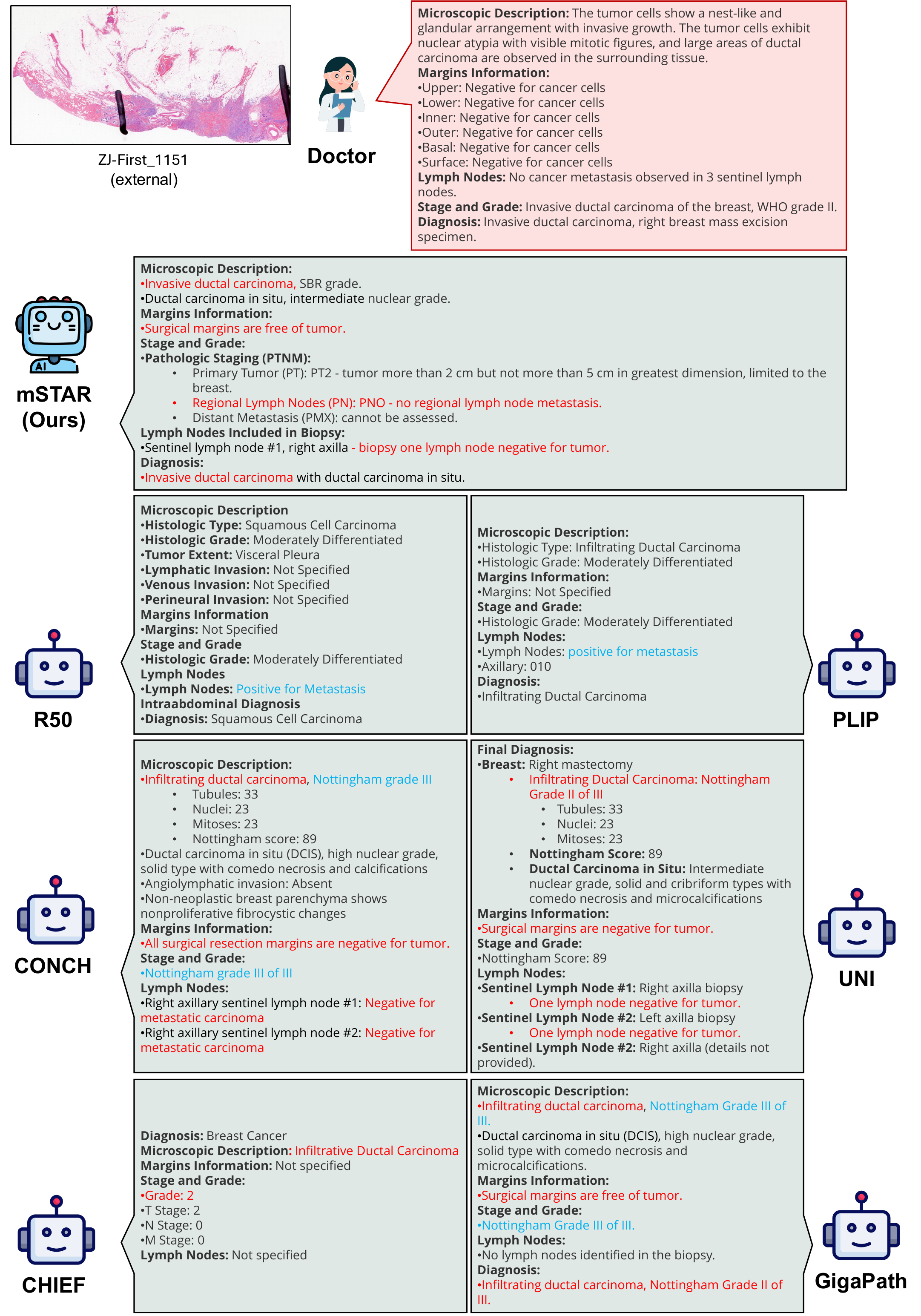}
    \subcaption{ZJ-First\_1151 (external)}
    \end{subfigure}
\end{figure*}
\begin{figure*}[h]\ContinuedFloat

\centering
    \begin{subfigure}{\textwidth}
    \centering
    \includegraphics[scale=0.4]{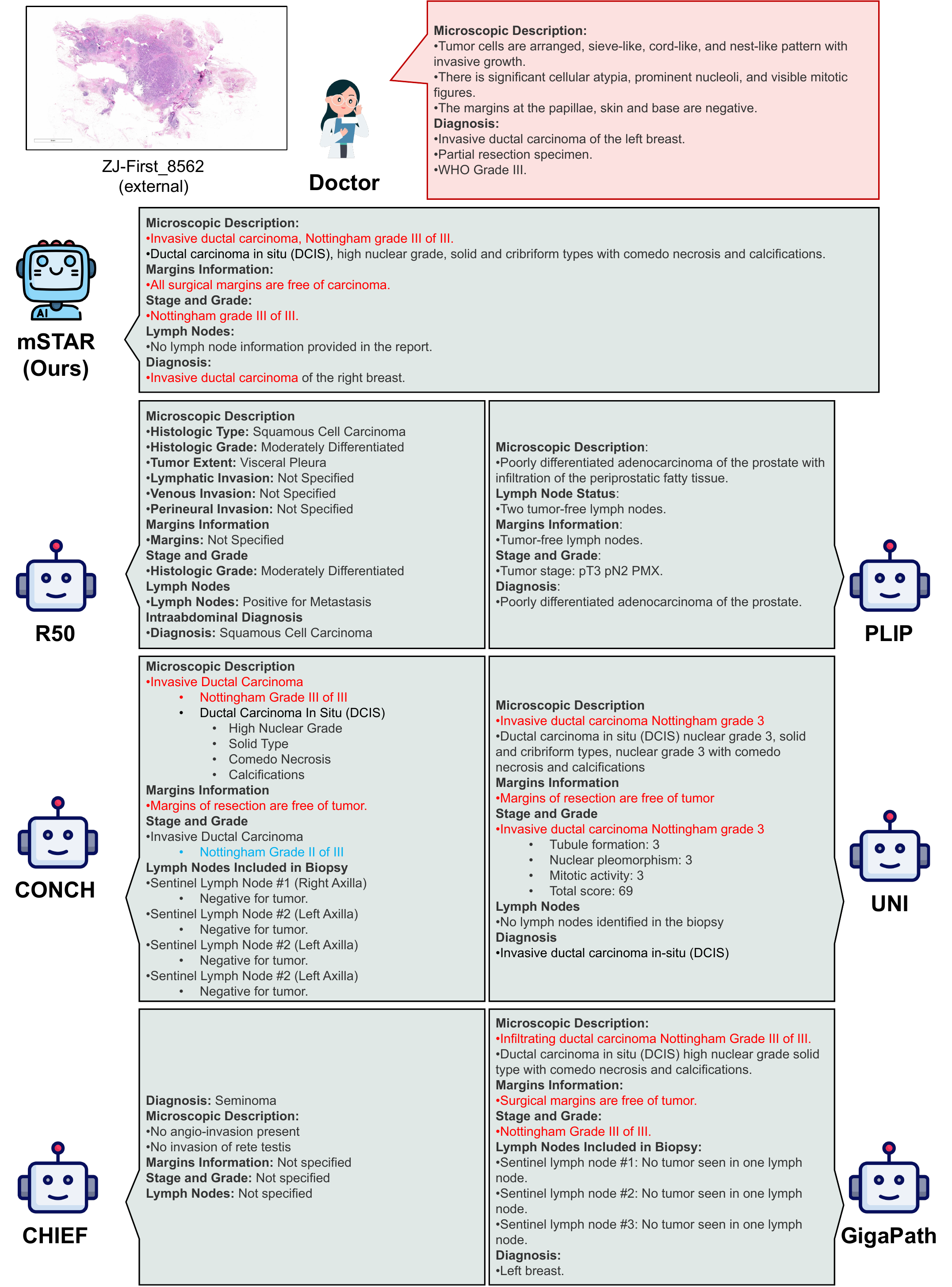}
    \subcaption{ZJ-First\_8562 (external)}
    \end{subfigure}
    \caption{\textbf{Pathology Reports Generation} for several cases. The words highlighted in {\color{red}red} are matched with the ground-truth report, while the ones highlighted in \textcolor[RGB]{0,191,255}{blue} are contradicted to the ground-truth report.}
    \label{fig:cases_report_gen}
\end{figure*}

\begin{table*}[htbp]
  \centering
  \caption{\textbf{Ablation study on different combinational modalities (Pathology, Reports and RNASeq) for survival analysis.} Average C-Index and its std across 9 TCGA survival datasets are reported.}
\begin{tabular}{l|ccc|c}
    \toprule
    \multicolumn{1}{c|}{\multirow{2}[4]{*}{Aggregator (TransMIL)}} & \multicolumn{3}{c|}{Pretrained Data} & \multirow{2}[4]{*}{Avg C-Index } \\
\cmidrule{2-4}          & \multicolumn{1}{c}{Pathology} & \multicolumn{1}{c}{Reports} & RNASeq &  \\
    \midrule
    w/o pretraining &  \checkmark     &       &       & 0.666±0.068 \\
\cmidrule{1-1}\cmidrule{5-5}    \multirow{3}[2]{*}{w/ pretraining} & \checkmark     & \checkmark     &       & 0.679±0.057 \\
          & \checkmark     &       & \checkmark     & 0.679±0.059 \\
          & \checkmark     & \checkmark     & \checkmark     & \pmb{0.689±0.055} \\
    \bottomrule
    \end{tabular}%
  \label{tab:abla_modal}%
\end{table*}%

\begin{table*}[htbp]
  \centering
  \caption{\textbf{Ablation study on different pertaining loss (Inter-modal and Inter-cancer Contrastive Losses) for survival analysis.} Average C-Index and its std across 9 TCGA survival datasets are reported.}
    \begin{tabular}{cc|c}
    \toprule
    \multicolumn{2}{c|}{Loss} & \multicolumn{1}{c}{\multirow{2}[4]{*}{Avg C-Index}} \\
\cmidrule{1-2}    Inter-modality & Inter-cancer &  \\
    \midrule
          & \checkmark & 0.6788$\pm$0.0731 \\
    \checkmark &       & 0.6724$\pm$0.0536 \\
    \checkmark & \checkmark & \pmb{0.6812$\pm$0.0666} \\
    \bottomrule
    \end{tabular}%
  \label{tab:abla_cancer}%
\end{table*}%

\clearpage
\begin{table*}[htbp]
  \centering
  \caption{\textbf{Macro-AUC of Pathological Diagnosis on 21 datasets}, including 10 external datasets, 8 independent datasets and 3 held-out datasets. The best-performing model for each metric is bolded. Std is given by bootstrapping with 1,000 bootstraps.}
    \scalebox{0.55}{
    \begin{tabular}{cc|ccccccc|c}
    \toprule
    \multicolumn{1}{c|}{Task} & dataset & R50   & PLIP  & CONCH & UNI   & CHIEF & GigaPath & mSTAR (Ours) & Evaluation \\
    \midrule
    \multicolumn{1}{c|}{\multirow{10}[2]{*}{Pathological Subtyping}} & BRCA-PathSubtype & 0.7577±0.0963 & 0.8540±0.0686 & 0.9395±0.0387 & 0.9391±0.0489 & 0.9167±0.0489 & \pmb{0.9467±0.0489} & 0.9314±0.0510 & Held-out \\
    \multicolumn{1}{c|}{} & EBrains\_PathSubtype & 0.5415±0.0186 & 0.8632±0.0364 & 0.9065±0.0131 & 0.9059±0.0152 & 0.8772±0.0264 & 0.8962±0.0177 & \pmb{0.9304±0.0134} & External \\
    \multicolumn{1}{c|}{} & GBMLGG\_PathSubtype & 0.9576±0.0682 & 0.9521±0.0607 & 0.9786±0.0552 & 0.9780±0.0566 & 0.9628±0.0654 & 0.9814±0.0627 & \pmb{0.9826±0.0568} & Held-out \\
    \multicolumn{1}{c|}{} & HANCOCK\_PathSubtype & 0.7886±0.0430 & 0.8289±0.0435 & 0.8332±0.0423 & 0.8398±0.0423 & 0.7923±0.0411 & 0.8248±0.0414 & \pmb{0.8422±0.0424} & Independent \\
    \multicolumn{1}{c|}{} & NFGC\_Lauren & 0.5933±0.0344 & 0.7076±0.0289 & 0.7005±0.0156 & 0.6807±0.0200 & 0.6989±0.0215 & 0.7232±0.0183 & \pmb{0.7307±0.0185} & External \\
    \multicolumn{1}{c|}{} & NFGC\_PathSubtype & 0.5190±0.0738 & 0.5915±0.0377 & 0.5816±0.0505 & 0.5854±0.0580 & \pmb{0.6366±0.0522} & 0.5737±0.0444 & 0.6202±0.0609 & External \\
    \multicolumn{1}{c|}{} & TCGA-NSCLC & 0.9109±0.0840 & 0.9403±0.0735 & 0.9727±0.0278 & 0.9702±0.0349 & 0.9643±0.0446 & 0.9708±0.0609 & \pmb{0.9733±0.0316} & Held-out \\
\bottomrule
    \end{tabular}%
    }
  \label{tab:diagnosis}%
\end{table*}%

\begin{table*}[htbp]
  \centering
  \caption{\textbf{Macro-AUC of Mutation Prediction on 18 datasets}. The best-performing model for each metric is bolded. 95\% CI is included in parentheses.}
  \scalebox{0.55}{
    \begin{tabular}{l|ccccccc}
    \toprule
          & \pmb{R50} & \pmb{PLIP} & \pmb{CONCH} & \pmb{UNI} & \pmb{CHIEF} & \pmb{GigaPath} & \pmb{mSTAR (Ours)} \\
    \midrule
    \pmb{BRCA\_GATA3} & 0.5299 (0.4690, 0.5908) & 0.4805 (0.4206, 0.5404) & 0.5715 (0.5169, 0.6261) & 0.5314 (0.4649, 0.5979) & 0.5874 (0.5379, 0.6369) & 0.5352 (0.4673, 0.6031) & \pmb{0.6194 (0.5665, 0.6723)} \\
    \pmb{BRCA\_PIK3CA} & 0.6168 (0.5763, 0.6573) & 0.6382 (0.6020, 0.6744) & 0.6584 (0.6234, 0.6934) & 0.6969 (0.6654, 0.7284) & 0.6651 (0.6311, 0.6991) & 0.6380 (0.6003, 0.6757) & \pmb{0.7215 (0.6897, 0.7533)} \\
    \pmb{BRCA\_TP53} & 0.6184 (0.5804, 0.6564) & 0.7231 (0.6907, 0.7555) & 0.8364 (0.8097, 0.8631) & 0.8341 (0.8075, 0.8607) & 0.8130 (0.7871, 0.8389) & \pmb{0.8403 (0.8144, 0.8662)} & 0.8366 (0.8105, 0.8627) \\
    \pmb{BRCA\_TTN} & 0.4319 (0.3709, 0.4929) & 0.4999 (0.4473, 0.5525) & \pmb{0.5895 (0.5366, 0.6424)} & 0.5320 (0.4684, 0.5956) & 0.4873 (0.4256, 0.5490) & 0.5228 (0.4549, 0.5907) & 0.5471 (0.4819, 0.6123) \\
    \pmb{CRC\_APC} & 0.4300 (0.3732, 0.4868) & 0.6812 (0.6338, 0.7286) & 0.6505 (0.5966, 0.7044) & \pmb{0.7273 (0.6739, 0.7807)} & 0.6053 (0.5486, 0.6620) & 0.5884 (0.5314, 0.6454) & 0.7238 (0.6760, 0.7716) \\
    \pmb{GBM\_EGFR} & 0.5102 (0.4376, 0.5828) & 0.4872 (0.4020, 0.5724) & 0.4221 (0.3321, 0.5121) & 0.4782 (0.4014, 0.5550) & 0.4879 (0.4047, 0.5711) & 0.5612 (0.4869, 0.6355) & \pmb{0.5649 (0.4879, 0.6419)} \\
    \pmb{GBM\_TP53} & 0.5458 (0.4680, 0.6236) & 0.6287 (0.5636, 0.6938) & 0.6526 (0.5826, 0.7226) & 0.8135 (0.7632, 0.8638) & 0.7179 (0.6483, 0.7875) & 0.7958 (0.7405, 0.8511) & \pmb{0.8282 (0.7780, 0.8784)} \\
\bottomrule
    \end{tabular}%
  }
  \label{tab:mutation_abmil}%
\end{table*}%

\begin{table*}[htbp]
  \centering
  \caption{\textbf{Macro-AUC of Mutation Prediction on 5 external datasets}. The best-performing model for each metric is bolded. 95\% CI is included in parentheses.}
  \scalebox{0.58}{
    \begin{tabular}{cc|lllllll|r}
    \toprule
    \multicolumn{1}{c|}{Task} & dataset & \multicolumn{1}{c}{R50} & \multicolumn{1}{c}{PLIP} & \multicolumn{1}{c}{CONCH} & \multicolumn{1}{c}{UNI} & \multicolumn{1}{c}{CHIEF} & \multicolumn{1}{c}{GigaPath} & \multicolumn{1}{c|}{mSTAR (Ours)} & \multicolumn{1}{c}{\pmb{Evaluation}} \\
    \midrule
    \multicolumn{1}{c|}{\multirow{5}[2]{*}{Mutation Prediction}} & \multicolumn{1}{l|}{CPTAC\_BRCA\_PIK3CA} & 0.5856±0.0382	 & 0.6957±0.0366	 & 0.6929±0.0379	 & 0.7143±0.0380	 & 0.6849±0.0355	 & 0.6678±0.0383	 & \pmb{0.7180±0.0353	} & \multicolumn{1}{l}{External} \\
    \multicolumn{1}{c|}{} & \multicolumn{1}{l|}{CPTAC\_BRCA\_TP53} & 0.6545±0.0353 & 0.7031±0.0350 & 0.7140±0.0333 & 0.7782±0.0303 & 0.7729±0.0324 & 0.7517±0.0321 & \pmb{0.7986±0.0293} & \multicolumn{1}{l}{External} \\
    \multicolumn{1}{c|}{} & \multicolumn{1}{l|}{CPTAC\_BRCA\_TTN} & 0.4780±0.0393 & \pmb{0.6122±0.0385} & 0.5794±0.0421 & 0.5699±0.0394 & 0.5349±0.0408 & 0.4571±0.0419 & 0.6104±0.0395 & \multicolumn{1}{l}{External} \\
    \multicolumn{1}{c|}{} & \multicolumn{1}{l|}{CPTAC\_LUAD\_EGFR} & 0.5085±0.0323 & 0.5542±0.0322 & 0.7066±0.0307 & 0.7264±0.0303 & 0.6799±0.0324 & 0.7163±0.0305 & \pmb{0.7470±0.0304} & \multicolumn{1}{l}{External} \\
    \multicolumn{1}{c|}{} & \multicolumn{1}{l|}{CPTAC\_LUAD\_KRAS} & 0.4336±0.0360 & 0.5928±0.0323 & 0.5583±0.0331 & 0.6101±0.0323 & 0.5703±0.0316 & 0.5491±0.0343 & \pmb{0.6212±0.0326} & \multicolumn{1}{l}{External} \\
    \midrule
    \multicolumn{2}{c|}{Overall Macro-AUC} & 0.5320±0.0788 & 0.6316±0.0585 & 0.6502±0.0671 & 0.6798±0.0774 & 0.6486±0.0858 & 0.6284±0.1096 & \pmb{0.6990±0.0728} &  \\
    \bottomrule
    \end{tabular}%
    }
  \label{tab:mutation_ext}%
\end{table*}%

\begin{table*}[htbp]
  \centering
  \caption{\textbf{Macro-AUC of IHC Biomarker Prediction on 10 datasets}. The best-performing model for each metric is bolded. Std is given by bootstrapping with 1,000 bootstraps.}
  \scalebox{0.6}{
    \begin{tabular}{l|lllllll|r}
    \toprule
    \multicolumn{1}{c|}{\pmb{dataset}} & \multicolumn{1}{c}{\pmb{R50}} & \multicolumn{1}{c}{\pmb{PLIP}} & \multicolumn{1}{c}{\pmb{CONCH}} & \multicolumn{1}{c}{\pmb{UNI}} & \multicolumn{1}{c}{\pmb{CHIEF}} & \multicolumn{1}{c}{\pmb{GigaPath}} & \multicolumn{1}{c|}{\pmb{mSTAR (Ours)}} & \multicolumn{1}{c}{\pmb{Evaluation}} \\
    \midrule
    IHC\_ZJ1\_ER (ext) & 0.6683±0.0107 & 0.7345±0.0093 & 0.8350±0.0076 & 0.8339±0.0074 & 0.8309±0.0073 & 0.8457±0.0074 & \pmb{0.8526±0.0071} & \multicolumn{1}{l}{External} \\
    IHC\_ZJ1\_HER2 (ext) & 0.3857±0.0194 & 0.6278±0.0160 & \pmb{0.6521±0.0175} & 0.6153±0.0173 & 0.6427±0.0162 & 0.6212±0.0171 & 0.6429±0.0162 & \multicolumn{1}{l}{External} \\
    IHC\_ZJ1\_PR (ext) & 0.4238±0.0065 & 0.5241±0.0056 & 0.5451±0.0051 & 0.5490±0.0052 & 0.5449±0.0052 & 0.4896±0.0059 & \pmb{0.5673±0.0047} & \multicolumn{1}{l}{External} \\
    \midrule
    TCGA\_BRCA\_ER (out) & 0.6863±0.0318 & 0.8635±0.0219 & 0.8685±0.0245 & 0.9026±0.0205 & 0.8710±0.0211 & 0.8638±0.0226 & \pmb{0.9300±0.0168} & \multicolumn{1}{l}{Held-out} \\
    TCGA\_BRCA\_HER2 (out) & 0.5907±0.0424 & 0.6840±0.0413 & 0.5867±0.0448 & 0.7636±0.0305 & 0.6929±0.0407 & 0.6208±0.0411 & \pmb{0.7679±0.0310} & \multicolumn{1}{l}{Held-out} \\
    TCGA\_BRCA\_PR (out) & 0.4926±0.0195 & 0.5211±0.0192 & 0.5057±0.0184 & 0.5028±0.0195 & 0.5121±0.0188 & 0.5203±0.0184 & \pmb{0.5254±0.0183} & \multicolumn{1}{l}{Held-out} \\
    \midrule
    IHC\_NF1\_CK7 (idpt) & 0.6865±0.1294 & 0.8467±0.0752 & 0.8862±0.0909 & 0.9247±0.0535 & 0.9070±0.0576 & 0.9012±0.0783 & \pmb{0.9335±0.0519} & \multicolumn{1}{l}{Independent} \\
\bottomrule
    \end{tabular}%
    }
  \label{tab:IHC}%
\end{table*}%

\begin{table*}[htbp]
  \centering
  \caption{\textbf{Macro-AUC of Molecular Subtyping on 7 datasets}. The best-performing model for each metric is bolded. Std is given by bootstrapping with 1,000 bootstraps.}
  \scalebox{0.6}{
    \begin{tabular}{l|lllllll|r}
    \toprule
    \multicolumn{1}{c|}{\pmb{dataset}} & \multicolumn{1}{c}{\pmb{R50}} & \multicolumn{1}{c}{\pmb{PLIP}} & \multicolumn{1}{c}{\pmb{CONCH}} & \multicolumn{1}{c}{\pmb{UNI}} & \multicolumn{1}{c}{\pmb{CHIEF}} & \multicolumn{1}{c}{\pmb{GigaPath}} & \multicolumn{1}{c|}{\pmb{mSTAR (Ours)}} & \multicolumn{1}{c}{\pmb{Evaluation}} \\
    \midrule
    EBrains\_MolSubtype (ext) & 0.7086±0.0353 & 0.9081±0.0206 & 0.9232±0.0189 & 0.9396±0.0166 & 0.9150±0.0194 & 0.9410±0.0162 & \pmb{0.9546±0.0138} & \multicolumn{1}{l}{External} \\
    HANCOCK\_HPV (ext) & 0.6814±0.0201 & 0.6972±0.0207 & 0.8161±0.0160 & 0.8189±0.0169 & 0.7905±0.0175 & 0.7347±0.0191 & \pmb{0.8265±0.0166} & \multicolumn{1}{l}{External} \\
    ZJ1\_Breast\_MolSubtype (ext) & 0.5891±0.0044 & 0.7109±0.0054 & 0.7244±0.0048 & 0.7844±0.0046 & 0.7918±0.0045 & 0.7013±0.0052 & \pmb{0.7946±0.0045} & \multicolumn{1}{l}{External} \\
    \midrule
    BRCA\_MolSubtype (out) & 0.6924±0.0213 & 0.7570±0.0215 & \pmb{0.8123±0.0196} & 0.7756±0.0219 & 0.7923±0.0208 & 0.7270±0.0202 & 0.8057±0.0191 & \multicolumn{1}{l}{Held-out} \\
    CRC\_MolSubtype (out) & 0.6286±0.0676 & 0.7440±0.0648 & 0.8365±0.0565 & 0.8549±0.0470 & 0.8520±0.0445 & 0.7697±0.0619 & \pmb{0.8556±0.0468} & \multicolumn{1}{l}{Held-out} \\
    GBMLGG\_MolSubtype (out) & 0.9335±0.0420 & 0.9462±0.0371 & 0.9567±0.0378 & 0.9763±0.0226 & 0.9513±0.0337 & 0.9837±0.0150 & \pmb{0.9847±0.0162} & \multicolumn{1}{l}{Held-out} \\
    TCGA\_HNSC\_HPV (out) & 0.8521±0.0585 & 0.9819±0.0098 & 0.9746±0.0141 & 0.9311±0.0253 & 0.9773±0.0114 & 0.9046±0.0357 & \pmb{0.9836±0.0114} & \multicolumn{1}{l}{Held-out} \\
    \midrule
    Avg   & 0.7265±0.1138 & 0.8208±0.1112 & 0.8634±0.0841 & 0.8687±0.0746 & 0.8672±0.0745 & 0.8231±0.1076 & \pmb{0.8865±0.0786} &  \\
    \bottomrule
    \end{tabular}%
    }
  \label{tab:molecular}%
\end{table*}%

\begin{table*}[htbp]
  \centering
  \caption{\textbf{Positive Rates} of mutation for every gene. The gene marked with * refers to the one related to FDA-approved drugs.}
    \begin{tabular}{l|l|c}
    \toprule
    \pmb{Source} & \pmb{Datasets} & \pmb{Positive Rate} \\
    \midrule
    \multirow{18}{*}{TCGA} & \pmb{BRCA\_GATA3} & 12.5\% \\
    ~& \pmb{BRCA\_PIK3CA*} & 36.0\% \\
    ~& \pmb{BRCA\_TP53} & 32.6\% \\
    ~& \pmb{BRCA\_TTN} & 17.5\% \\
    ~& \pmb{CRC\_APC} & 77.2\% \\
    ~& \pmb{GBM\_EGFR} & 22.9\% \\
    ~& \pmb{GBM\_TP53} & 31.3\% \\
    ~& \pmb{HNSC\_TP53} & 74.2\% \\
    ~& \pmb{LGG\_CIC} & 21.0\% \\
    ~& \pmb{LUAD\_EGFR*} & 12.5\% \\
    ~& \pmb{LUAD\_KRAS*} & 29.7\% \\
    ~& \pmb{LUSC\_TP53} & 50.9\% \\
    ~& \pmb{NSCLC\_TMB*} & 22.2\% \\
    ~& \pmb{SKCM\_BRAF*} & 53.4\% \\
    ~& \pmb{SKCM\_DNAH5} & 52.2\% \\
    ~& \pmb{UCEC\_ARID1A} & 43.3\% \\
    ~& \pmb{UCEC\_PTEN} & 64.7\% \\
    ~& \pmb{UCEC\_TTN} & 39.5\% \\
    \midrule
    \multirow{5}{*}{CPTAC} & \pmb{BRCA\_PIK3CA*} & 36.2\% \\
    ~ & \pmb{BRCA\_TP53} & 32.8\% \\
    ~ & \pmb{BRCA\_TTN} & 30.0\% \\
    ~ & \pmb{LUAD\_KRAS*} & 32.6\% \\
    ~ & \pmb{LUAD\_EGFR*} & 30.3\% \\
    \bottomrule
    \end{tabular}%
  \label{tab:pos_rate}%
\end{table*}%

\begin{table*}[htbp]
  \centering
  \caption{\textbf{Label Distribution of IHC Biomarker Prediction} for binary classification.}
    \begin{tabular}{l|cc|c}
    \toprule
    Dataset & Negative     & Positive     & Sum \\
    \midrule
    TCGA\_BRCA\_ER  & 214   & 735   & 949 \\
    TCGA\_BRCA\_HER2  & 404   & 142   & 546 \\
    TCGA\_BRCA\_PR  & 307   & 641   & 948 \\
    IHC\_ZJ1\_ER  & 499   & 1049  & 1548 \\
    IHC\_ZJ1\_PR  & 623   & 933   & 1556 \\
    IHC\_ZJ1\_HER2  & 164   & 1180  & 1344 \\
    IHC\_ZJ1\_CK5 & 753   & 208   & 961 \\
    \bottomrule
    \end{tabular}%
  \label{tab:pos_rate_ihc}%
\end{table*}%

\begin{table*}[htbp]
  \centering
  \caption{\textbf{Label Distribution of IHC Biomarker Expression Level Prediction}.}
    \begin{tabular}{l|cccc|c}
    \toprule
    Dataset & Level 0 & Level 1 & Level 2 & Level 3 & Sum \\
    \midrule
    \textcolor[rgb]{ .063,  .282,  .38}{IHC\_ZJ1\_ER \_Level} & 499   & 268   & 421   & 360   & 1548 \\
    IHC\_ZJ1\_HER2 \_Level & 164   & 347   & 424   & 409   & 1344 \\
    \textcolor[rgb]{ .063,  .282,  .38}{IHC\_NF1\_CK7 } & 224   & 24    & 171   & -     & 419 \\
    \bottomrule
    \end{tabular}%
  \label{tab:ihc_level}%
\end{table*}%

\begin{table*}[htbp]
  \centering
  \caption{\textbf{Macro-AUC of Zero-shot Slide Classification on 6 datasets}. Best performing model for each metric is in bold. 95\% CI is included in parentheses.}
  \scalebox{0.8}{
     \begin{tabular}{l|ccc}
    \toprule
    dataset & PLIP  & CONCH & mSTAR(Ours) \\
    \midrule
    CAMELYON & 0.622 (0.534,0.710) & 0.638 (0.550,0.726) & \pmb{0.742 (0.668,0.817)} \\
    PANDA & 0.561 (0.547,0.575) & \pmb{0.582 (0.566,0.598)} & 0.462 (0.446,0.478) \\
    UBC-OCEAN & \pmb{0.762 (0.691,0.833)} & 0.370 (0.307,0.433) & 0.692 (0.621,0.763) \\
    BCNB-ER & 0.563 (0.473,0.653) & 0.500 (0.500,0.500) & \pmb{0.725 (0.641,0.809)} \\
    BCNB-PR & 0.523 (0.435,0.611) & 0.500 (0.500,0.500) & \pmb{0.722 (0.644,0.800)} \\
    BCNB-HER2 & 0.516 (0.430,0.602) & \pmb{0.518 (0.428,0.608)} & 0.435 (0.345,0.525) \\
    \midrule
    Avg Macro-AUC & 0.5912 & 0.518 & \pmb{0.6297} \\
    \bottomrule
    \end{tabular}%
    }
  \label{tab:zeroshot_overall}%
\end{table*}%

\begin{table*}[htbp]
  \centering
  \caption{\textbf{Performance (Macro-AUC) of Zero-shot Retrieval for Image-to-Text and Text-to-Image tasks} on one held-out dataset and one external dataset.}
    \begin{tabular}{l|l|ccc|ccc}
    \toprule
    \multicolumn{1}{c|}{\multirow{2}[4]{*}{dataset}} & \multicolumn{1}{c|}{\multirow{2}[4]{*}{model}} & \multicolumn{3}{c|}{Image2Text} & \multicolumn{3}{c}{Text2Image} \\
\cmidrule{3-8}          &       & @5    & @10   & @50   & @5    & @10   & @50 \\
    \midrule
    TCGA-Report (held-out) & PLIP  & 0.0119 & 0.0321 & 0.1536 & 0.0202 & 0.0393 & 0.1774 \\
    TCGA-Report (held-out) & CONCH & 0.0083 & 0.0167 & 0.0655 & 0.0107 & 0.0190 & 0.0810 \\
    TCGA-Report (held-out) & mSTAR (Ours) & \pmb{0.0643} & \pmb{0.1274} & \pmb{0.5262} & \pmb{0.0667} & \pmb{0.1298} & \pmb{0.5405} \\
    \midrule
    Breast\&Lung (external) & PLIP  & 0.0200 & 0.0400 & 0.1880 & 0.0260 & 0.0340 & 0.1220 \\
    Breast\&Lung (external) & CONCH & 0.0060 & 0.0120 & 0.1040 & 0.0080 & 0.0220 & 0.1140 \\
    Breast\&Lung (external) & mSTAR (Ours) & \pmb{0.0360} & \pmb{0.0520} & \pmb{0.2240} & \pmb{0.0340} & \pmb{0.0600} & \pmb{0.2160} \\
    \bottomrule
    \end{tabular}%
  \label{tab:retrieval}%
\end{table*}%

\begin{table*}[htbp]
  \centering
  \caption{\textbf{Performance of Pathological Report Generation} on one held-out dataset and two external datasets. Best performing model for each metric is in bold. 6 metrics are reported.}
  \scalebox{0.85}{
    \begin{tabular}{c|lccccccc|c}
    \toprule
    Source & Metric & R50   & PLIP  & CONCH & UNI   & CHIEF & GigaPath & mSTAR (Ours) & Evaluation \\
    \midrule
    \multirow{6}[2]{*}{TCGA} & BLEU\_1 & 0.2541 & 0.3124 & 0.2447 & 0.3552 & 0.3087 & 0.3545 & \pmb{0.3624} & \multirow{6}[2]{*}{held-out} \\
          & BLEU\_2 & 0.1501 & 0.2078 & 0.1590 & 0.2403 & 0.2028 & 0.2388 & \pmb{0.2474} &  \\
          & BLEU\_3 & 0.0942 & 0.1457 & 0.1092 & 0.1716 & 0.1399 & 0.1694 & \pmb{0.1775} &  \\
          & BLEU\_4 & 0.0558 & 0.1057 & 0.0765 & 0.1274 & 0.0994 & 0.1253 & \pmb{0.1329} &  \\
          & METEOR & 0.1291 & 0.1348 & 0.1177 & 0.1489 & 0.1343 & 0.1491 & \pmb{0.1518} &  \\
          & ROUGE\_L & 0.2185 & 0.2638 & 0.2372 & 0.2931 & 0.2623 & 0.2921 & \pmb{0.3012} &  \\
    \midrule
    \multirow{6}[2]{*}{Nanfang} & BLEU\_1 & 0.1705 & 0.1186 & 0.2136 & 0.2361 & 0.2017 & 0.2097 & \pmb{0.3052} & \multirow{6}[2]{*}{external} \\
\bottomrule
    \end{tabular}%
    }
  \label{tab:report_gen}%
\end{table*}%

\begin{table*}[htbp]
  \centering
  \caption{\textbf{C-Index of Survival Prediction on 16 datasets}, including 10 held-out datasets, 4 external datasets and 2 independent datasets. }
  \scalebox{0.55}{
    \begin{tabular}{cc|cccccccc}
    \toprule
    \multicolumn{1}{c|}{Task} & dataset & R50   & PLIP  & CONCH & UNI   & CHIEF & GigaPath & \multicolumn{1}{c|}{mSTAR (Ours)} & Evaluation \\
    \midrule
    \multicolumn{1}{c|}{\multirow{3}[1]{*}{Disease-Free Survival}} & DFS\_BRCA  & 0.5504±0.1201 & 0.6368±0.0964 & 0.6357±0.1254 & \pmb{0.6695±0.119} & 0.6494±0.1131 & 0.6298±0.1177 & \multicolumn{1}{c|}{0.6663±0.1209} & Held-out \\
    \multicolumn{1}{c|}{} & DFS\_YN1  & 0.5037±0.0691 & 0.5277±0.0722 & 0.4810±0.0704 & 0.5737±0.0717 & 0.5040±0.0715 & 0.4935±0.0774 & \multicolumn{1}{c|}{\pmb{0.5775±0.0698}} & External \\
    \multicolumn{1}{c|}{} & DFS\_ZJ1  & 0.5556±0.0693 & 0.5491±0.0651 & 0.5850±0.0665 & 0.5548±0.0696 & 0.5539±0.0710 & 0.5367±0.0715 & \multicolumn{1}{c|}{\pmb{0.5959±0.0655}} & External \\
    \hline
    \multicolumn{1}{c|}{\multirow{12}[1]{*}{Overall Survival}} & OS\_BRCA  & 0.5696±0.1250 & 0.5915±0.1151 & 0.6940±0.0888 & 0.6908±0.1048 & 0.6698±0.0972 & 0.6528±0.1067 & \multicolumn{1}{c|}{\pmb{0.7076±0.0896}} & Held-out \\
    \multicolumn{1}{c|}{} & OS\_CRC  & 0.5710±0.1148 & 0.7076±0.0890 & 0.7026±0.0877 & 0.6906±0.0835 & \pmb{0.7288±0.0870} & 0.6815±0.1011 & \multicolumn{1}{c|}{0.6895±0.0836} & Held-out \\
    \multicolumn{1}{c|}{} & OS\_GBMLGG  & 0.7810±0.0385 & 0.7716±0.0409 & 0.7704±0.0435 & 0.7905±0.0434 & 0.7876±0.0389 & 0.7839±0.0415 & \multicolumn{1}{c|}{\pmb{0.7923±0.0426}} & Held-out \\
    \multicolumn{1}{c|}{} & OS\_HANCOCK  & 0.5084±0.0425 & 0.5343±0.0439 & 0.5198±0.0417 & 0.5573±0.0450 & 0.5062±0.0423 & 0.5378±0.0447 & \multicolumn{1}{c|}{\pmb{0.5639±0.0448}} & External \\
\bottomrule
    \end{tabular}%
    }
  \label{tab:survival}%
\end{table*}%

\begin{table*}[htbp]
  \centering
  \caption{\textbf{C-Index of Multimodal Survival Analysis} on average across 4 multimodal fusion approaches on 9 TCGA held-out datasets. Best performing model for each metric is bolded.}
  \scalebox{0.75}{
    \begin{tabular}{l|cllllll}
    \toprule
    dataset & R50   & \multicolumn{1}{c}{PLIP} & \multicolumn{1}{c}{CONCH} & \multicolumn{1}{c}{UNI} & \multicolumn{1}{c}{CHIEF} & \multicolumn{1}{c}{GigaPath} & \multicolumn{1}{c}{mSTAR (Ours)} \\
    \midrule
    BRCA  & 0.7182±0.0125 & 0.7080±0.0224 & 0.6876±0.0397 & 0.7178±0.0234 & 0.7006±0.0124 & 0.6905±0.0293 & \pmb{0.7400±0.0081} \\
    CRC   & 0.6674±0.0338 & 0.6665±0.0022 & 0.6594±0.0051 & 0.6661±0.0133 & 0.6683±0.0212 & 0.6818±0.0171 & \pmb{0.7073±0.0100} \\
    GBMLGG & 0.8789±0.0046 & 0.8772±0.0033 & 0.8840±0.0039 & 0.8918±0.0075 & 0.8871±0.0018 & 0.8899±0.0026 & \pmb{0.8948±0.0065} \\
    HNSC  & 0.6838±0.0087 & 0.6691±0.0145 & 0.6700±0.0039 & 0.6858±0.0159 & 0.6894±0.0070 & 0.6731±0.0104 & \pmb{0.7111±0.0131} \\
    KIRC  & 0.6880±0.0158 & 0.6797±0.0068 & 0.6922±0.0073 & 0.6977±0.0125 & 0.6915±0.0060 & 0.6876±0.0111 & \pmb{0.7017±0.0106} \\
    LUAD  & 0.6499±0.0102 & 0.6375±0.0117 & 0.6523±0.0223 & 0.6495±0.0165 & 0.6575±0.0199 & 0.6706±0.0338 & \pmb{0.6708±0.0288} \\
    LUSC  & 0.5457±0.0102 & 0.5349±0.0190 & 0.5358±0.0278 & 0.5713±0.0156 & 0.5550±0.0175 & 0.5689±0.0373 & \pmb{0.5902±0.0278} \\
    SKCM  & \pmb{0.6570±0.0132} & 0.6502±0.0169 & 0.6411±0.0099 & 0.6399±0.0240 & 0.6452±0.0195 & 0.6266±0.0546 & 0.6476±0.0191 \\
    UCEC  & 0.6498±0.0303 & 0.6564±0.0239 & 0.6814±0.0171 & 0.6928±0.0274 & 0.6453±0.0293 & 0.6670±0.0479 & \pmb{0.7111±0.0352} \\
    \midrule
    Avg   & 0.6821±0.0827 & 0.6755±0.0844 & 0.6782±0.0853 & 0.6903±0.0819 & 0.6822±0.0831 & 0.6840±0.0814 & \pmb{0.7083±0.0781} \\
    \bottomrule
    \end{tabular}%
    }
  \label{tab:fusion_box}%
\end{table*}%
\begin{table*}[htbp]
  \centering
  \caption{\textbf{C-Index of Multimodal Survival Analysis with MCAT on 9 TCGA held-out datasets}. Best performing model for each metric is bolded. 95\% CI is included in parentheses.}
  \scalebox{0.55}{
    \begin{tabular}{l|ccccccc}
    \toprule
    \pmb{dataset} & \pmb{R50} & \pmb{PLIP} & \pmb{CONCH} & \pmb{UNI} & \pmb{CHIEF} & \pmb{GigaPath} & \pmb{mSTAR (Ours)} \\
    \midrule
    \pmb{BRCA} & 0.7215 (0.6382, 0.8048) & 0.7304 (0.6428, 0.8180) & 0.6266 (0.5019, 0.7513) & 0.6928 (0.6034, 0.7822) & 0.6850 (0.5892, 0.7808) & 0.6556 (0.5566, 0.7546) & \pmb{0.7489 (0.6703, 0.8275)} \\
    \pmb{CRC} & \pmb{0.6996 (0.5994, 0.7998)} & 0.6676 (0.5669, 0.7683) & 0.6635 (0.5601, 0.7669) & 0.6481 (0.5474, 0.7488) & 0.6935 (0.6020, 0.7850) & 0.6897 (0.5837, 0.7957) & 0.6989 (0.5943, 0.8035) \\
    \pmb{GBMLGG} & 0.8713 (0.8343, 0.9083) & 0.8784 (0.8433, 0.9135) & 0.8886 (0.8553, 0.9219) & 0.8810 (0.8471, 0.9149) & 0.8872 (0.8574, 0.9170) & 0.8885 (0.8599, 0.9171) & \pmb{0.8896 (0.8610, 0.9182)} \\
    \pmb{HNSC} & 0.6830 (0.6054, 0.7606) & 0.6444 (0.5677, 0.7211) & 0.6646 (0.6008, 0.7284) & 0.6874 (0.6125, 0.7623) & 0.6839 (0.6157, 0.7521) & 0.6659 (0.5987, 0.7331) & \pmb{0.6907 (0.6157, 0.7657)} \\
    \pmb{KIRC} & 0.6856 (0.6130, 0.7582) & 0.6830 (0.6088, 0.7572) & 0.6996 (0.6335, 0.7657) & \pmb{0.7091 (0.6291, 0.7891)} & 0.6962 (0.6200, 0.7724) & 0.6968 (0.6149, 0.7787) & 0.7084 (0.6407, 0.7761) \\
    \pmb{LUAD} & 0.6375 (0.5326, 0.7424) & 0.6505 (0.5576, 0.7434) & 0.6628 (0.5681, 0.7575) & 0.6595 (0.5733, 0.7457) & 0.6823 (0.5943, 0.7703) & 0.6546 (0.5731, 0.7361) & \pmb{0.6953 (0.6073, 0.7833)} \\
    \pmb{LUSC} & 0.5373 (0.4420, 0.6326) & 0.5499 (0.4654, 0.6344) & 0.5526 (0.4660, 0.6392) & 0.5493 (0.4488, 0.6498) & \pmb{0.5823 (0.4841, 0.6805)} & 0.5472 (0.4694, 0.6250) & 0.5537 (0.4590, 0.6484) \\
\bottomrule
    \end{tabular}%
    }
  \label{tab:fusion_mcat}%
\end{table*}%

\begin{table*}[htbp]
  \centering
  \caption{\textbf{C-Index of Multimodal Survival Analysis with Porpoise on 9 TCGA held-out datasets}. Best performing model for each metric is bolded. 95\% CI is included in parentheses.}
  \scalebox{0.55}{
    \begin{tabular}{l|ccccccc}
    \toprule
    \pmb{dataset} & \pmb{R50} & \pmb{PLIP} & \pmb{CONCH} & \pmb{UNI} & \pmb{CHIEF} & \pmb{GigaPath} & \pmb{mSTAR (Ours)} \\
    \midrule
    \pmb{BRCA} & 0.7366 (0.6480, 0.8252) & 0.7244 (0.6297, 0.8191) & 0.7259 (0.6369, 0.8149) & \pmb{0.7474 (0.6651, 0.8297)} & 0.7196 (0.6353, 0.8039) & 0.7312 (0.6434, 0.8190) & 0.7456 (0.6515, 0.8397) \\
    \pmb{CRC} & 0.6113 (0.4978, 0.7248) & 0.6639 (0.5571, 0.7707) & 0.6517 (0.5535, 0.7499) & 0.6599 (0.5621, 0.7577) & 0.6351 (0.5224, 0.7478) & 0.6579 (0.5591, 0.7567) & \pmb{0.7020 (0.6085, 0.7955)} \\
    \pmb{GBMLGG} & 0.8805 (0.8472, 0.9138) & 0.8748 (0.8407, 0.9089) & 0.8786 (0.8406, 0.9166) & 0.8948 (0.8627, 0.9269) & 0.8898 (0.8547, 0.9249) & 0.8888 (0.8557, 0.9219) & \pmb{0.8955 (0.8643, 0.9267)} \\
    \pmb{HNSC} & 0.6982 (0.6271, 0.7693) & 0.6806 (0.5977, 0.7635) & 0.6730 (0.5981, 0.7479) & 0.7111 (0.6415, 0.7807) & 0.6858 (0.6152, 0.7564) & 0.6860 (0.6164, 0.7556) & \pmb{0.7189 (0.6474, 0.7904)} \\
    \pmb{KIRC} & 0.6681 (0.5934, 0.7428) & 0.6715 (0.5921, 0.7509) & 0.6802 (0.5973, 0.7631) & 0.6767 (0.6046, 0.7488) & 0.6831 (0.6076, 0.7586) & 0.6693 (0.5870, 0.7516) & \pmb{0.6834 (0.6115, 0.7553)} \\
    \pmb{LUAD} & \pmb{0.6656 (0.5792, 0.7520)} & 0.6222 (0.5266, 0.7178) & 0.6139 (0.5265, 0.7013) & 0.6295 (0.5454, 0.7136) & 0.6413 (0.5527, 0.7299) & 0.6243 (0.5369, 0.7117) & 0.6300 (0.5455, 0.7145) \\
    \pmb{LUSC} & 0.5364 (0.4490, 0.6238) & 0.5100 (0.4136, 0.6064) & 0.4911 (0.3980, 0.5842) & 0.5794 (0.4859, 0.6729) & 0.5572 (0.4617, 0.6527) & 0.5469 (0.4601, 0.6337) & \pmb{0.5895 (0.5027, 0.6763)} \\
\bottomrule
    \end{tabular}%
  }
  \label{tab:fusion_porpoise}%
\end{table*}%

\begin{table*}[htbp]
  \centering
  \caption{\textbf{C-Index of Multimodal Survival Analysis with MOTCat on 9 TCGA held-out datasets}. Best performing model for each metric is bolded. 95\% CI is included in parentheses.}
  \scalebox{0.55}{
    \begin{tabular}{l|ccccccc}
    \toprule
    \pmb{dataset} & \pmb{R50} & \pmb{PLIP} & \pmb{CONCH} & \pmb{UNI} & \pmb{CHIEF} & \pmb{GigaPath} & \pmb{mSTAR (Ours)} \\
    \midrule
    \pmb{BRCA} & 0.7030 (0.6075, 0.7985) & 0.6732 (0.5685, 0.7779) & 0.6781 (0.5789, 0.7773) & 0.6970 (0.5814, 0.8126) & 0.6979 (0.5893, 0.8065) & 0.6711 (0.5723, 0.7699) & \pmb{0.7278 (0.6276, 0.8280)} \\
    \pmb{CRC} & 0.6867 (0.5926, 0.7808) & 0.6695 (0.5635, 0.7755) & 0.6643 (0.5553, 0.7733) & 0.6835 (0.5765, 0.7905) & 0.6759 (0.5554, 0.7964) & 0.7040 (0.6056, 0.8024) & \pmb{0.7244 (0.6327, 0.8161)} \\
    \pmb{GBMLGG} & 0.8838 (0.8526, 0.9150) & 0.8736 (0.8375, 0.9097) & 0.8867 (0.8555, 0.9179) & \pmb{0.8899 (0.8583, 0.9215)} & 0.8863 (0.8545, 0.9181) & 0.8878 (0.8547, 0.9209) & 0.8889 (0.8583, 0.9195) \\
    \pmb{HNSC} & 0.6776 (0.6045, 0.7507) & 0.6731 (0.6023, 0.7439) & 0.6679 (0.5852, 0.7506) & 0.6736 (0.5930, 0.7542) & 0.7014 (0.6303, 0.7725) & 0.6801 (0.6091, 0.7511) & \pmb{0.7256 (0.6546, 0.7966)} \\
    \pmb{KIRC} & \pmb{0.7124 (0.6467, 0.7781)} & 0.6890 (0.6108, 0.7672) & 0.6927 (0.6163, 0.7691) & 0.7050 (0.6286, 0.7814) & 0.6980 (0.6198, 0.7762) & 0.6878 (0.6170, 0.7586) & 0.7085 (0.6403, 0.7767) \\
    \pmb{LUAD} & 0.6463 (0.5559, 0.7367) & 0.6472 (0.5549, 0.7395) & 0.6685 (0.5758, 0.7612) & 0.6710 (0.5875, 0.7545) & 0.6715 (0.5921, 0.7509) & \pmb{0.7125 (0.6261, 0.7989)} & 0.7004 (0.6138, 0.7870) \\
    \pmb{LUSC} & 0.5618 (0.4748, 0.6488) & 0.5564 (0.4621, 0.6507) & 0.5643 (0.4704, 0.6582) & 0.5654 (0.4780, 0.6528) & 0.5450 (0.4568, 0.6332) & 0.5478 (0.4614, 0.6342) & \pmb{0.5855 (0.4957, 0.6753)} \\
\bottomrule
    \end{tabular}%
  }
  \label{tab:fusion_motcat}%
\end{table*}%

\begin{table*}[htbp]
  \centering
  \caption{\textbf{C-Index of Multimodal Survival Analysis with CMTA on 9 TCGA held-out datasets}. Best performing model for each metric is bolded. 95\% CI is included in parentheses.}
  \scalebox{0.55}{
    \begin{tabular}{l|ccccccc}
    \toprule
    \pmb{dataset} & \pmb{R50} & \pmb{PLIP} & \pmb{CONCH} & \pmb{UNI} & \pmb{CHIEF} & \pmb{GigaPath} & \pmb{mSTAR (Ours)} \\
    \midrule
    \pmb{BRCA} & 0.7118 (0.6224, 0.8012) & 0.7039 (0.5945, 0.8133) & 0.7196 (0.6157, 0.8235) & 0.7339 (0.6351, 0.8327) & 0.6998 (0.6098, 0.7898) & 0.7040 (0.6023, 0.8057) & \pmb{0.7375 (0.6560, 0.8190)} \\
    \pmb{CRC} & 0.6719 (0.5610, 0.7828) & 0.6651 (0.5514, 0.7788) & 0.6580 (0.5469, 0.7691) & 0.6727 (0.5714, 0.7740) & 0.6687 (0.5623, 0.7751) & 0.6754 (0.5790, 0.7718) & \pmb{0.7040 (0.6080, 0.8000)} \\
    \pmb{GBMLGG} & 0.8798 (0.8447, 0.9149) & 0.8820 (0.8479, 0.9161) & 0.8822 (0.8501, 0.9143) & 0.9016 (0.8726, 0.9306) & 0.8849 (0.8494, 0.9204) & 0.8944 (0.8640, 0.9248) & \pmb{0.9051 (0.8798, 0.9304)} \\
    \pmb{HNSC} & 0.6764 (0.6051, 0.7477) & 0.6782 (0.6059, 0.7505) & 0.6744 (0.5964, 0.7524) & 0.6710 (0.5928, 0.7492) & 0.6864 (0.6062, 0.7666) & 0.6602 (0.5871, 0.7333) & \pmb{0.7092 (0.6430, 0.7754)} \\
    \pmb{KIRC} & 0.6858 (0.6158, 0.7558) & 0.6752 (0.5984, 0.7520) & 0.6963 (0.6175, 0.7751) & 0.7000 (0.6100, 0.7900) & 0.6888 (0.6243, 0.7533) & 0.6963 (0.6203, 0.7723) & \pmb{0.7065 (0.6279, 0.7851)} \\
    \pmb{LUAD} & 0.6502 (0.5536, 0.7468) & 0.6301 (0.5386, 0.7216) & 0.6639 (0.5645, 0.7633) & 0.6380 (0.5376, 0.7384) & 0.6348 (0.5468, 0.7228) & \pmb{0.6911 (0.6033, 0.7789)} & 0.6574 (0.5684, 0.7464) \\
    \pmb{LUSC} & 0.5473 (0.4569, 0.6377) & 0.5234 (0.4303, 0.6165) & 0.5352 (0.4413, 0.6291) & 0.5909 (0.4980, 0.6838) & 0.5355 (0.4465, 0.6245) & \pmb{0.6335 (0.5402, 0.7268)} & 0.6320 (0.5497, 0.7143) \\
\bottomrule
    \end{tabular}%
  }
  \label{tab:fusion_cmta}%
\end{table*}%

\begin{table*}[h]
    \centering
    \caption{\textbf{Data splits (cases) of 9 cancer datasets on TCGA}. CRC=COAD+READ and GBMLGG=GBM+LGG. Apart from these specific cancer types, other TCGA datasets were used for pretraining without data partitioning.}
    \begin{tabular}{l|ccc|r}
    \toprule
        Cancer Type & Train & Validation & Test & Total \\
    \midrule
        BRCA & 716 & 102 & 307 & 1023 \\
        CRC & 405 & 58 & 116 & 579 \\
        GBMLGG & 581 & 83 & 166 & 830 \\
        HNSC & 308 & 44 & 89 & 441 \\
        KIRC & 348 & 50 & 100 & 498 \\
        LUAD & 318 & 45 & 92 & 455 \\
        LUSC & 316 & 45 & 91 & 452 \\
        SKCM & 290 & 41 & 84 & 415 \\
        UCEC & 346 & 49 & 100 & 495 \\
    \bottomrule
    \end{tabular}
    
    \label{tab:tcga_splits}
\end{table*}
\begin{table*}[h]
    \centering
    \caption{\textbf{Prompts} used in GPT-4 for cleaning pathology reports.}
    \begin{tabular}{l}
    \\
     \textit{Helping me to check the formatting and spelling of the supplied text,} \\
     \textit{including some incorrect use of punctuation like mis-using of 'X' and 'x'} \\
     \textit{and capitalization and deletion of some words of unknown meaning as well.} \\
     \\
    \end{tabular}
    \label{tab:prompt_gpt4}
\end{table*}

\begin{table*}[h]
    \centering
    \caption{\textbf{Prompts} used in GPT-4o-mini to filter out information unseen under microscopic examination from pathology reports for the task of report generation.}
    \begin{tabular}{l}
    \\
     \textit{I need you to be a professional pathologist and help me organize the contents of the pathology report.} \\
     \textit{The report I give may be a chaos in organization and have information I don't need.} \\
     \textit{You need to remove the following contents: 1. patient's information like patient's id.} \\
     \textit{Only keep the following contents if available: 1. Microscopic description: a detailed description of}\\
     \textit{what the pathologist sees during microscopic exam of the specimen 2. Margins Information:}\\
     \textit{The margins are an area of normal cells around a tumor. 3. stage and grade}\\
     \textit{4. Lymph nodes included in a biopsy 5. Diagnosis}
     \\
    \end{tabular}
    \label{tab:prompt_repogen}
\end{table*}

\begin{table*}[h]
    \centering
    
    \caption{\textbf{Hyperparameters of ABMIL used for Slide-level Prediction}. A single 80GB NVIDIA H800 GPU was used for training.}
    \begin{tabular}{l|r}
    \toprule
        Hyperparameter & Value \\
    \midrule
        Input dim & 512/1024 \\
        Hidden dim & 512 \\
        Dropout & 0.25 \\
    \midrule
        Batch size & 1 \\
        Epochs & 30 \\
        Optimizer & Adam \\
        Learning rate & 2e-4\\
        Scheduler & Cosine\\
        Weight decay & 1e-5\\ 
    \bottomrule
    
    \end{tabular}
    \label{tab:mil_cls}
\end{table*}

\begin{table*}[h]
    \centering
    \caption{\textbf{Architecture Hyperparameters of Porpoise used for Multimodal Survival Analysis}. A single 80GB NVIDIA H800 GPU was used for training. $D$ is the number of genes, varying across different datasets.}
    \begin{tabular}{l|r r}
    \toprule
        Hyperparameter & WSI & RNASeq \\
    \midrule
        Input dim & 512/1024 & $D$ \\
        Hidden dim & $512 \rightarrow 256$ & 256 \\
        Dropout & 0.1 & 0.1 \\
    \midrule
        Feature dim after fusion & \multicolumn{2}{c}{256} \\
    \bottomrule
        
    \end{tabular}
    
    \label{tab:mm_porpoise}
\end{table*}

\begin{table*}[h]
    \centering
    \caption{\textbf{Architecture Hyperparameters of MCAT and MOTCat used for Multimodal Survival Analysis}. A single 80GB NVIDIA H800 GPU was used for training. $D$ is the number of genes, varying across different unique functional categories and different datasets. $256\times 2 \rightarrow 256$ refers to 256-dimensional features from two modalities were fused into a 256-dimensional slide-level feature.}
    \begin{tabular}{l|r r}
    \toprule
        Hyperparameter & WSI & RNASeq \\
    \midrule
        Input dim & 512/1024 & $6\times D$ \\
        Hidden dim & 256 & $6 \times 256$ \\
        Dropout & 0.25 & 0.25 \\
    \midrule
        Feature dim after fusion & \multicolumn{2}{c}{$256\times 2 \rightarrow 256$} \\
    \bottomrule
        
    \end{tabular}
    
    \label{tab:mm_mcat_motcat}
\end{table*}

\begin{table*}[h]
    \centering
    \caption{\textbf{Architecture Hyperparameters of CMTA used for Multimodal Survival Analysis}. A single 80GB NVIDIA H800 GPU was used for training. $D$ is the number of genes, varying across different datasets. $256\times 2 \rightarrow 256$ refers to 256-dimensional features from two modalities were fused into a 256-dimensional slide-level feature.}
    \begin{tabular}{l|r r}
    \toprule
        Hyperparameter & WSI & RNASeq \\
    \midrule
        Input dim & 512/1024 & $6\times D$ \\
        Hidden dim & 256 & $6 \times (1024 \rightarrow 256)$ \\
        Dropout & 0.25 & 0.25 \\
    \midrule
        Feature dim after fusion & \multicolumn{2}{c}{$256\times 2 \rightarrow 256$} \\
    \bottomrule
        
    \end{tabular}
    
    \label{tab:mm_cmta}
\end{table*}

\begin{table*}[h]
    \centering
    \caption{\textbf{Training Hyperparameters used in Multimodal Survival Analysis}. A single 80GB NVIDIA H800 GPU was used for training.}
    \begin{tabular}{l|r}
    \toprule
        Hyperparameter & Value \\
    \midrule
        Batch size & 1 \\
        Epochs & 30 \\
        Optimizer & Adam \\
        Learning rate & 2e-4\\
        Scheduler & Cosine\\
        Weight decay & 1e-5\\ 
    \bottomrule
        
    \end{tabular}
    
    \label{tab:mm_train}
\end{table*}

\begin{table*}[h]
    \centering
    \caption{\textbf{Architecture Hyperparameters of HistGen used for Pathology Report Generation}. A single 80GB NVIDIA H800 GPU was used for training. $D$ is the feature dimension of pathological patch, varying across different FMs.}
    \begin{tabular}{l|r}
    \toprule
        Hyperparameter & Value \\
    \midrule
        Input dim & $D$ \\
        Hidden dim & 512 \\
        Layers of decoder & 3 (8 heads) \\
        Dimension of cross-modal context & $512\times 2048$ \\
    \midrule
        Optimizer & Adam \\
        Learning rate & 1e-4 \\
        Beam size & 3 \\
    \bottomrule
        
    \end{tabular}
    
    \label{tab:param_histgen}
\end{table*}
\begin{table*}[h]
    \centering
    \caption{Prompt templates for all tasks that required prompts. In these templates, the placeholder "CLASSNAME" was replaced with the actual name of the class. The class prompts of CLASSNAME for each task are presented in ~\ref{tab:cls_camel}-\ref{tab:cls_her2}.}
    \begin{tabular}{c|l}
    \toprule
        \# & Prompt \\
    \midrule
        1 & CLASSNAME.  \\
        2 & a photomicrograph showing CLASSNAME. \\
        3 & a photomicrograph of CLASSNAME. \\
        4 & an image of CLASSNAME. \\
        5 & an image showing CLASSNAME. \\
        6 & an example of CLASSNAME. \\
        7 & CLASSNAME is shown. \\
\bottomrule
    \end{tabular}
    
    \label{tab:prompts}
\end{table*}

\begin{table*}[h]
    \centering
    \caption{Class prompts for Breast Metastasis Detection.}
    \begin{tabular}{l|l|l}
    \toprule
        Task & Class & CLASSNAME \\
    \midrule
        \multirow{8}{*}{Breast Metastasis Detection} & \multirow{4}{*}{Tumor} & metastatic cancer \\
         & ~ & lymph node with metastatic cancer \\
         & ~ & metastatic breast cancer in lymph node \\
         & ~ & tumor \\
    \cmidrule{2-3}
         & \multirow{4}{*}{Normal} & non-metastatic \\
         & ~ & lymph node without metastatic cancer \\
         & ~ & non-metastatic lymph node \\
         & ~ & normal \\
    \bottomrule
    \end{tabular}
    
    \label{tab:cls_camel}
\end{table*}

\begin{table*}[h]
    \centering
    \caption{Class prompts of PANDA for Prostate ISUP grading.}
    \begin{tabular}{l|c|l}
    \toprule
    Task  & Class & CLASSNAME \\
    \midrule
    \multirow{24}[12]{*}{Prostate ISUP grading} & \multirow{4}[2]{*}{0} & benign tissue, \\
          &       & healthy tissue, \\
          &       & non-cancerous tissue, \\
          &       & normal tissue \\
\cmidrule{2-3}          & \multirow{4}[2]{*}{1} & Gleason score 6, \\
          &       & ISUP grade 1, \\
          &       & low-grade cancer, \\
          &       & Gleason pattern 3+3 \\
\cmidrule{2-3}          & \multirow{4}[2]{*}{2} & Gleason score 7 (3+4), \\
          &       & ISUP grade 2, \\
          &       & moderately differentiated cancer, \\
          &       & Gleason pattern 3+4 \\
\cmidrule{2-3}          & \multirow{4}[2]{*}{3} & Gleason score 7 (4+3), \\
          &       & ISUP grade 3, \\
          &       & moderately differentiated cancer, \\
          &       & Gleason pattern 4+3 \\
\cmidrule{2-3}          & \multirow{4}[2]{*}{4} & Gleason score 8, \\
          &       & ISUP grade 4, \\
          &       & high-grade cancer, \\
          &       & Gleason pattern 4+4, 3+5, or 5+3 \\
\cmidrule{2-3}          & \multirow{4}[2]{*}{5} & Gleason score 9 or 10, \\
          &       & ISUP grade 5, \\
          &       & high-grade aggressive cancer, \\
          &       & Gleason pattern 4+5, 5+4, or 5+5 \\
    \bottomrule
    \end{tabular}%
    
    \label{tab:cls_panda}
\end{table*}

\begin{table*}[htbp]
  \centering
  \caption{Class prompts of UBC-OCEAN for Ovarian Cancer Subtyping.}
    \begin{tabular}{l|c|l}
    \toprule
    Task  & Class & CLASSNAME \\
    \midrule
    \multicolumn{1}{l|}{\multirow{15}[10]{*}{Ovarian Cancer Subtyping}} & \multirow{3}[2]{*}{HGSC} & high-grade serous carcinoma, \\
          &       & ovarian high-grade serous carcinoma, \\
          &       & high-grade serous carcinoma of ovary \\
\cmidrule{2-3}          & \multirow{3}[2]{*}{LGSC} & low-grade serous carcinoma, \\
          &       & ovarian low-grade serous carcinoma, \\
          &       & low-grade serous carcinoma of ovary \\
\cmidrule{2-3}          & \multirow{3}[2]{*}{EC} & endometrioid carcinoma, \\
          &       & ovarian endometrioid carcinoma, \\
          &       & endometrioid carcinoma of ovary \\
\cmidrule{2-3}          & \multirow{3}[2]{*}{CC} & clear cell carcinoma, \\
          &       & ovarian clear cell carcinoma, \\
          &       & clear cell carcinoma of ovary \\
\cmidrule{2-3}          & \multirow{3}[2]{*}{MC} & mucinous carcinoma, \\
          &       & ovarian mucinous carcinoma, \\
          &       & mucinous carcinoma of ovary \\
\bottomrule    \end{tabular}%
  \label{tab:cls_ubc}%
\end{table*}%

\begin{table*}[htbp]
  \centering
  \caption{Class prompts of BCNB for ER Prediction.}
    \begin{tabular}{l|c|l}
    \toprule
    Task  & Class & CLASSNAME \\
    \midrule
    \multicolumn{1}{l|}{\multirow{14}[4]{*}{ER Prediction}} & \multirow{7}[2]{*}{Positive} & ER positive, \\
          &       & ER+, \\
          &       & ER positive breast cancer, \\
          &       & Estrogen Receptor positive, \\
          &       & ER+ breast cancer, \\
          &       & ER positive carcinoma, \\
          &       & Estrogen Receptor positive breast cancer \\
\cmidrule{2-3}          & \multirow{7}[2]{*}{Negative} & ER negative, \\
          &       & ER-, \\
          &       & ER negative breast cancer, \\
          &       & Estrogen Receptor negative, \\
          &       & ER- breast cancer, \\
          &       & ER negative carcinoma, \\
          &       & Estrogen Receptor negative breast cancer \\
    \bottomrule
    \end{tabular}%
  \label{tab:cls_er}%
\end{table*}%

\begin{table*}[htbp]
  \centering
  \caption{Class prompts of BCNB for PR Prediction.}
    \begin{tabular}{l|c|l}
    \toprule
    Task  & Class & CLASSNAME \\
    \midrule
    \multicolumn{1}{l|}{\multirow{14}[4]{*}{PR Prediction}} & \multirow{7}[2]{*}{Positive} & PR positive, \\
          &       & PR+, \\
          &       & PR positive breast cancer, \\
          &       & Progesterone Receptor positive, \\
          &       & PR+ breast cancer, \\
          &       & PR positive carcinoma, \\
          &       & Progesterone Receptor positive breast cancer \\
\cmidrule{2-3}          & \multirow{7}[2]{*}{Negative} & PR negative, \\
          &       & PR-, \\
          &       & PR negative breast cancer, \\
          &       & Progesterone Receptor negative, \\
          &       & PR- breast cancer, \\
          &       & PR negative carcinoma, \\
          &       & Progesterone Receptor negative breast cancer \\
    \bottomrule
    \end{tabular}%
  \label{tab:cls_er}%
\end{table*}%

\begin{table*}[htbp]
  \centering
  \caption{Class prompts of BCNB for HER2 Prediction.}
    \begin{tabular}{l|c|l}
    \toprule
    Task  & Class & CLASSNAME \\
    \midrule
    \multicolumn{1}{l|}{\multirow{15}[4]{*}{HER2 Prediction}} & \multirow{8}[2]{*}{Positive} & HER2 positive, \\
          &       & HER2+, \\
          &       & HER2 positive breast cancer, \\
          &       & HER2 overexpression, \\
          &       & HER2+ breast cancer, \\
          &       & HER2 positive carcinoma, \\
          &       & Human Epidermal growth factor Receptor 2 positive, \\
          &       & Human Epidermal growth factor Receptor 2 overexpression \\
\cmidrule{2-3}          & \multirow{7}[2]{*}{Negative} & HER2 negative, \\
          &       & HER2-, \\
          &       & HER2 negative breast cancer, \\
          &       & no HER2 overexpression, \\
          &       & HER2- breast cancer, \\
          &       & HER2 negative carcinoma, \\
          &       & Human Epidermal growth factor Receptor 2 negative \\
    \bottomrule
    \end{tabular}%
  \label{tab:cls_her2}%
\end{table*}%

\begin{table*}[htbp]
  \centering
  \caption{Source of datasets for 97 oncological tasks.}
  \scalebox{0.65}{
    \begin{tabular}{l|l|l|l|l}
    \toprule
    \textcolor[rgb]{ .063,  .282,  .38}{\textbf{No.}} & \textcolor[rgb]{ .063,  .282,  .38}{\textbf{Application}} & \multicolumn{1}{l|}{\textcolor[rgb]{ .063,  .282,  .38}{\textbf{Task}}} & \multicolumn{1}{l|}{\textcolor[rgb]{ .063,  .282,  .38}{\textbf{Dataset}}} & \textcolor[rgb]{ .063,  .282,  .38}{\textbf{URL/Source}} \\
    \midrule
    \rowcolor[rgb]{ .753,  .902,  .961} \textcolor[rgb]{ .063,  .282,  .38}{1} & \textcolor[rgb]{ .063,  .282,  .38}{Pathlogical Diagnosis} & \textcolor[rgb]{ .063,  .282,  .38}{Pathological Subtyping} & \textcolor[rgb]{ .063,  .282,  .38}{BRCA-PathSubtype } & https://portal.gdc.cancer.gov/ \\
    \textcolor[rgb]{ .063,  .282,  .38}{2} & \textcolor[rgb]{ .063,  .282,  .38}{Pathlogical Diagnosis} & \textcolor[rgb]{ .063,  .282,  .38}{Pathological Subtyping} & \textcolor[rgb]{ .063,  .282,  .38}{GBMLGG\_PathSubtype } & https://portal.gdc.cancer.gov/ \\
    \rowcolor[rgb]{ .753,  .902,  .961} \textcolor[rgb]{ .063,  .282,  .38}{3} & \textcolor[rgb]{ .063,  .282,  .38}{Pathlogical Diagnosis} & \textcolor[rgb]{ .063,  .282,  .38}{Pathological Subtyping} & \textcolor[rgb]{ .063,  .282,  .38}{HANCOCK\_PathSubtype } & \textcolor[rgb]{ .063,  .282,  .38}{https://hancock.research.fau.eu/download} \\
    \textcolor[rgb]{ .063,  .282,  .38}{4} & \textcolor[rgb]{ .063,  .282,  .38}{Pathlogical Diagnosis} & \textcolor[rgb]{ .063,  .282,  .38}{Pathological Subtyping} & \textcolor[rgb]{ .063,  .282,  .38}{TCGA-NSCLC } & https://portal.gdc.cancer.gov/ \\
    \rowcolor[rgb]{ .753,  .902,  .961} \textcolor[rgb]{ .063,  .282,  .38}{5} & \textcolor[rgb]{ .063,  .282,  .38}{Pathlogical Diagnosis} & \textcolor[rgb]{ .063,  .282,  .38}{Pathological Subtyping} & \textcolor[rgb]{ .063,  .282,  .38}{EBrains\_PathSubtype } & https://www.ebrains.eu/ \\
\bottomrule
    \end{tabular}%
    }
\end{table*}%

\begin{table*}\ContinuedFloat
\centering
\caption{(Continuous Table) Source of datasets for 97 oncological tasks.}
\scalebox{0.65}{
\begin{tabular}{l|l|l|l|l}
    \toprule
    \textcolor[rgb]{ .063,  .282,  .38}{\textbf{No.}} & \textcolor[rgb]{ .063,  .282,  .38}{\textbf{Application}} & \multicolumn{1}{l|}{\textcolor[rgb]{ .063,  .282,  .38}{\textbf{Task}}} & \multicolumn{1}{l|}{\textcolor[rgb]{ .063,  .282,  .38}{\textbf{Dataset}}} & \textcolor[rgb]{ .063,  .282,  .38}{\textbf{URL}} \\
    \midrule
\textcolor[rgb]{ .063,  .282,  .38}{62} & \textcolor[rgb]{ .063,  .282,  .38}{Survival Prediction} & \textcolor[rgb]{ .063,  .282,  .38}{Overall Survival} & \textcolor[rgb]{ .063,  .282,  .38}{OS\_BRCA } & https://portal.gdc.cancer.gov/ \\
    \rowcolor[rgb]{ .753,  .902,  .961} \textcolor[rgb]{ .063,  .282,  .38}{63} & \textcolor[rgb]{ .063,  .282,  .38}{Survival Prediction} & \textcolor[rgb]{ .063,  .282,  .38}{Overall Survival} & \textcolor[rgb]{ .063,  .282,  .38}{OS\_CRC } & https://portal.gdc.cancer.gov/ \\
    \textcolor[rgb]{ .063,  .282,  .38}{64} & \textcolor[rgb]{ .063,  .282,  .38}{Survival Prediction} & \textcolor[rgb]{ .063,  .282,  .38}{Overall Survival} & \textcolor[rgb]{ .063,  .282,  .38}{OS\_GBMLGG } & https://portal.gdc.cancer.gov/ \\
    \rowcolor[rgb]{ .753,  .902,  .961} \textcolor[rgb]{ .063,  .282,  .38}{65} & \textcolor[rgb]{ .063,  .282,  .38}{Survival Prediction} & \textcolor[rgb]{ .063,  .282,  .38}{Overall Survival} & \textcolor[rgb]{ .063,  .282,  .38}{OS\_HNSC } & https://portal.gdc.cancer.gov/ \\
    \textcolor[rgb]{ .063,  .282,  .38}{66} & \textcolor[rgb]{ .063,  .282,  .38}{Survival Prediction} & \textcolor[rgb]{ .063,  .282,  .38}{Overall Survival} & \textcolor[rgb]{ .063,  .282,  .38}{OS\_KIRC } & https://portal.gdc.cancer.gov/ \\
    \rowcolor[rgb]{ .753,  .902,  .961} \textcolor[rgb]{ .063,  .282,  .38}{67} & \textcolor[rgb]{ .063,  .282,  .38}{Survival Prediction} & \textcolor[rgb]{ .063,  .282,  .38}{Overall Survival} & \textcolor[rgb]{ .063,  .282,  .38}{OS\_LUAD } & https://portal.gdc.cancer.gov/ \\
\bottomrule
    \end{tabular}%
}
    \label{tab:public_source}%
\end{table*}




\end{appendices}



\end{document}